\documentclass[lettersize,journal]{IEEEtran}
\usepackage{amsmath,amsfonts}

\usepackage{array}
\usepackage[caption=false,font=normalsize,labelfont=sf,textfont=sf]{subfig}
\usepackage{textcomp}
\usepackage{stfloats}
\usepackage{url}
\usepackage{verbatim}
\usepackage{graphicx}
\def\BibTeX{{\rm B\kern-.05em{\sc i\kern-.025em b}\kern-.08em
    T\kern-.1667em\lower.7ex\hbox{E}\kern-.125emX}}
\usepackage{balance}
\usepackage{hyperref}
\usepackage{eqnarray}
\usepackage{cite}
\usepackage{nicefrac}
\usepackage{bm}
\usepackage{amsthm}
\usepackage{amssymb}
\usepackage{mathtools}
\usepackage{bbm}
\usepackage{color}
\usepackage{caption}
\usepackage{algorithm}
\usepackage[pdftex,dvipsnames]{xcolor}  % Coloured text etc.
\usepackage[colorinlistoftodos,prependcaption]{todonotes}
\newtheorem{theorem}{Theorem}
\newtheorem{lemma}{Lemma}

\newtheorem{proposition}{Proposition}
\newtheorem{assumption}{Assumption}

\usepackage[noend]{algpseudocode}

\newcommand{\prox}{\textbf{\rm \bf prox}_{\gamma^k F_i}}

\begin{document}

\title{On the Convergence of Federated Learning Algorithms without Data Similarity}
\author{Ali Beikmohammadi,
%\IEEEmembership{Fellow, IEEE}
Sarit Khirirat, and Sindri Magn\'usson 
\IEEEmembership{Member, IEEE}
\thanks{This work was partially supported by the Swedish Research Council through grant agreement no. 2020-03607 and in part by Sweden's Innovation Agency (Vinnova). 
The computations were enabled by resources provided by the National Academic Infrastructure for Supercomputing in Sweden (NAISS) at Chalmers Centre for Computational Science and Engineering (C3SE) partially funded by the Swedish Research Council through grant agreement no. 2022-06725.}
\thanks{A. Beikmohammadi and S. Magn\'usson are with the Department of Computer and System Science, Stockholm University, 16425 Stockholm, Sweden (e-mail: beikmohammadi@dsv.su.se; sindri.magnusson@dsv.su.se).}
\thanks{S. Khirirat contributed to this paper before he joined the King Abdullah University of Science and Technology (KAUST), Thuwal, Saudi Arabia. He is currently a postdoctoral fellow at KAUST (e-mail: sarit.khirirat@kaust.edu.sa).
}
}

\markboth{IEEE Transactions on Big Data, June~2024}%
{On the Convergence of Federated Learning Algorithms without Data Similarity}

\maketitle

\begin{abstract}
Data similarity assumptions have traditionally been relied upon to understand the convergence behaviors of federated learning methods. Unfortunately, this approach often demands fine-tuning step sizes based on the level of data similarity. When data similarity is low, these small step sizes result in an unacceptably slow convergence speed for federated methods. In this paper, we present a novel and unified framework for analyzing the convergence of federated learning algorithms without the need for data similarity conditions. Our analysis centers on an inequality that captures the influence of step sizes on algorithmic convergence performance. By applying our theorems to well-known federated algorithms, we derive precise expressions for three widely used step size schedules: fixed, diminishing, and step-decay step sizes, which are independent of data similarity conditions. Finally, we conduct comprehensive evaluations of the performance of these federated learning algorithms, employing the proposed step size strategies to train deep neural network models on benchmark datasets under varying data similarity conditions. Our findings demonstrate significant improvements in convergence speed and overall performance, marking a substantial advancement in federated learning research.
\end{abstract}


\section{Introduction}
%\paragraph{Federated learning.} 
\IEEEPARstart{F}{ederated} learning has gained significant popularity as a framework for training cutting-edge machine learning models using vast amounts of data collected from numerous resource-constrained devices, such as phones, tablets, and IoT devices. This approach allows these devices to individually train models using their private datasets without compromising sensitive information \cite{mcmahan2017communication}. %,konevcny2016federated
One common implementation of federated learning is the server-worker architecture, where a server aggregates information from local workers to update model parameters, which are then broadcast back to the workers. However, designing effective federated learning methods faces challenges such as dealing with high degrees of systems and statistical heterogeneity, as well as managing communication costs \cite{li2020federated}.
 
A popular approach in federated learning involves designing stochastic and distributed optimization algorithms that facilitate local updating. One such algorithm is FedAvg \cite{mcmahan2017communication}, also known as Local SGD \cite{gorbunov2021local,khaled2020tighter},
which draws inspiration from stochastic gradient descent \cite{needell2014stochastic}.
In this approach, each worker computes its local models based on a stochastic gradient using its private data and then communicates these models to the server responsible for updating the global models. 
Numerous other federated algorithms have emerged to amplify the training efficacy of FedAvg.
For instance, FedProx~\cite{li2020federated,yuan2022convergence} and Proxskip~\cite{mishchenko2022proxskip} utilize proximal updates.
 Similarly, SCAFFOLD~\cite{karimireddy2020scaffold},  FedSplit \cite{pathak2020fedsplit}, and FedPD \cite{zhang2021fedpd} harness variance reduction, operator splitting, and ADMM techniques respectively. FedPD was later refined into FedADMM \cite{gong2022fedadmm} to expedite convergence.

The convergence behaviors of federated optimization algorithms in both homogeneous and heterogeneous data settings have been investigated in the literature. To model data heterogeneity,  existing research often relies on data similarity assumptions \cite{yuan2022convergence,zhang2021fedpd}.
\textcolor{black}{One commonly used assumption measures the similarity between the local gradient at each worker and the global gradient}. However, many existing works, e.g. \cite{li2020federated,karimireddy2020scaffold}, require the step size to be tuned based on the data similarity to establish convergence. As a result, the step sizes tend to become extremely small, especially when the similarity between the local gradient at each worker and the global gradient is low. Moreover, these step sizes are generally impractical to compute since the data similarity is typically unknown in practice. In addition, many of the results only apply when the data similarity is high enough. 
Without these assumptions, the convergence of FedAvg for (strongly-) convex problems and FedADMM for non-convex problems are shown by \cite{gorbunov2021local} and \cite{gong2022fedadmm}, respectively. 
Their existing proof techniques and results cannot be applied to other federated learning algorithms, and they are limited to fixed step size strategies.

\subsection{Contributions}
\begin{figure}[t]
	\centering
\includegraphics[width=1.0\columnwidth]{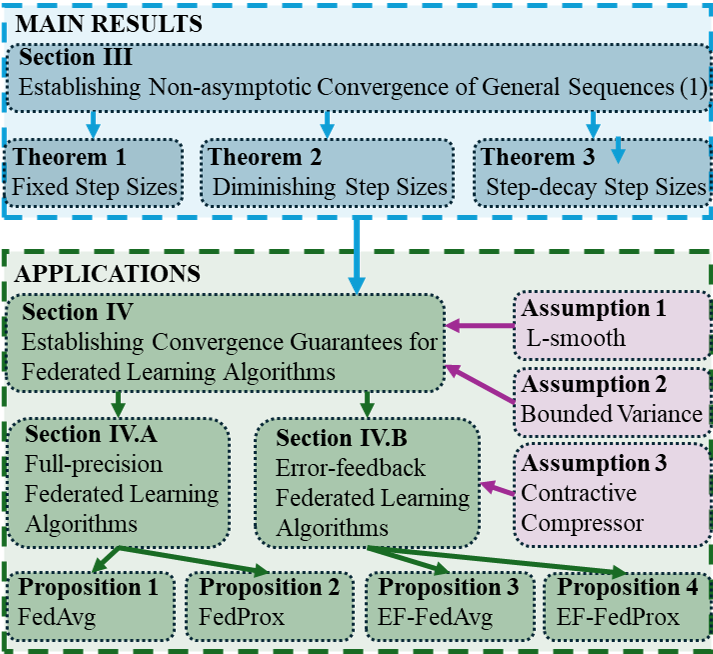} 
	\caption{\textcolor{black}{Visual workflow of our analysis.}}
	\label{vis}
\end{figure}

\noindent {\color{black}
The goal of this paper is to propose a unified framework for analyzing a broad family of federated learning algorithms for non-convex problems without data similarity assumptions, as shown in our analysis workflow in Figure \ref{vis}.}
Our analysis is based on a general descent inequality that captures the convergence behaviors of several federated algorithms of interest. 
We derive novel sequence convergence theorems for three step size schedules commonly used in practice: fixed, diminishing, and step-decay step sizes. 
{\color{black}
By applying these results, we establish convergence guarantees for popular federated algorithms.  
In particular, our convergence bound for FedAvg does not require restrictive assumptions unlike existing works in \cite{yu2019parallel,wang2021cooperative,haddadpour2019local,glasgow2022sharp}, while our result for FedProx does not have the dependency of the step-size on the data similarity parameter in contrast to \cite{li2020federated}.
}
Finally, we demonstrate the effectiveness of these federated learning algorithms for deep neural network training over MNIST and FashionMNIST under different data similarity conditions.

\subsection{Notations}
\noindent For $x,y\in\mathbb{R}^d$, $\langle x,y \rangle:= x^T y$ is the inner product and $\Vert  x \Vert = \sqrt{\langle x,x \rangle}$ is the $\ell_2$-norm. Next, for a real-valued function $f:\mathbb{R}^d\rightarrow \mathbb{R}$, its infimum is denoted by $f^{\inf}$, i.e. $f^{\inf} \leq f(x)$ for any $x \in \mathbb{R}^d$,  while its proximal operator with a positive parameter $\gamma$ is defined by
\begin{align*}
    \textbf{prox}_{\gamma f}(x) := \underset{y\in\mathbb{R}^d}{\text{argmin}} \left\{ f(y) + \frac{1}{2\gamma}\Vert y-x \Vert^2 \right\}. 
\end{align*}
Finally, for the fixed-point operator $\mathcal{T}:\mathbb{R}^d\rightarrow\mathbb{R}^d$ and a positive integer $T$, we denote $\mathcal{T}^T(x)=\underbrace{\mathcal{T}\circ\mathcal{T}\circ\ldots\circ\mathcal{T}}_{T \ \text{times}}(x)$.

\section{Prior Works}\label{sec:Priorwork}

\noindent In this section, we review relevant research that provides context for our work. We cover three key areas: device heterogeneity, efficient communication, and step size schedules.

\subsection{Data Similarity Assumptions}
\noindent Two classical algorithms in federated learning include FedAvg \cite{pmlr-v54-mcmahan17a} and FedProx \cite{li2020federated,yuan2022convergence}. 
While FedAvg updates the global model by averaging local stochastic gradient descent updates,  FedProx computes the model based on an average of local proximal updates.  
The convergence of both algorithms has been extensively analyzed under both homogeneous  and heterogeneous data conditions.
{\color{black}The data heterogeneity is often captured by different assumptions on data similarity which imply theoretical convergence performance of federated and decentralized algorithms \cite{li2020federated,woodworth2020local}.
For instance, FedAvg and FedProx are shown in \cite{woodworth2020local} and \cite{li2020federated}, respectively, to converge slowly especially when the level of data similarity is low. 
While some limited works, e.g. \cite{khaled2020tighter,gorbunov2021local,glasgow2022sharp},  %gong2022fedadmm
have derived convergence results for federated learning algorithms without relying on data similarity assumptions, their results are confined to specific algorithms with fixed step sizes and cannot be extended to analyze other federated algorithms.  
%{\color{black}
 Notably, our work makes a significant contribution by expanding the results in \cite{khaled2020tighter,gorbunov2021local}, which only cover (strongly) convex problems, and by generalizing the results in \cite{glasgow2022sharp}, which requires restrictive assumptions on the Lipschitz continuity of the Hessian and on the bounded $4^\text{th}$-moment of the variance, i.e. $\mathbf{E}\| \nabla F_i(x;\xi)-\nabla f_i(x) \|^4 \leq \sigma^4$ where $\nabla F_i(x;\xi)$ is the unbiased stochastic estimator of $\nabla f_i(x)$. In contrast, our results cover non-convex problems under standard assumptions, frequently encountered in the training of neural networks}. In addition, we study various step size selection strategies, encompassing fixed, diminishing, and step-decay step sizes.

\subsection{Communication-efficient Federated Optimization}
\noindent Communication bandwidth is a major performance bottleneck for federated algorithms \cite{alistarh2017qsgd}. This challenge becomes particularly pronounced in scenarios of high network latency, limited communication bandwidth, and when the communicated models are high dimensional. To alleviate the communication bottleneck, there are two common approaches. The first approach is to increase the number of local updates to reduce the number of communication rounds, often at the price of slow convergence speed \cite{mcmahan2017communication}. 
The second approach is to reduce the number of communicated bits by applying compression \cite{khirirat2021flexible, alistarh2017qsgd}. 
Compression can be \emph{sparsification} (which keeps a few important vector elements) and/or \emph{quantization} (which maps each vector element with infinite values into a smaller set of finite values). To further improve solution accuracy of algorithms using compression while saving communicated bits, error feedback mechanisms \cite{khirirat2020compressed} 
and their variants, e.g. EF21 \cite{richtarik2021ef21}, 
have been extensively studied.  The benefits of utilizing these approaches in federated learning have been explored by several works, e.g. \cite{khirirat2022eco,basu2019qsparse}. 
Unlike these prior works, our framework can establish the convergence of error-feedback federated algorithms without data similarity assumptions. 
Our results also apply for stochastic, non-convex optimization unlike \cite{khirirat2022eco}, %mitra2021linear,
and do not assume bounded gradient-norm conditions unlike \cite{basu2019qsparse}.

\subsection{Step Size Schedules for Stochastic Optimization}
\noindent Tuning step sizes is crucial to optimize the convergence performance of stochastic optimization algorithms. 
Using  fixed step sizes for stochastic optimization algorithms guarantees the convergence towards the solution with the residual error, \cite{bertsekas2011incremental,needell2014stochastic}. 
To ensure the convergence of these algorithms towards the exact optimal solution, two common approaches are to use diminishing step sizes \cite{10026503,nguyen2018sgd}, 
or step-decay step sizes \cite{ge2019step,wang2021convergence}. 
More recently, several works \cite{schaipp2023stochastic,loizou2021stochastic} 
have proposed strategies for adjusting step sizes automatically to maximize the performance.  
However, theoretical convergence behaviors under different step size schedules are underexplored for federated learning.
In this work, we unify the convergence of popular federated algorithms without data similarity assumptions when the step sizes are fixed, diminishing, and step-decay.

\section{Main Convergence Theorems}\label{sec:DescentIneq}

\noindent We now proceed to develop our novel sequence results that will serve as the foundation for our proofs for federated learning algorithms without the data-similarity assumption.\footnote{It is worth noting that the general nature of these results renders them potentially valuable for investigating the convergence rates of various algorithms beyond FL.} In particular, we consider  general non-negative sequences  $V_k$ and $W_k$ that satisfy the inequality: 
\begin{align}\label{eqn:mainIneq}
    V_{k+1} \leq (1+ b_1 \gamma_k^2 )V_k - b_2 \gamma_k W_k + b_3 \gamma_k^2,~~ \forall k \geq 0
\end{align}
where $b_1,b_2,b_3$ are non-negative constants and $\gamma_k$ are positive step sizes.  

The system in \eqref{eqn:mainIneq} has been studied by \cite{robbins1971convergence}. 
They prove the almost sure convergence of $\mathop{\min}_{0\leq k \leq K-1} W_k$ when using appropriately chosen diminishing step sizes $\gamma_k$. However, to the best of our knowledge, non-asymptotic results for this system remain unexplored. We aim to fill this gap by presenting non-asymptotic results for various step size selections. Our next three theorems establish the convergence of sequences satisfying the inequality \eqref{eqn:mainIneq} when step sizes are fixed, diminishing, and step-decay. 
\begin{theorem}[Fixed step sizes]\label{thm:fixed}
 Consider the system \eqref{eqn:mainIneq}.
   If $\gamma_k = \gamma = c/\sqrt{K}$ for $c>0$ and $K\in\mathbb{N}$, then 
    \begin{align*}
\mathop{\min}_{0\leq k \leq K-1} W_k
&  \leq \frac{1}{\sqrt{K}} \left( \frac{\exp(b_1 c^2) V_0}{b_2c}  + \frac{b_3c}{b_2} \right). 
\end{align*}
\end{theorem}
%\begin{proof}
%See Appendix~ ...    
%\end{proof}
\begin{theorem}[Diminishing step sizes]\label{thm:diminishing}
	Consider the system \eqref{eqn:mainIneq}.
	If $\gamma_k = c/(k+1)^{\nu}$ for $c>0$, $\nu \in (1/2,1)$ and $k\in\mathbb{N}$, then
	\begin{align*}
		\mathop{\min}_{0\leq k \leq K-1} W_k
		&\leq \frac{1}{K^{1-\nu}}\left( \frac{V_0}{b_2}+ \frac{b_3}{b_2} \frac{2\nu c^2}{2\nu-1} \right) \frac{\exp\left( b_1 \frac{2\nu c^2}{2\nu-1}  \right)}{ c}.
	\end{align*}
\end{theorem}
%\begin{proof}
%See Appendix~ ...
%\end{proof}

\begin{theorem}[Step-decay step sizes]\label{thm:stepdecay}
	Consider the system \eqref{eqn:mainIneq}. 
	Let $K=MT$ for any  $M \geq 1$.
	If $0 \leq V_k \leq R$ for some positive constant $R$, and $\gamma_k =  \gamma_0 / \alpha^{\lfloor k/T \rfloor}$ for $\alpha>1$ and $T = 2K/\log_\alpha K$,  then 
	\begin{align*}
		\mathop{\min}_{0\leq k \leq K-1} W_k
		\leq   \frac{1}{b_2 \gamma_0}\frac{R}{\sqrt{K}} + C\frac{  B}{ \gamma_0} \frac{\log_{\alpha}(K)}{2\sqrt{K}},
	\end{align*}	
	where $B = \exp \left( 2b_1 \gamma_0^2 \frac{1}{\min(\log_{\alpha} 2, 1)} \right)$ and $C= (R+b_3/b_1)/b_2$. 
\end{theorem}
\begin{proof}
 The proofs of Theorem \ref{thm:fixed}, \ref{thm:diminishing}, and \ref{thm:stepdecay} can be found in the supplementary material.     
\end{proof}
Theorem~\ref{thm:fixed},~\ref{thm:diminishing} and~\ref{thm:stepdecay} establish the convergence rates of $\mathop{\min}_{0\leq k \leq K-1} W_k$ towards zero for  three different step sizes. In particular, we obtain the rates $\mathcal{O}(1/K^{1/2})$, $\mathcal{O}(1/K^{1-\nu})$, and $\mathcal{O}(\log_\alpha K/\sqrt{K})$, respectively,   for fixed step sizes, diminishing step sizes, and step-decay step sizes. 
{\color{black}
Note that in the federated learning application, we have in \eqref{eqn:mainIneq}  $V_k= \mathbf{E}[f(x^k) - f^{\inf}]$ and $W_k=\mathbf{E}\| \nabla f(x^k) \|^2$, where $f(\cdot)$ is a loss function and $f^{\inf}$ is its lower bound.
Our theorems will then establish convergence towards a stationary point.
}

 By rearranging the bounds in the theorems, they can also be used to establish iteration complexity. In particular, to reach $\epsilon$-accurate solution (i.e., such that $\mathop{\min}_{0\leq k \leq K-1} W_k\leq \epsilon$) by Theorem~\ref{thm:fixed} we need in the worst case
 \begin{align*}
	K = \frac{1}{\epsilon^2} \left( \frac{\exp(b_1 c^2) V_0}{b_2c}  + \frac{b_3c}{b_2} \right)^2 \quad \text{iterations}.
\end{align*}
 Therefore, we can establish that the iteration complexity with fixed step sizes is on the order of $\mathcal{O}(1/\epsilon^2)$.
Similarly, % if we use a diminishing step size 
we can establish that the iteration complexity with diminishing step sizes is on the order of $\mathcal{O}(1/\epsilon^{1/(1-\nu)})$ for $\nu \in (1/2,1)$. In particular, by rearranging the bound in Theorem~\ref{thm:diminishing} we observe that to reach an $\epsilon$-accurate solution requires in the worst case 
\begin{align*}
	\frac{1}{\epsilon^{\frac{1}{1-\nu}}}\left[ \left( \frac{V_0}{b_2}+ \frac{b_3}{b_2} \frac{2\nu c^2}{2\nu-1} \right) \frac{\exp\left( b_1 \frac{2\nu c^2}{2\nu-1}  \right)}{ c} \right]^{\frac{1}{1-\nu}} \ \text{iterations}.
\end{align*}

\section{Applications in Federated Learning}\label{sec:applications}

\noindent In this section, we demonstrate how our novel sequence results (Theorem~\ref{thm:fixed}, \ref{thm:diminishing}, and \ref{thm:stepdecay}) can be effectively applied to establish convergence guarantees for federated learning under broader assumptions than the previous literature considered. 
Notably, our approach does not necessitate data similarity assumptions and incorporates different step size schedules. The central idea underlying all proofs is to initially derive the worst-case convergence bound in the form of \eqref{eqn:mainIneq} for each algorithm. Subsequently, we leverage Theorem \ref{thm:fixed}, \ref{thm:diminishing}, and \ref{thm:stepdecay} to determine the convergence rate for the algorithm. This methodology enables us to achieve more general and robust convergence results for federated learning.

We consider the typical federated learning set-up where  $n$ workers wish to collaboratively solve a  finite-sum minimization problem on the form
\begin{align}\label{eq:Problem}
	\mathop{\text{minimize}}\limits_{x\in\mathbb{R}^d} \quad f(x) := \frac{1}{n}\sum_{i=1}^n f_i(x),
\end{align}	
where $x\in\mathbb{R}^d$ is a vector storing model parameters, and each worker accesses a single private objective function  $f_i:\mathbb{R}^d\rightarrow\mathbb{R}$ which is often on the form: 
\begin{align}\label{eq:f_i_exp}
	f_i(x) := \mathbb{E}_{\xi_i\sim\mathcal{D}_i} F_i(x;\xi_i). 
\end{align}	
Here, $\xi_i$ is a random variable vector sampled from the distribution of data points $\mathcal{D}_i$  stored privately at worker $i$.

To facilitate our analysis, we impose two standard assumptions on the objective functions of Problem~\eqref{eq:Problem}. 
The first assumption is the Lipschitz continuity of $\nabla f_i(x)$.
\begin{assumption}\label{assum:L_Lipschitz}
Each local function $f_i:\mathbb{R}^d\rightarrow\mathbb{R}$ is bounded from below by an infimum $f^{\inf}_i \in\mathbb{R}$, is differentiable, and has $L$-Lipschitz continuous gradient, i.e. for all $x,y\in\mathbb{R}^d$, 
\begin{align}\label{eq:L_Lipschitz}
\| \nabla f_i(x) - \nabla f_i(y)\| \leq L \| x-y\|.
\end{align}	
\end{assumption}	
From Assumption \ref{assum:L_Lipschitz} and by Cauchy-Schwartz's inequality, the whole objective function $f(x)$ in \eqref{eq:Problem} also has $L$-Lipschitz continuous gradient. 
Furthermore, the following inequalities  are direct consequences from Assumption~\ref{assum:L_Lipschitz} \cite{li2019convergence, khaled2022better}: 
\begin{align}
	&f_i(y)  \leq f_i(x) + \langle \nabla f_i(x), y-x \rangle + ({L}/{2})\| y-x\|^2,  \label{eq:L_Lipschitz_ineq} 
\end{align}
and
% \intertext{and} 
\begin{align}
	&\| \nabla f_i(x)\|^2  \leq  2L[f_i(x) - f_i^{\inf}]. \label{eq:L_Lipschitz_trick}
\end{align}		
%\end{lemma}	
%
The second assumption we impose is the bounded variance of a stochastic local gradient $\nabla F_i(x;\xi_i)$ with respect to a full local gradient $\nabla f_i(x)$. 
\begin{assumption}\label{assum:variance_stoc}
	The variance of the stochastic gradient in each  node is bounded, i.e. for all $x\in\mathbb{R}^d$
	\begin{align}\label{eq:variance_stoc}
		\mathbb{E}_{\xi_i} \| \nabla F_i(x;\xi_i)  - \nabla f_i(x) \|^2 \leq \sigma^2.
	\end{align}	
\end{assumption}	
Assumption~\ref{assum:variance_stoc} is standard to analyze the convergence of optimization methods using stochastic gradients \cite{alistarh2017qsgd,feyzmahdavian2016asynchronous}. 
In addition, since $\nabla f_i(x) = \mathbb{E}_{\xi_i}\nabla F_i(x;\xi_i)$, this assumption implies that
\begin{align*}
\mathbb{E}_{\xi_i} \| \nabla F_i(x;\xi_i)  \|^2 \leq \sigma^2 + \| \nabla f_i(x) \|^2.
\end{align*}
Assumption~\ref{assum:variance_stoc} is thus more general than the bounded second moment assumption, i.e. $\mathbb{E}_{\xi}\| \nabla F_i(x;\xi_i) \|^2 \leq \sigma^2$, which is used to derive the convergence of federated learning algorithms, see, e.g., \cite{basu2019qsparse,xu2021step}. 
{
\color{black} Also, this assumption is more relaxed than the bounded $4^\text{th}$-moment of the variance, i.e. $\mathbb{E}_{\xi}\| \nabla F_i(x;\xi_i) - \nabla f_i(x) \|^4 \leq \sigma^4$ studied in \cite{glasgow2022sharp}, because $\mathbf{E}\| \nabla F_i(x;\xi_i) - \nabla f_i(x) \|^2 \leq \sqrt{\mathbf{E}\| \nabla F_i(x;\xi_i) - \nabla f_i(x)\|^4}$.
}

Next, we derive the convergence bounds in \eqref{eqn:mainIneq} and establish rate-convergence results for full-precision and communication-efficient federated learning algorithms for Problem~\eqref{eq:Problem} without data similarity assumptions.

\subsection{Full-precision Federated Learning Algorithms}

\begin{figure}[t]
	\centering
\includegraphics[width=1.0\columnwidth]{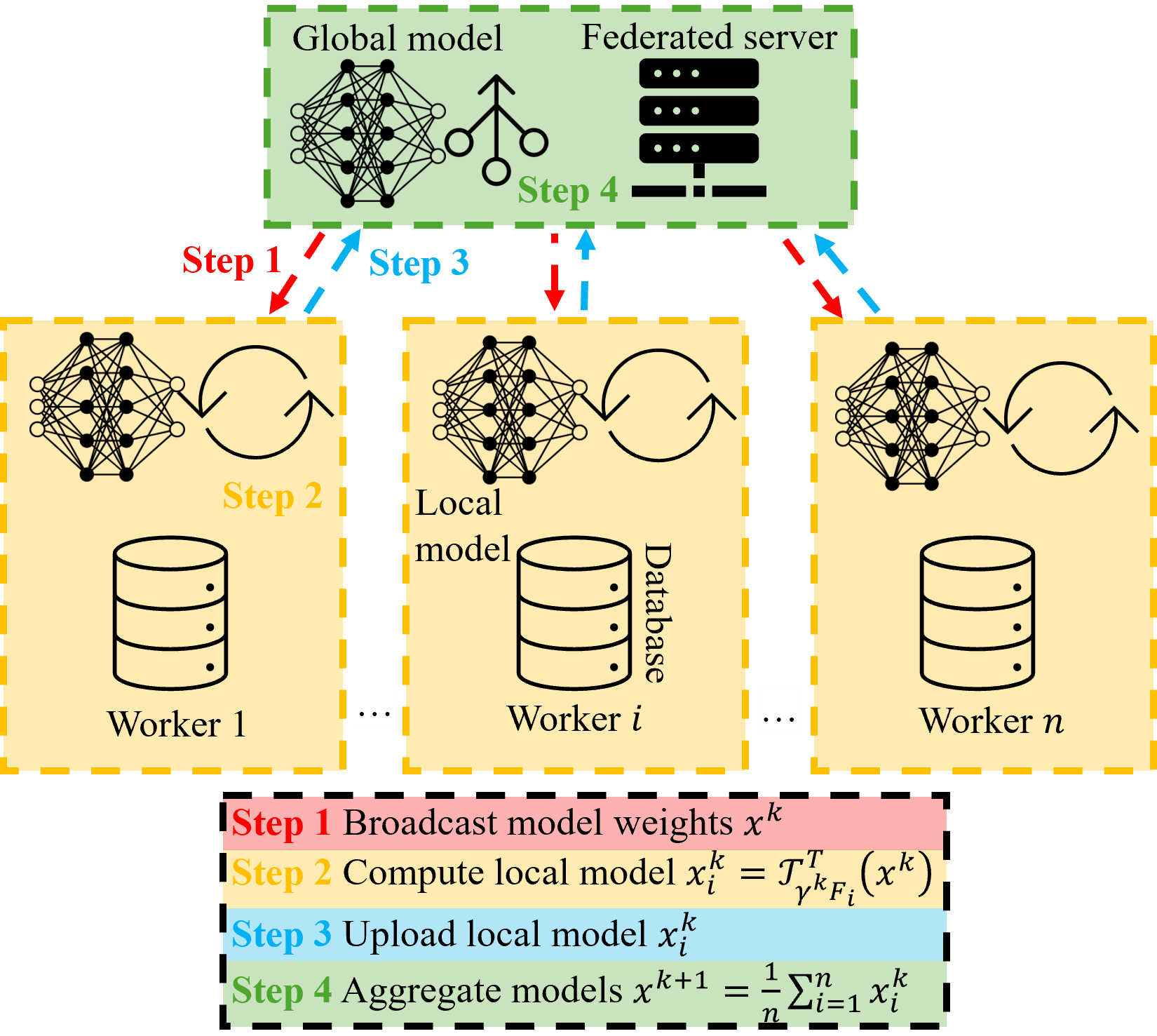} 
	\caption{\textcolor{black}{Visual workflow of the full-precision federated learning algorithms.}}
	\label{fl_vig}
\end{figure}

\noindent We start by considering full-precision federated learning algorithms to solve the problem in~\eqref{eq:Problem} where Assumptions~\ref{assum:L_Lipschitz} and~\ref{assum:variance_stoc} hold. 
 In these algorithms, the server updates the global model parameters based on the local model parameters from the workers.  
Given the initial point $x^0\in\mathbb{R}^d$ and the step size schedule $\gamma^k>0$, these algorithms proceed in $K$ communication rounds. 
In each communication round $k\in\{0,1,\ldots,K-1\}$, every worker updates its local model parameters $x_i^k$ by performing $T$ local fixed-point iterations according to:
\begin{align*}
	x_i^k =  \mathcal{T}_{\gamma^k F_i}^T(x^k),
\end{align*}
where $\mathcal{T}_{\gamma F}:\mathbb{R}^d\rightarrow\mathbb{R}^d$ is a fixed-point operator with a private function $F(x)$ and a positive step size $\gamma$. 
Then, the server updates the global models $x^{k+1}$  by averaging the local models from every worker:
\begin{align*}
	 x^{k+1} = \frac{1}{n}\sum_{i=1}^n x_i^{k}.
\end{align*}
\textcolor{black}{
The full-precision federated learning algorithms are described formally in Algorithm~\ref{alg:fp_fl} and its visual workflow is shown in Figure \ref{fl_vig}.}

\begin{algorithm}[t]
	\caption{ {Full-precision Federated Learning Algorithms} }\label{alg:fp_fl}
	\begin{algorithmic}
		\State \textbf{Input:} The number of iterations $K,T$, the step size $\gamma^k > 0$, and the initial point $x^0\in\mathbb{R}^d$.
		\For{$k=0,1,\ldots,K-1$}
		\State The server broadcasts $x^k$ to every worker node 
		\For{every worker $i=1,\ldots,n$}
		\State Compute $x_i^k = \mathcal{T}_{\gamma^k F_i}^T(x^k)$
		\State Send $x_i^{k}$ to the server 	
		\EndFor 
		\State The server updates $x^{k+1} = \frac{1}{n}\sum_{i=1}^n x_i^{k}$	
		\EndFor
	\end{algorithmic}
\end{algorithm}

Now, we derive \eqref{eqn:mainIneq} and convergence results without data similarity assumptions for two popular full-precision federated algorithms: FedAvg and FedProx.

\subsubsection{FedAvg} FedAvg is the special case of Algorithm~\ref{alg:fp_fl} when $\mathcal{T}_{\gamma F}(x) =x-\gamma \nabla F(x)$. In this case, we obtain the following form of %the system inequality in 
\eqref{eqn:mainIneq}.

\begin{proposition}[FedAvg]\label{prop:fedavg}
Consider Algorithm~\ref{alg:fp_fl} with $\mathcal{T}_{\gamma F}(x) =x-\gamma \nabla F(x)$ for	the problem in~\eqref{eq:Problem} where Assumptions~\ref{assum:L_Lipschitz} and~\ref{assum:variance_stoc} hold.  The iterates $\{x^k\}$ generated by this algorithm with $\gamma^k = \alpha^k/T$  and $\alpha^k \leq 1/(\sqrt{6}L)$ satisfies \eqref{eqn:mainIneq},
	%\begin{eqnarray*}
	%	V_{k+1}
	%	& \leq & (1+\sqrt{6}(\alpha^k)^2 L^2 T) V_k - \frac{\alpha^k}{2} W_k + (\alpha^k)^2 e,
	%\end{eqnarray*}		
	where 
	\begin{align*}
		V_k &= \mathbf{E}[f(x^k) - f^{\inf}], \quad   W_k=\mathbf{E}\| \nabla f(x^k) \|^2\\ 
		\gamma_k&=\alpha^k, \quad  b_1 = \sqrt{6}L^2 T,  \quad b_2=1/2, \quad \text{and} \\ 
		\quad b_3 &= \sqrt{6}L^2 T \Delta^{\inf}+ L [1 + (3/\sqrt{6}) T] \sigma^2.
	\end{align*}	
	%$V_k=  \mathbf{E}[f(x^k) - f^{\inf}]$, $W_k=\mathbf{E}\| \nabla f(x^k) \|^2$, $\gamma_k=\alpha^k$,  $b_1=\sqrt{6}L^2 T$, $b_2=1/2$, $b_3=\sqrt{6}L^2 T \Delta^{\inf}+ (3/\sqrt{6}) L T \sigma^2 + L  \sigma^2$, and 
	Here, $\Delta^{\inf}=(1/n)\sum_{i=1}^n [f^{\inf} - f_i^{\inf}] \geq 0$.  
\end{proposition}

By Proposition~\ref{prop:fedavg}  and Theorem~\ref{thm:fixed}, FedAvg  attains the $\mathcal{O}(1/K^{1/2})$ convergence for non-convex problems when $\gamma^k=\alpha^k/T$ and  $\alpha^k = (\sqrt{6}L)^{-1}/\sqrt{K}$.
{
\color{black}
Our result for FedAvg with fixed step sizes does not require data similarity assumptions, provides a  faster rate than \cite{glasgow2022sharp}, and also does not require additional  assumptions that restrict problem classes, e.g. the PL condition 	\cite{haddadpour2019local}, the bounded gradient-norm condition \cite{yu2019parallel},  the condition that all eigenvalues of the mixing matrix must be less than $1$ 	\cite{wang2021cooperative}, and the $3^\text{rd}$-order smoothness on a function (or the Lipschitz continuity on the Hessian )\cite{glasgow2022sharp}. }
These comparisons between our convergence theorems and existing works were summarized in Table~\ref{tab:fedavg}.
Furthermore, FedAvg converges at the $\mathcal{O}(1/K^{1-\nu})$ rate when $\alpha^k = (\sqrt{6}L)^{-1}/(k+1)^{\nu}$ for $\nu\in(1/2,1)$ from Theorem~\ref{thm:diminishing}, and at the $\mathcal{O}(1/\log_{\alpha}(K))$ rate when $\alpha^k = (\sqrt{6}L)^{-1}/\alpha^{\lfloor k/T \rfloor}$ for $\alpha > 1$ and $T=2K/\log_\alpha K$  from Theorem~\ref{thm:stepdecay}.

\begin{table}
	\centering 
        \small
	\begin{tabular}{c|c|c|c}  \hline 
		Ref. & Rate     & Data similarity  & Extra assumption \\ \hline 
		\cite{yu2019parallel} & $\mathcal{O}\left( \frac{1}{\sqrt{K}} \right)$& Yes  &  $\mathbb{E}_{\xi_i}\| \nabla F_i(x;\xi_i)\|\leq \sigma^2$ \\  
		\cite{wang2021cooperative} & $\mathcal{O}\left( \frac{1}{\sqrt{K}}\right)$& No  &  $\lambda_i(W)<1$ for all $i$ \\  
		\cite{haddadpour2019local} & $\mathcal{O}\left( \frac{1}{K} \right)$ & No   & PL  \\
    {\color{black}  \cite{glasgow2022sharp}} & {\color{black}$\mathcal{O}\left(\frac{1}{K^{2/5}}\right)$} & {\color{black}No} &  {\color{black}$3^\text{th}$-order smoothness} \\
		{\bf Ours} & $\mathcal{O}\left( \frac{1}{\sqrt{K}} \right)$& {\bf No}   & \bf No \\ \hline 
	\end{tabular}	
	\caption{Comparisons of convergence results for FedAvg on non-convex problems in non-iid data settings. Here, the result of ours derive from Proposition~\ref{prop:fedavg} and Theorem~\ref{thm:fixed}, and $W$ is the mixing matrix.}
	\label{tab:fedavg}
\end{table}	

\subsubsection{{FedProx}} FedProx is the special case of Algorithm~\ref{alg:fp_fl} when $\mathcal{T}_{\gamma F}(x)=\textbf{prox}_{\gamma F}(x)$ and $T=1$. Similarly, as above, we obtain the following form of %the system inequality in 
\eqref{eqn:mainIneq}.
\begin{proposition}[FedProx]\label{prop:FedProx}
Consider 	Algorithm~\ref{alg:fp_fl} with $\mathcal{T}_{\gamma F}(x)=\textrm{\bf prox}_{\gamma F}(x)$ and $T=1$ for the problem in~\eqref{eq:Problem} where Assumptions~\ref{assum:L_Lipschitz} and~\ref{assum:variance_stoc} hold. Then, the iterates $\{x^k\}$ generated by this algorithm with $\gamma^k \leq 1/(\sqrt{6} L)$ satisfy
\eqref{eqn:mainIneq}, where 
\begin{align*}
V_k & = \mathbf{E}[f(x^{k}) - f^{\inf}], \quad W_k= \mathbf{E}\| \nabla f(x^k) \|^2 \\
b_1& =\sqrt{6}L^2, \quad b_2=1/2, \quad \text{and} \\
\quad b_3  &= \sqrt{6}L^2\Delta^{\inf} + L(1+3/\sqrt{6})\sigma^2.
\end{align*}
Here, $\Delta^{\inf}=(1/n)\sum_{i=1}^n [f^{\inf} - f_i^{\inf}] \geq 0$.  
%\begin{eqnarray*}
%V_{k+1}
%	& \leq & (1+\sqrt{6}L^2(\gamma^k)^2)V_k  - \frac{\gamma^k}{2} W_k + (\gamma^k)^2 e, 
%\end{eqnarray*}	
%where $V_k:= \mathbf{E}[f(x^{k}) - f^{\inf}]$, $W_k:= \mathbf{E}\| \nabla f(x^k) \|^2$, $e:= \sqrt{6}L^2\Delta^{\inf} + L(1+3/\sqrt{6})\sigma^2$, and $\Delta^{\inf}:=(1/n)\sum_{i=1}^n [f^{\inf} - f_i^{\inf}]>0$.  
\end{proposition}
By  Proposition~\ref{prop:FedProx} and Theorem~\ref{thm:fixed}, FedProx achieves the $\mathcal{O}(1/K^{1/2})$ convergence for  non-convex problems.
This result with fixed step sizes does not assume the data similarity assumption and additional restrictive assumptions on objective functions by prior works in \cite{li2020federated,nguyen2020fast,yuan2022convergence}.  

Unlike Theorem 4 of  \cite{li2020federated} and Theorem 1 of \cite{nguyen2020fast}, our result does not require the data similarity assumption on each local function $f_i(x)$ with respect to the whole function $f(x)$, i.e. where there exists a data-similarity parameter $B \geq 0$ such that 
\begin{align*}
	\mathbf{E}\| \nabla f_i(x)\|^2 \leq B^2 \| \nabla f(x)\|^2, \quad \forall x\in\mathbb{R}^d.
\end{align*}	
%
%Moreover, we guarantee convergence under more relaxed step size selections that do not depend on data similarity $B$. 
%
%On the one hand, 
Theorem 4 in  \cite{li2020federated} ensures FedProx convergence only when the fixed step size $\gamma>0$ is chosen based on data similarity $B$ and other parameters $\mu,\bar\mu,K>0$ such that 
\begin{align*}
    \rho = {1}/{\mu} - {\gamma B}/{\mu} - (1+\gamma)C_1 B - (1+\gamma^2) C_2 B^2 >0,
\end{align*}
where $C_1 = \sqrt{2}/(\bar\mu \sqrt{K}) + L/(\bar\mu \mu)$ and $C_2=L/(2\bar\mu)^2 + L(2\sqrt{2K}+2)/(K\bar\mu^2)$.
When the data-similarity $B$ is too large, then there is no step size $\gamma>0$ fulfilling this condition.
On the other hand, we show that FedProx converges for any fixed step size $\gamma^k=\gamma$ satisfying 
\(
	0 <\gamma \leq {(\sqrt{6}L)^{-1}}/{\sqrt{K}}. 
 \)
Thus, we guarantee convergence under more relaxed step size selections that do not depend on data similarity $B$.

To the best of our knowledge, the only convergence-rate result for FedProx that does not assume data similarity is~\cite{yuan2022convergence}. However, our results are more general as we do not impose the assumption of Lipschitz continuity on each local function, a requirement made in~\cite{yuan2022convergence}. This means, for example, that their results do not even cover quadratic loss functions. Moreover,~\cite{yuan2022convergence} only consider fixed step sizes, whereas our results cover both diminishing step sizes (by Theorem~\ref{thm:diminishing}) and step-decay step sizes (by Theorem~\ref{thm:stepdecay}).

\subsection{Error-feedback Federated Learning Algorithms}

\begin{figure}[t]
	\centering
\includegraphics[width=1.0\columnwidth]{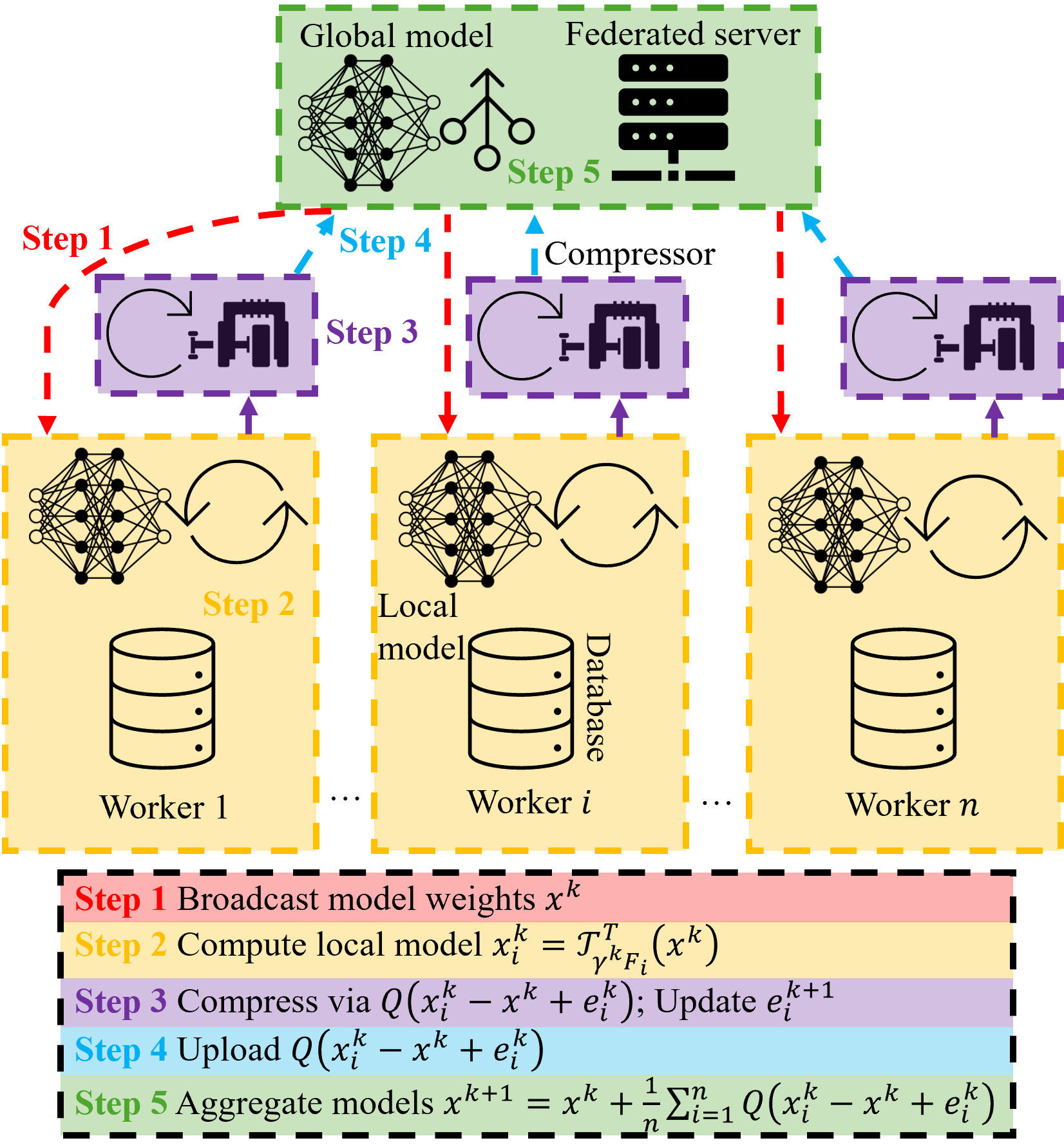} 
	\caption{\textcolor{black}{Visual workflow of the error-feedback federated learning algorithms.}}
	\label{ef_fl_vig}
\end{figure}

\noindent To improve communication efficiency while maintaining the strong convergence performance of full-precision federated learning algorithms, we turn our attention to error-feedback federated learning algorithms.
These algorithms contain two communication-saving approaches: (1) local updating and (2) error-compensated message passing. 
In each communication round $k\in\{0,1,\ldots,K-1\}$ of these algorithms, the server broadcasts the current global model $x^k$ to all workers, and each worker performs $T$ local fixed-point updates. 
In particular, worker $i$ updates its local model via: 
    \begin{align*}
        x_i^k = \mathcal{T}^T_{\gamma^k F_i}(x^k).
    \end{align*}
After $T$ local updates, each worker uploads a compressed message vector $Q(x_i^k-x^k+e^k_i)$ to the server and updates the compression error $e^{k+1}_i$ according to: 
    \begin{align*}
        e_i^{k+1} = x_i^k-x^k+e^k_i - Q(x_i^k-x^k+e^k_i).
    \end{align*}
Then, the server receives compressed message vectors and computes the next global model via: 
    \begin{align*}
        x^{k+1} = x^k + ({1}/{n})\sum_{i=1}^n Q(x_i^{k} - x^k + e_i^k).
    \end{align*}
\textcolor{black}{These algorithms are formally summarized in Algorithm~\ref{alg:EC_FL}, and its visual workflow is shown in Figure \ref{ef_fl_vig}.}  
Note that Algorithm~\ref{alg:EC_FL} recovers FedPAQ \cite{reisizadeh2020fedpaq} when we let  $e_i^k=0$ for all $i,k$ and $\mathcal{T}_{\gamma F}(x) =x-\gamma \nabla F(x)$, and becomes Algorithm~\ref{alg:fp_fl} when we set $e_i^k=0$ for all $i,k$ and $Q(x)=x$ . 

To analyze these error-feedback algorithms, we impose Assumptions~\ref{assum:L_Lipschitz} and~\ref{assum:variance_stoc} and also consider a contractive compressor that covers several compressors of interest.
\begin{assumption}[Contractive compressor]\label{assum:contractive_comp}
 The compressor $Q:\mathbb{R}^d\rightarrow\mathbb{R}^d$ is contractive with a scalar $\alpha \in (0,1]$, i.e.
 \begin{equation}\label{eq:contractive_comp}
     \| Q(v) - v \|^2 \leq (1-\alpha)\| v \|^2~~~\text{ for all } v\in\mathbb{R}^d. 
 \end{equation}
\end{assumption}
From Assumption~\ref{assum:contractive_comp},  $\alpha$ implies the precision of a contractive compressor. 
For extreme cases, $Q(v)$ becomes close to $v$ as $\alpha$ is close to one.
%
%Furthermore, Assumption~\ref{assum:contractive_comp} is satisfied by many compressors. 
%
Contractive compressors cover the ternary quantizer \cite{khirirat2018gradient} with $\alpha=1/d$,  the scaled sign quantizer \cite{karimireddy2019error} with $\alpha=1/d$, and 
the Top-$K$ sparisifier \cite{khirirat2018gradient,richtarik2021ef21} %,karimireddy2019error
with $\alpha=K/d$.

\begin{algorithm}[t]
	\caption{Error-feedback Federated Learning Algorithms}\label{alg:EC_FL}
	\begin{algorithmic}
		\State \textbf{Input:} The number of iterations $K,T$, the step size $\gamma^k > 0$, the initial point $x^0\in\mathbb{R}^d$, and $e_i^0=0$ for all $i$.
		\For{$k=0,1,\ldots,K-1$}
		\State The server broadcasts $x^k$ to every worker node 
		\For{every worker $i=1,\ldots,n$}
		\State Compute $x_i^k = \mathcal{T}_{\gamma^k F_i}^T(x^k)$
		\State Send $Q(x_i^{k} - x^k + e_i^k)$ to the server
		\State Update $e_i^{k+1} =x_i^{k} - x^k + e_i^k - Q(x_i^{k} - x^k + e_i^k)$ 	
		\EndFor 
		\State The server: $x^{k+1} = x^k + \frac{1}{n}\sum_{i=1}^n Q(x_i^{k} - x^k + e_i^k)$	
		\EndFor
	\end{algorithmic}
\end{algorithm}

Now, we derive the convergence bound in \eqref{eqn:mainIneq}  for two error-feedback federated algorithms:  error-feedback FedAvg and error-feedback FedProx.

\subsubsection{Error-feedback {FedAvg}}
Error-feedback FedAvg is a special case of Algorithm~\ref{alg:EC_FL} with $\mathcal{T}_{\gamma F}(x) =x-\gamma \nabla F(x)$.
The next result shows this algorithm follows  \eqref{eqn:mainIneq}. 
\begin{proposition}[Error-feedback FedAvg]\label{prop:EF_fedavg}
	Consider Algorithm~\ref{alg:EC_FL} with $\mathcal{T}_{\gamma F}(x) =x-\gamma \nabla F(x)$ for	the problem in~\eqref{eq:Problem} where Assumptions~\ref{assum:L_Lipschitz},~\ref{assum:variance_stoc} and~\ref{assum:contractive_comp} hold.  The iterates $\{x^k\}$ generated by this algorithm with $\gamma^k = \alpha^k/T$  and  $\alpha^k \leq \hat\alpha := \frac{1}{L}\min\left( \frac{1}{6} , \sqrt{\frac{3\alpha}{64(1-\alpha)(1+2/\alpha)}}  \right)$ satisfies \eqref{eqn:mainIneq}, where 
	\begin{align*}
		V_k & = \mathbf{E}[f(z^{k}) - f^{\inf}] +\frac{4 (1+1.5L) L^2 \alpha^k}{\alpha n}  \sum_{i=1}^n \mathbf{E} \| e_i^k \|^2,  \\
         W_k & = \mathbf{E}\| \nabla f(x^k) \|^2, \quad 
        \gamma_k = \alpha^k, \quad b_1 = 2L\tilde C_2, \\
		b_2 & =\frac{1}{4}, \quad \text{and} \quad  b_3  = 2L \tilde C_2 \Delta^{\inf} + \tilde C_3 \sigma^2.
	\end{align*}		
	Here,  $\Delta^{\inf}=(1/n)\sum_{i=1}^n [f^{\inf} - f_i^{\inf}] \geq 0$, $\tilde C_2 =  \frac{16(1-\alpha)(1+2/\alpha)A}{3} + \frac{3 L }{2}$, $\tilde C_3 =  \frac{14(1-\alpha)(1+2/\alpha) A}{3}  +  \frac{13L}{8}$, and $A = \frac{4 (1+1.5L) L^2}{\alpha} \hat \alpha$.
\end{proposition}

From Proposition~\ref{prop:EF_fedavg} and Theorem~\ref{thm:fixed}, \ref{thm:diminishing}, and \ref{thm:stepdecay}, error-feedback FedAvg enjoys the $\mathcal{O}(1/K^{1/2})$, $\mathcal{O}(1/K^{1-\nu})$ and $\mathcal{O}(\log_\alpha K/\sqrt{K})$ convergence  without data similarity assumptions, respectively, when $\alpha^k$ is fixed, diminishing and step-decay. 
%
%when  $\alpha^k = \alpha = \hat\alpha/\sqrt{K}$, when $\alpha^k = \hat\alpha/(k+1)^\nu$ for $\nu\in(1/2,1)$, and when $\alpha^k = \hat\alpha/ \alpha^{\lfloor k/T \rfloor}$ for $\alpha>1$ and $T = 2K/\log_\alpha K$. 
%
Our $\mathcal{O}(1/K^{1-\nu})$ rate with diminishing step sizes is stronger than the $\mathcal{O}(1/\ln(K))$ rate by \cite[Theorem 2]{basu2019qsparse}.
In addition,   in contrast to \cite[Theorem 1]{basu2019qsparse},  our result with fixed step sizes does not assume that the second moment is bounded, which is more restrictive than Assumption~\ref{assum:variance_stoc}.

{\color{black}
To the best of our knowledge, the only paper investigating error-feedback federated averaging algorithms without data similarity assumptions is \cite{gao2021convergence}. However, the authors do not provide proof for their statements, neither in the main paper nor in the extended version of their paper posted on ArXiv.
In contrast to \cite{gao2021convergence}, our analysis framework can also be applied to derive the convergence of error-feedback FedProx without data similarity assumptions, as shown next.
}

\subsubsection{Error-feedback FedProx} 
Error-feedback FedProx is the special case of  Algorithm~\ref{alg:EC_FL} with $\mathcal{T}_{\gamma F}(x)=\textrm{\bf prox}_{\gamma F}(x)$ and $T=1$, which 
follows~\eqref{eqn:mainIneq}.
\begin{proposition}[Error-feedback FedProx]\label{prop:EF_fedprox}
Consider 	Algorithm~\ref{alg:EC_FL} with $\mathcal{T}_{\gamma F}(x)=\textrm{\bf prox}_{\gamma F}(x)$ and $T=1$ for the problem in~\eqref{eq:Problem} where Assumptions~\ref{assum:L_Lipschitz},~\ref{assum:variance_stoc} and~\ref{assum:contractive_comp} hold.
Then, the iterates $\{x^k\}$ generated by this algorithm with $\gamma^k \leq \gamma := \min\left( \frac{1}{6L}, \frac{1}{2}\sqrt{\frac{\alpha}{C_1}} \right)$  satisfy
	\eqref{eqn:mainIneq}, where 
\begin{align*}
	V_k & = \mathbf{E}[f(z^{k}) - f^{\inf}] +\frac{3L^2 \gamma^k}{\alpha n}\sum_{i=1}^n \mathbf{E} \| e_i^k \|^2, \quad \\ 
 W_k & = \mathbf{E}\| \nabla f(x^k) \|^2, \quad 
 b_1  = 2L\left( \frac{3L}{2} + A C_2 \right), \quad  b_2 =\frac{1}{4}, \quad \\
	b_3 & = 2L\left( \frac{3L}{2} + A C_2 \right) \Delta^{\inf} + \left( \frac{9L}{4}+ A C_3 \right) \sigma^2.
\end{align*}		
Here, $\Delta^{\inf}=(1/n)\sum_{i=1}^n [f^{\inf} - f_i^{\inf}] \geq 0$,
% $W_k= \mathbf{E}\| \nabla f(x^k) \|^2$, $b_1= 2L(3L^2/2 + A C_2)$, $b_2=1/4$, $b_3=2L(3L^2/2 + A C_2) \Delta^{\inf} + (3L^2/4 + 3L/2 + A C_3) \sigma^2$, and $\Delta^{\inf}:=(1/n)\sum_{i=1}^n [f^{\inf} - f_i^{\inf}] \geq 0$.  
$A= 3L^2\gamma/\alpha$, $C_1 = (1-\alpha)(1+2/\alpha)(4+4L^2/3)$, $C_2 =(1-\alpha)(1+2/\alpha)(4+4/3)$ and  $C_3 =(1-\alpha)(1+2/\alpha)(4+2/3)$.
\end{proposition}	

Similarly to error-feedback FedAvg, we apply Proposition~\ref{prop:EF_fedprox}, and Theorem~\ref{thm:fixed}, \ref{thm:diminishing}, and \ref{thm:stepdecay} to establish the  $\mathcal{O}(1/K^{1/2})$, $\mathcal{O}(1/K^{1-\nu})$ and $\mathcal{O}(\log_\alpha K/\sqrt{K})$ convergence without data similarity assumptions for error-feedback FedProx, respectively, using fixed, diminishing and step-decay step sizes $\gamma^k$.

\section{Numerical Experiments} \label{Numerical_Experiments}

\noindent {\color{black} We finally evaluated the performance of four different federated learning algorithms using three step size strategies to train deep neural network models over two distinct datasets: MNIST \cite{lecun1998mnist} and FashionMNIST \cite{xiao2017fashion}, under various data similarity conditions.
Although it is feasible to explore additional datasets, we chose to focus on these two while considering different data distributions described in the following section to better highlight our key findings. 
}  
Both datasets contain $60000$ training images and $10000$ test images. 
Each $28\times 28$ grayscale image of the MNIST and FashionMNIST datasets is, respectively, one out of ten handwritten digits and one out of ten distinct fashion items. 
In particular, we implemented FedAvg, FedProx, error-feedback FedAvg, and error-feedback FedProx to solve the convolutional neural network (CNN) model using PyTorch \cite{imambi2021pytorch}. 
This architecture contains a CNN with two 5x5 convolution layers, in which the first layer and second layer have 20 channels, and 50 channels, respectively, and each is followed by a 2x2 max pooling and ReLU activation function.
Then, a fully connected layer with 500 units, a ReLU activation function, and a final softmax output layer formed our selected architecture.
The total number of trainable parameters of this model is thus 431,080. 
All the numerical experiments were implemented in Python 3.8.6 and conducted on a computing server equipped with an NVIDIA Tesla T4 GPU with 16GB RAM. 
All source codes required for conducting and analyzing the experiments are made available online\footnote{\url{https://github.com/AliBeikmohammadi/FedAlgo_WO_DataSim/}}.

\begin{figure}[t]
	\centering
	\includegraphics[width=0.9\columnwidth]{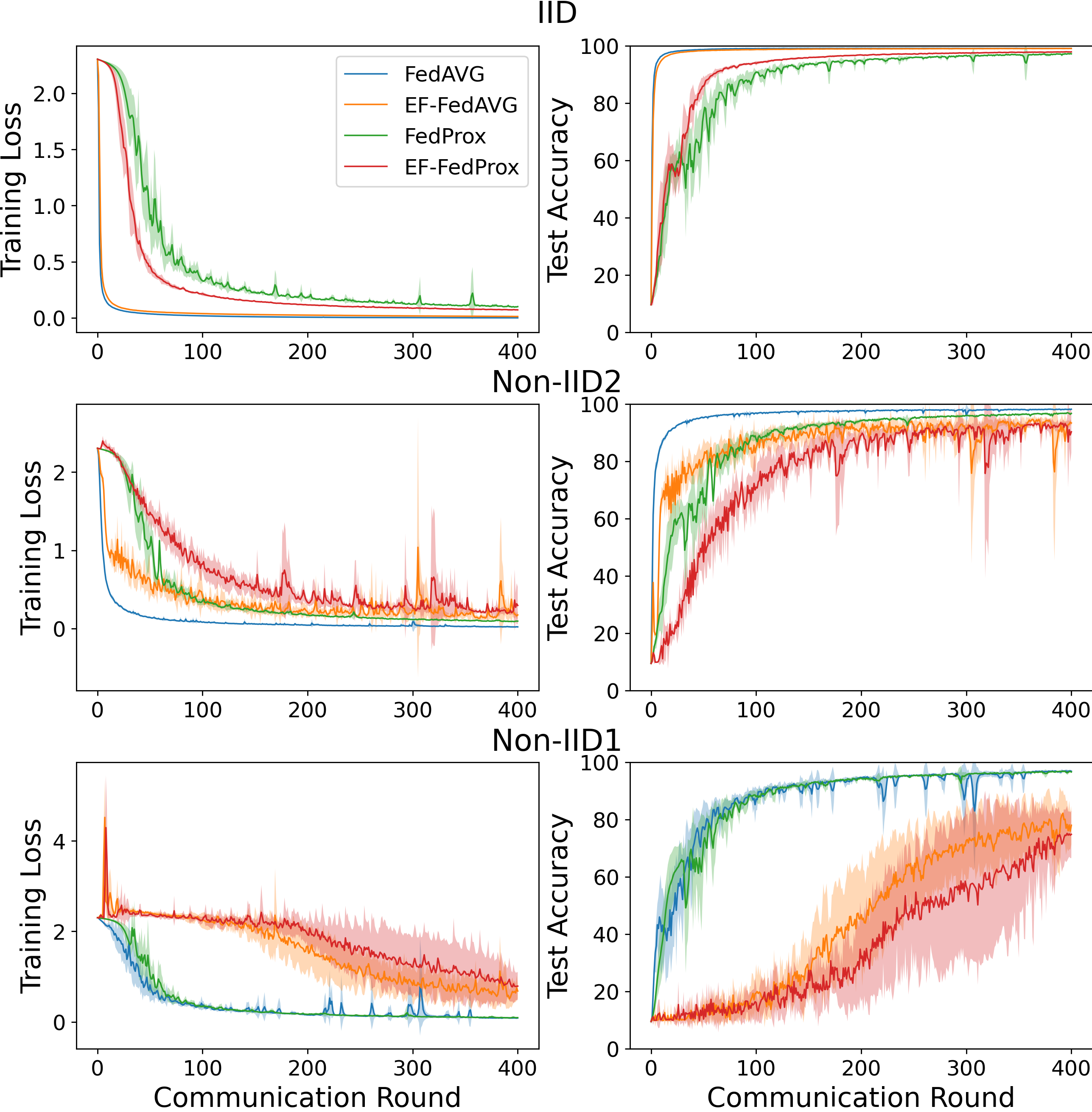} 
	\caption{Performance of FedAvg, error-feedback FedAvg, FedProx, and error-feedback FedProx with the fixed step size in (left plots -) training loss and (right plots -) test accuracy on MNIST dataset considering three different partitioned data among the workers.}
	\label{mnist-fix}
\end{figure}

\begin{figure}[t]
	\centering
	\includegraphics[width=0.9\columnwidth]{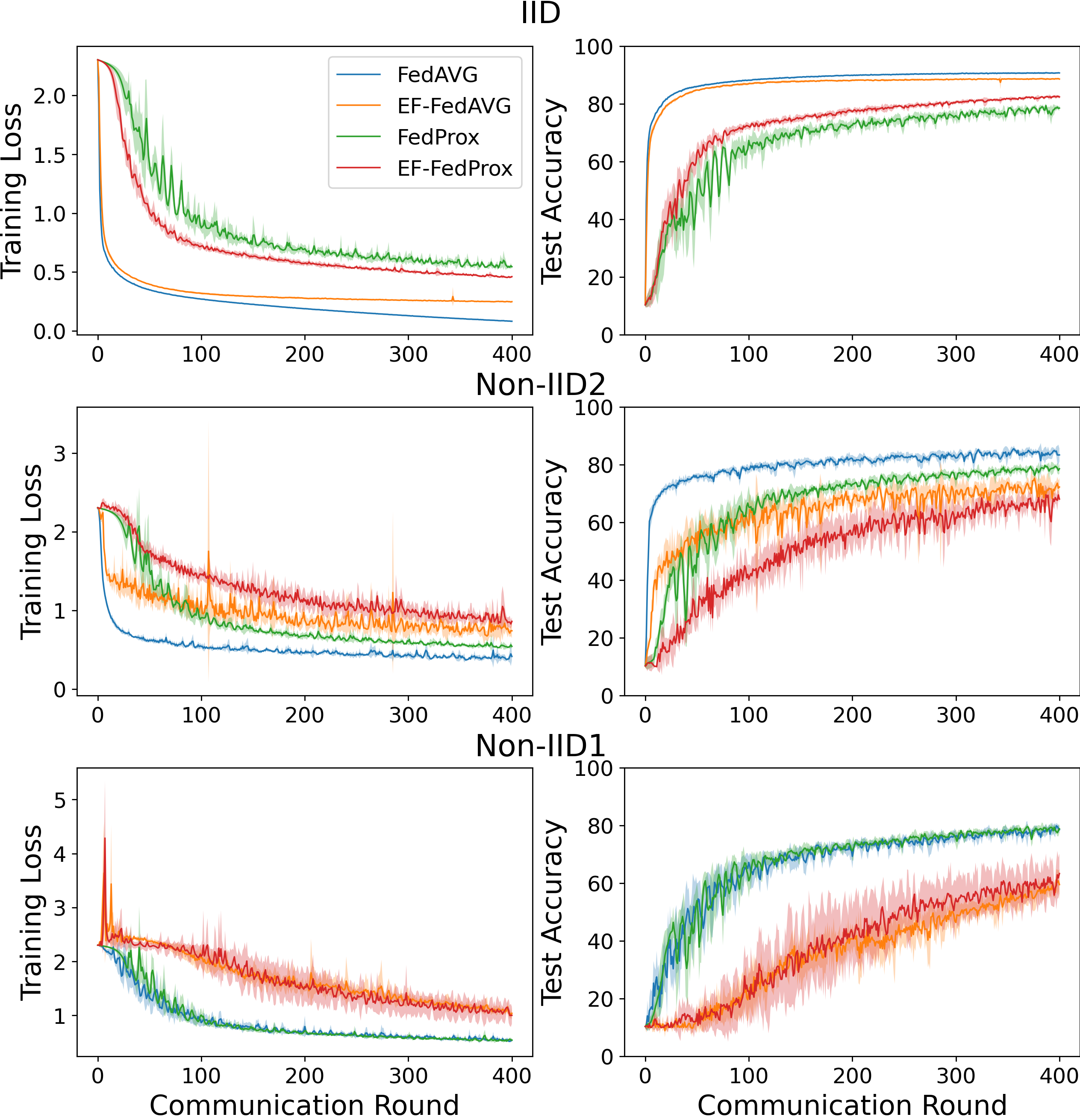} 
	\caption{Performance of FedAvg, error-feedback FedAvg, FedProx, and error-feedback FedProx with the fixed step size in (left plots -) training loss and (right plots -) test accuracy on FashionMNIST dataset considering three different partitioned data among the workers.}
	\label{fmnist-fix}
\end{figure}

\subsection{Data Similarity Conditions}
\noindent We evaluate federated algorithms under three data similarity conditions. 
In particular, we use three cases, i.e. IID, Non-IID2, and Non-IID1, for partitioning each dataset among the workers.
The IID case yields high data similarity by ensuring that each worker has the same data partition having the same size of samples with ten classes assigned according to a uniform distribution. 
The Non-IID2 case gives low data similarity, where each worker has data samples with only two classes. In particular, each worker is assigned two data chunks randomly from 20 data chunks representing the whole dataset with sorted classes. 
The Non-IID1 case provides extremely low data similarity by assigning a data partition containing samples with only a single class to each worker.

\subsection{Hyper-parameters}
\noindent For all algorithms, we set the number of communication rounds at $K=400$, chose the mini-batch size at $64$, and initialized the neural network weights using the default random initialization routines of the PyTorch framework. 
We chose the number of local updates at $T=30$ for FedAvg and error-feedback FedAvg, the learning rate of the inner solver for proximal updates at $0.1$ for FedProx and error-feedback FedProx, and $k$ to be $1\%$ of the trainable parameters (i.e. $k=4310$) for the top-$k$ sparsifier for error-feedback FedAvg and error-feedback FedProx. 
Furthermore, we employed three step size %scheduling 
strategies: a) fixed step size with $c=2$, b) diminishing step size with  $c=0.8$ and $\nu=0.51$, and c) step-decay step size with $\gamma_0=0.8$, $\alpha=2$, and $T=50$.
For fair empirical comparisons, we ran the experiments using five distinct random seeds for network initialization. 
Figures~\ref{mnist-fix} and~\ref{fmnist-fix} plot the average and standard deviation of training loss and test accuracy from running the algorithms with fixed step sizes over MNIST and FashionMNIST, respectively. 
We included additional experiments from running FedAvg, FedProx, error-feedback FedAvg, and error-feedback FedAvg with diminishing and step-decay step sizes over the MNIST dataset and FashionMNIST datasets. 
Particularly, we reported the result from running the algorithms with diminishing step sizes in Figures~\ref{mnist-diminishing} and~\ref{fmnist-diminishing}, and with the step-decay step sizes in Figures~\ref{mnist-step-decay} and~\ref{fmnist-step-decay}.

{\color{black}
It is worth mentioning that state-of-the-art methods might achieve higher test accuracy by employing more complex models and extensively tuning hyperparameters. However, the models we introduced, along with the specified settings for algorithms and step sizes, adequately serve our purpose: evaluating the optimization methods in the presence of various data similarities rather than achieving the highest possible accuracy on these tasks.}

\subsection{Discussions}
\begin{figure}[t]
	\centering
	\includegraphics[width=0.8\columnwidth]{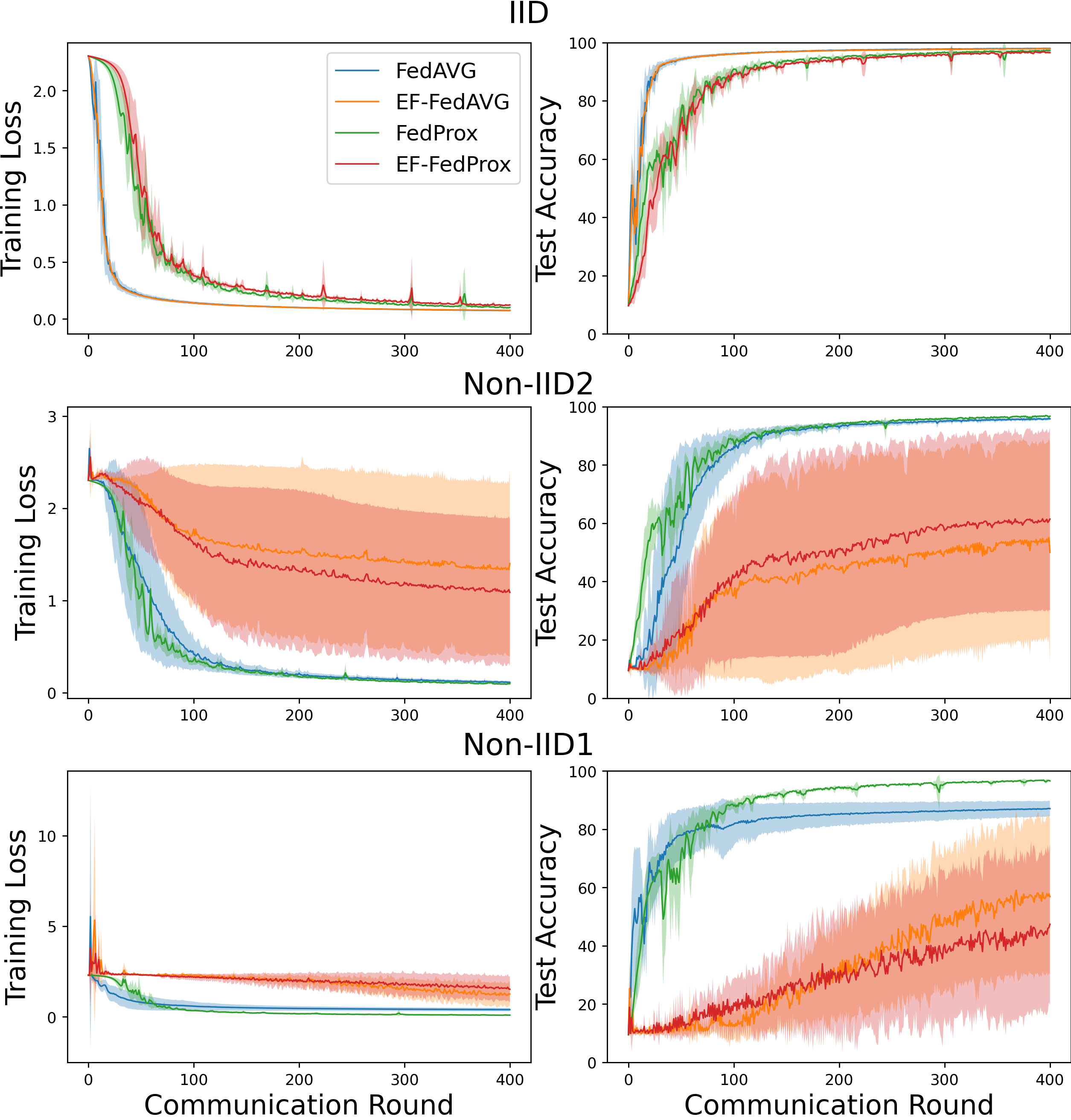} 
	\caption{Performance of FedAvg, error-feedback FedAvg, FedProx, and error-feedback FedProx with the diminishing step size in (left plots -) training loss and (right plots -) test accuracy on MNIST dataset considering three different partitioning data among the workers.}
	\label{mnist-diminishing}
\end{figure}

\begin{figure}[t]
	\centering
	\includegraphics[width=0.8\columnwidth]{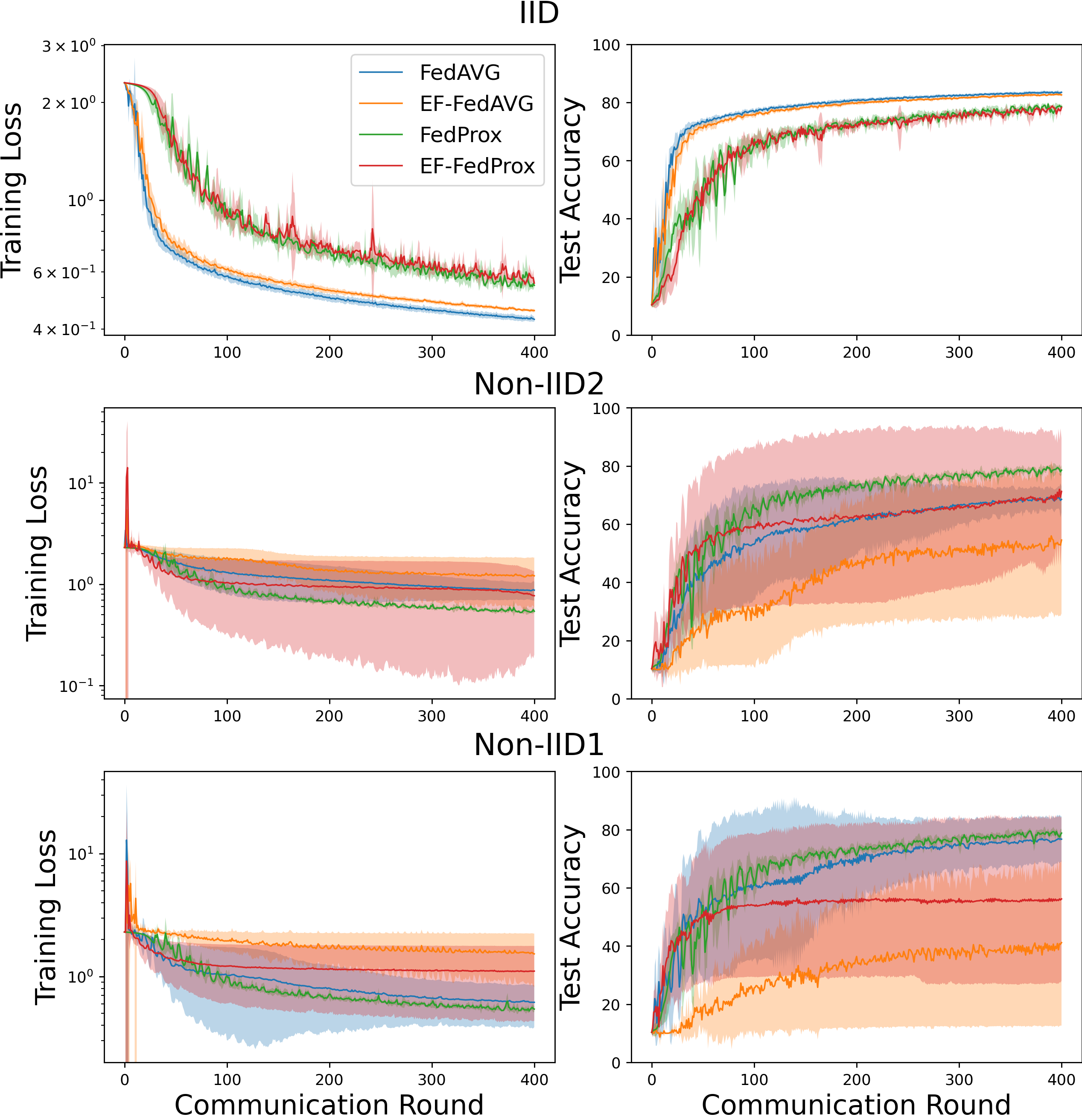} 
	\caption{Performance of FedAvg, error-feedback FedAvg, FedProx, and error-feedback FedProx with the diminishing step size in (left plots -) training loss and (right plots -) test accuracy on FashionMNIST dataset considering three different partitioning data among the workers.}
	\label{fmnist-diminishing}
\end{figure}

\begin{figure}[t]
	\centering
	\includegraphics[width=0.8\columnwidth]{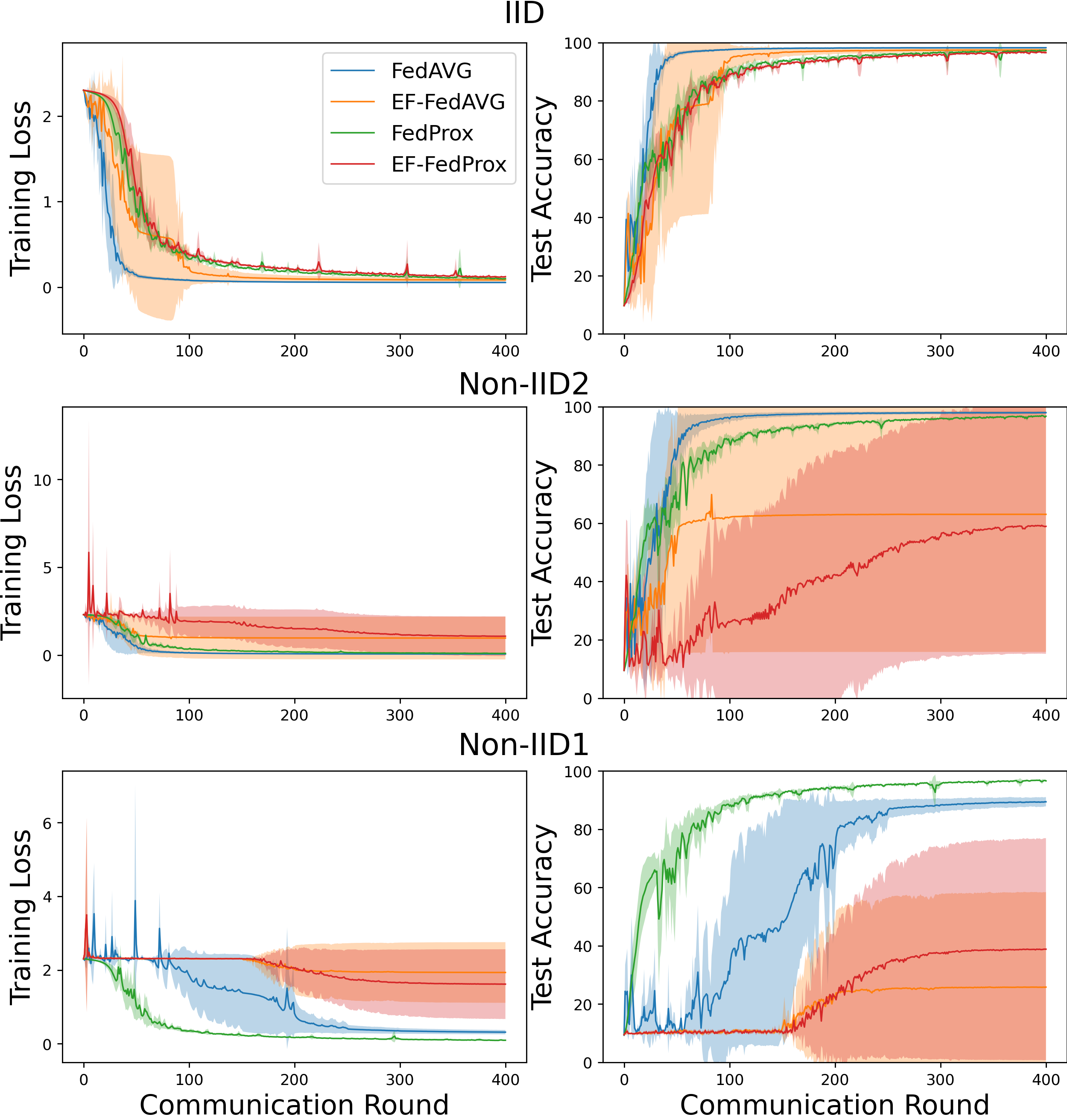} 
	\caption{Performance of FedAvg, error-feedback FedAvg, FedProx, and error-feedback FedProx with the step-decay step size in (left plots -) training loss and (right plots -) test accuracy on MNIST dataset considering three different partitioning data among the workers.}
	\label{mnist-step-decay}
\end{figure}

\begin{figure}[t]
	\centering
	\includegraphics[width=0.8\columnwidth]{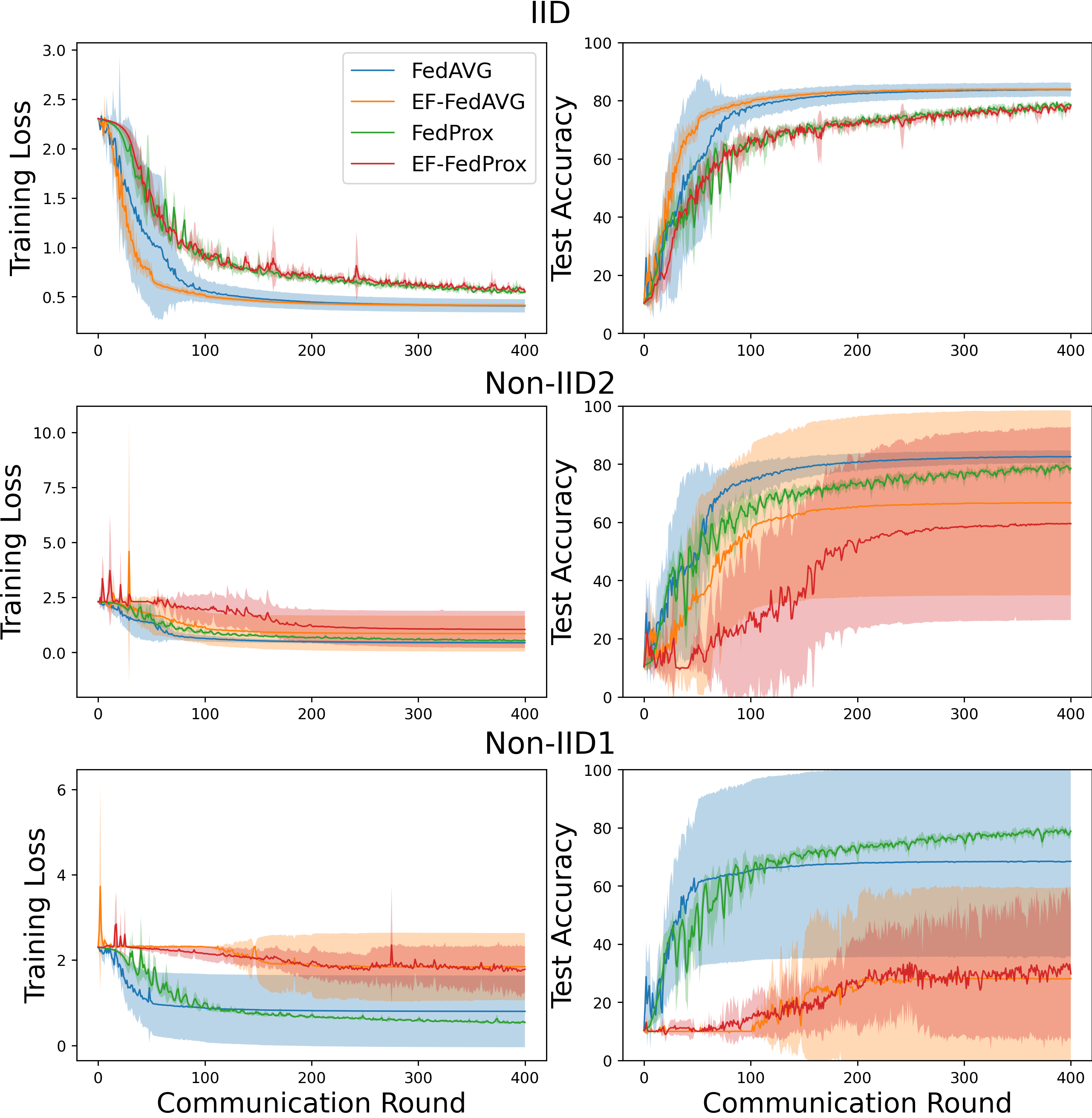} 
	\caption{Performance of FedAvg, error-feedback FedAvg, FedProx, and error-feedback FedProx with the step-decay step size in (left plots -) training loss and (right plots -) test accuracy on FashionMNIST dataset considering three different partitioning data among the workers.}
	\label{fmnist-step-decay}
\end{figure}

%\noindent 
{\color{black}
\subsubsection{(Error-feedback) FedAvg vs. (Error-feedback) FedProx Algorithm}
Figures ~\ref{mnist-fix} and ~\ref{fmnist-fix} show that with the same fixed step size, FedAvg surpasses FedProx in both full-precision and error-feedback updates in terms of solution accuracy and convergence speed, particularly when data similarity is high. For instance, in the IID case at $K=100$, FedAvg achieves an 80\% test accuracy, whereas FedProx only reaches 65\%. This disparity arises because, with a small learning rate of 0.1, the proximal update's regularization term in FedProx becomes dominant, causing the next local iterate $x_i^k$ to be nearly identical to the current global iterate $x^k$.
\subsubsection{Error-Feedback vs. Full-Precision Federated Learning Algorithms}
We also observed that error-feedback algorithms generally underperform compared to their full-precision counterparts, especially when data similarity is low. For example, in the Non-IID1 scenario at $K=200$, error-feedback algorithms achieve a 40\% test accuracy, whereas full-precision algorithms attain 70\%. This performance gap is due to the top-$k$ sparsifier in error-feedback algorithms introducing biased information, unlike in full-precision algorithms.
\subsubsection{Effect of Different Step Size Regimes}
Figures ~\ref{mnist-diminishing}, ~\ref{fmnist-diminishing}, ~\ref{mnist-step-decay}, and ~\ref{fmnist-step-decay} demonstrate consistent trends with the fixed step size results. Similar to those results, FedAvg tends to outperform FedProx, and error-feedback algorithms generally exhibit poorer performance than their full-precision counterparts under diminishing and step-decay step size regimes. Furthermore, our theoretical findings are validated, showing that these different algorithms can converge without requiring step sizes to be coupled to data similarity.
}

\section{Conclusions}
\noindent {\color{black} In this work, we have introduced a unified analysis framework for federated algorithms on non-convex problems without relying on data similarity assumptions. 
This framework employs the worst-case convergence bound in the general non-negative system ~\eqref{eqn:mainIneq} and utilizes convergence theorems that incorporate fixed, diminishing, and step-decay step size schedules. 
We demonstrated how to apply this framework to achieve strong convergence results for both full-precision and error-feedback federated algorithms. This includes FedAvg, FedProx, error-feedback FedAvg, and error-feedback FedProx, all with step sizes that are independent of data similarity parameters under standard conditions on objective functions. 
Finally, we substantiated our theoretical findings with numerical experiments, training CNN models on the MNIST and FashionMNIST datasets. These experiments showcase the performance of these federated algorithms under various data similarity conditions.
}

%\balance
\bibliographystyle{IEEEtran}
\bibliography{sample_R1.bib}

\begin{thebibliography}{10}
\providecommand{\url}[1]{#1}
\csname url@samestyle\endcsname
\providecommand{\newblock}{\relax}
\providecommand{\bibinfo}[2]{#2}
\providecommand{\BIBentrySTDinterwordspacing}{\spaceskip=0pt\relax}
\providecommand{\BIBentryALTinterwordstretchfactor}{4}
\providecommand{\BIBentryALTinterwordspacing}{\spaceskip=\fontdimen2\font plus
\BIBentryALTinterwordstretchfactor\fontdimen3\font minus \fontdimen4\font\relax}
\providecommand{\BIBforeignlanguage}[2]{{%
\expandafter\ifx\csname l@#1\endcsname\relax
\typeout{** WARNING: IEEEtran.bst: No hyphenation pattern has been}%
\typeout{** loaded for the language `#1'. Using the pattern for}%
\typeout{** the default language instead.}%
\else
\language=\csname l@#1\endcsname
\fi
#2}}
\providecommand{\BIBdecl}{\relax}
\BIBdecl

\bibitem{mcmahan2017communication}
B.~McMahan, E.~Moore, D.~Ramage, S.~Hampson, and B.~A. y~Arcas, ``Communication-efficient learning of deep networks from decentralized data,'' in \emph{Artificial intelligence and statistics}.\hskip 1em plus 0.5em minus 0.4em\relax PMLR, 2017, pp. 1273--1282.

\bibitem{li2020federated}
T.~Li, A.~K. Sahu, M.~Zaheer, M.~Sanjabi, A.~Talwalkar, and V.~Smith, ``Federated optimization in heterogeneous networks,'' \emph{Proceedings of Machine learning and systems}, vol.~2, pp. 429--450, 2020.

\bibitem{gorbunov2021local}
E.~Gorbunov, F.~Hanzely, and P.~Richt{\'a}rik, ``Local {SGD}: Unified theory and new efficient methods,'' in \emph{International Conference on Artificial Intelligence and Statistics}.\hskip 1em plus 0.5em minus 0.4em\relax PMLR, 2021, pp. 3556--3564.

\bibitem{khaled2020tighter}
A.~Khaled, K.~Mishchenko, and P.~Richt{\'a}rik, ``Tighter theory for local {SGD} on identical and heterogeneous data,'' in \emph{International Conference on Artificial Intelligence and Statistics}.\hskip 1em plus 0.5em minus 0.4em\relax PMLR, 2020, pp. 4519--4529.

\bibitem{needell2014stochastic}
D.~Needell, R.~Ward, and N.~Srebro, ``Stochastic gradient descent, weighted sampling, and the randomized kaczmarz algorithm,'' \emph{Advances in neural information processing systems}, vol.~27, 2014.

\bibitem{yuan2022convergence}
X.~Yuan and P.~Li, ``On convergence of fedprox: Local dissimilarity invariant bounds, non-smoothness and beyond,'' \emph{Advances in Neural Information Processing Systems}, vol.~35, pp. 10\,752--10\,765, 2022.

\bibitem{mishchenko2022proxskip}
K.~Mishchenko, G.~Malinovsky, S.~Stich, and P.~Richt{\'a}rik, ``Proxskip: Yes! local gradient steps provably lead to communication acceleration! finally!'' in \emph{International Conference on Machine Learning}.\hskip 1em plus 0.5em minus 0.4em\relax PMLR, 2022, pp. 15\,750--15\,769.

\bibitem{karimireddy2020scaffold}
S.~P. Karimireddy, S.~Kale, M.~Mohri, S.~Reddi, S.~Stich, and A.~T. Suresh, ``Scaffold: Stochastic controlled averaging for federated learning,'' in \emph{International conference on machine learning}.\hskip 1em plus 0.5em minus 0.4em\relax PMLR, 2020, pp. 5132--5143.

\bibitem{pathak2020fedsplit}
R.~Pathak and M.~J. Wainwright, ``Fedsplit: An algorithmic framework for fast federated optimization,'' \emph{Advances in neural information processing systems}, vol.~33, pp. 7057--7066, 2020.

\bibitem{zhang2021fedpd}
X.~Zhang, M.~Hong, S.~Dhople, W.~Yin, and Y.~Liu, ``Fedpd: A federated learning framework with adaptivity to non-iid data,'' \emph{IEEE Transactions on Signal Processing}, vol.~69, pp. 6055--6070, 2021.

\bibitem{gong2022fedadmm}
Y.~Gong, Y.~Li, and N.~M. Freris, ``{FedADMM:} a robust federated deep learning framework with adaptivity to system heterogeneity,'' in \emph{2022 IEEE 38th International Conference on Data Engineering (ICDE)}.\hskip 1em plus 0.5em minus 0.4em\relax IEEE, 2022, pp. 2575--2587.

\bibitem{yu2019parallel}
H.~Yu, S.~Yang, and S.~Zhu, ``Parallel restarted {SGD} with faster convergence and less communication: Demystifying why model averaging works for deep learning,'' in \emph{Proceedings of the AAAI Conference on Artificial Intelligence}, vol.~33, 2019, pp. 5693--5700.

\bibitem{wang2021cooperative}
J.~Wang and G.~Joshi, ``Cooperative {SGD}: A unified framework for the design and analysis of local-update {SGD} algorithms,'' \emph{The Journal of Machine Learning Research}, vol.~22, no.~1, pp. 9709--9758, 2021.

\bibitem{haddadpour2019local}
F.~Haddadpour, M.~M. Kamani, M.~Mahdavi, and V.~Cadambe, ``Local {SGD} with periodic averaging: Tighter analysis and adaptive synchronization,'' \emph{Advances in Neural Information Processing Systems}, vol.~32, 2019.

\bibitem{glasgow2022sharp}
M.~R. Glasgow, H.~Yuan, and T.~Ma, ``Sharp bounds for federated averaging (local {SGD}) and continuous perspective,'' in \emph{International Conference on Artificial Intelligence and Statistics}.\hskip 1em plus 0.5em minus 0.4em\relax PMLR, 2022, pp. 9050--9090.

\bibitem{pmlr-v54-mcmahan17a}
\BIBentryALTinterwordspacing
B.~McMahan, E.~Moore, D.~Ramage, S.~Hampson, and B.~A.~y. Arcas, ``{Communication-efficient learning of deep networks from decentralized data},'' in \emph{Proceedings of the 20th International Conference on Artificial Intelligence and Statistics}, ser. Proceedings of Machine Learning Research, A.~Singh and J.~Zhu, Eds., vol.~54.\hskip 1em plus 0.5em minus 0.4em\relax PMLR, 20--22 Apr 2017, pp. 1273--1282. [Online]. Available: \url{https://proceedings.mlr.press/v54/mcmahan17a.html}
\BIBentrySTDinterwordspacing

\bibitem{woodworth2020local}
B.~Woodworth, K.~K. Patel, S.~Stich, Z.~Dai, B.~Bullins, B.~Mcmahan, O.~Shamir, and N.~Srebro, ``Is local {SGD} better than minibatch {SGD}?'' in \emph{International Conference on Machine Learning}.\hskip 1em plus 0.5em minus 0.4em\relax PMLR, 2020, pp. 10\,334--10\,343.

\bibitem{alistarh2017qsgd}
D.~Alistarh, D.~Grubic, J.~Li, R.~Tomioka, and M.~Vojnovic, ``{QSGD:} communication-efficient {SGD} via gradient quantization and encoding,'' \emph{Advances in neural information processing systems}, vol.~30, 2017.

\bibitem{khirirat2021flexible}
S.~Khirirat, S.~Magn{\'u}sson, A.~Aytekin, and M.~Johansson, ``A flexible framework for communication-efficient machine learning,'' in \emph{Proceedings of the AAAI Conference on Artificial Intelligence}, vol.~35, 2021, pp. 8101--8109.

\bibitem{khirirat2020compressed}
S.~Khirirat, S.~Magn{\'u}sson, and M.~Johansson, ``Compressed gradient methods with hessian-aided error compensation,'' \emph{IEEE Transactions on Signal Processing}, vol.~69, pp. 998--1011, 2020.

\bibitem{richtarik2021ef21}
P.~Richt{\'a}rik, I.~Sokolov, and I.~Fatkhullin, ``{EF21}: A new, simpler, theoretically better, and practically faster error feedback,'' \emph{Advances in Neural Information Processing Systems}, vol.~34, pp. 4384--4396, 2021.

\bibitem{khirirat2022eco}
S.~Khirirat, S.~Magn{\'u}sson, and M.~Johansson, ``{Eco-Fedsplit:} federated learning with error-compensated compression,'' in \emph{ICASSP 2022-2022 IEEE International Conference on Acoustics, Speech and Signal Processing (ICASSP)}.\hskip 1em plus 0.5em minus 0.4em\relax IEEE, 2022, pp. 5952--5956.

\bibitem{basu2019qsparse}
D.~Basu, D.~Data, C.~Karakus, and S.~Diggavi, ``{Qsparse-local-SGD:} distributed {SGD} with quantization, sparsification and local computations,'' \emph{Advances in Neural Information Processing Systems}, vol.~32, 2019.

\bibitem{bertsekas2011incremental}
D.~P. Bertsekas \emph{et~al.}, ``Incremental gradient, subgradient, and proximal methods for convex optimization: A survey,'' \emph{Optimization for Machine Learning}, vol. 2010, no. 1-38, p.~3, 2011.

\bibitem{10026503}
S.~Khirirat, X.~Wang, S.~Magnússon, and M.~Johansson, ``Improved step-size schedules for proximal noisy gradient methods,'' \emph{IEEE Transactions on Signal Processing}, vol.~71, pp. 189--201, 2023.

\bibitem{nguyen2018sgd}
L.~Nguyen, P.~H. Nguyen, M.~Dijk, P.~Richt{\'a}rik, K.~Scheinberg, and M.~Tak{\'a}c, ``{SGD} and hogwild! convergence without the bounded gradients assumption,'' in \emph{International Conference on Machine Learning}.\hskip 1em plus 0.5em minus 0.4em\relax PMLR, 2018, pp. 3750--3758.

\bibitem{ge2019step}
R.~Ge, S.~M. Kakade, R.~Kidambi, and P.~Netrapalli, ``The step decay schedule: A near optimal, geometrically decaying learning rate procedure for least squares,'' \emph{Advances in neural information processing systems}, vol.~32, 2019.

\bibitem{wang2021convergence}
X.~Wang, S.~Magn{\'u}sson, and M.~Johansson, ``On the convergence of step decay step-size for stochastic optimization,'' \emph{Advances in Neural Information Processing Systems}, vol.~34, pp. 14\,226--14\,238, 2021.

\bibitem{schaipp2023stochastic}
F.~Schaipp, R.~M. Gower, and M.~Ulbrich, ``A stochastic proximal polyak step size,'' \emph{Transactions on Machine Learning Research}, 2023.

\bibitem{loizou2021stochastic}
N.~Loizou, S.~Vaswani, I.~H. Laradji, and S.~Lacoste-Julien, ``Stochastic polyak step-size for {SGD}: An adaptive learning rate for fast convergence,'' in \emph{International Conference on Artificial Intelligence and Statistics}.\hskip 1em plus 0.5em minus 0.4em\relax PMLR, 2021, pp. 1306--1314.

\bibitem{robbins1971convergence}
H.~Robbins and D.~Siegmund, ``A convergence theorem for non negative almost supermartingales and some applications,'' in \emph{Optimizing methods in statistics}.\hskip 1em plus 0.5em minus 0.4em\relax Elsevier, 1971, pp. 233--257.

\bibitem{li2019convergence}
X.~Li and F.~Orabona, ``On the convergence of stochastic gradient descent with adaptive stepsizes,'' in \emph{The 22nd international conference on artificial intelligence and statistics}.\hskip 1em plus 0.5em minus 0.4em\relax PMLR, 2019, pp. 983--992.

\bibitem{khaled2022better}
\BIBentryALTinterwordspacing
A.~Khaled and P.~Richt{\'a}rik, ``Better theory for {SGD} in the nonconvex world,'' \emph{Transactions on Machine Learning Research}, 2023. [Online]. Available: \url{https://openreview.net/forum?id=AU4qHN2VkS}
\BIBentrySTDinterwordspacing

\bibitem{feyzmahdavian2016asynchronous}
H.~R. Feyzmahdavian, A.~Aytekin, and M.~Johansson, ``An asynchronous mini-batch algorithm for regularized stochastic optimization,'' \emph{IEEE Transactions on Automatic Control}, vol.~61, no.~12, pp. 3740--3754, 2016.

\bibitem{xu2021step}
A.~Xu, Z.~Huo, and H.~Huang, ``Step-ahead error feedback for distributed training with compressed gradient,'' in \emph{Proceedings of the AAAI Conference on Artificial Intelligence}, vol.~35, 2021, pp. 10\,478--10\,486.

\bibitem{nguyen2020fast}
H.~T. Nguyen, V.~Sehwag, S.~Hosseinalipour, C.~G. Brinton, M.~Chiang, and H.~V. Poor, ``Fast-convergent federated learning,'' \emph{IEEE Journal on Selected Areas in Communications}, vol.~39, no.~1, pp. 201--218, 2020.

\bibitem{reisizadeh2020fedpaq}
A.~Reisizadeh, A.~Mokhtari, H.~Hassani, A.~Jadbabaie, and R.~Pedarsani, ``Fedpaq: A communication-efficient federated learning method with periodic averaging and quantization,'' in \emph{International Conference on Artificial Intelligence and Statistics}.\hskip 1em plus 0.5em minus 0.4em\relax PMLR, 2020, pp. 2021--2031.

\bibitem{khirirat2018gradient}
S.~Khirirat, M.~Johansson, and D.~Alistarh, ``Gradient compression for communication-limited convex optimization,'' in \emph{2018 IEEE Conference on Decision and Control (CDC)}.\hskip 1em plus 0.5em minus 0.4em\relax IEEE, 2018, pp. 166--171.

\bibitem{karimireddy2019error}
S.~P. Karimireddy, Q.~Rebjock, S.~Stich, and M.~Jaggi, ``Error feedback fixes sign{SGD} and other gradient compression schemes,'' in \emph{International Conference on Machine Learning}.\hskip 1em plus 0.5em minus 0.4em\relax PMLR, 2019, pp. 3252--3261.

\bibitem{gao2021convergence}
H.~Gao, A.~Xu, and H.~Huang, ``On the convergence of communication-efficient local {SGD} for federated learning,'' in \emph{Proceedings of the AAAI Conference on Artificial Intelligence}, vol.~35, 2021, pp. 7510--7518.

\bibitem{lecun1998mnist}
Y.~LeCun, ``The mnist database of handwritten digits,'' \emph{http://yann. lecun. com/exdb/mnist/}, 1998.

\bibitem{xiao2017fashion}
H.~Xiao, K.~Rasul, and R.~Vollgraf, ``Fashion-mnist: a novel image dataset for benchmarking machine learning algorithms,'' \emph{arXiv preprint arXiv:1708.07747}, 2017.

\bibitem{imambi2021pytorch}
S.~Imambi, K.~B. Prakash, and G.~Kanagachidambaresan, ``Pytorch,'' \emph{Programming with TensorFlow: Solution for Edge Computing Applications}, pp. 87--104, 2021.

\bibitem{roberts2022simplified}
L.~Roberts and E.~Smyth, ``A simplified convergence theory for byzantine resilient stochastic gradient descent,'' \emph{EURO Journal on Computational Optimization}, vol.~10, p. 100038, 2022.

\end{thebibliography}
\begin{IEEEbiography}
[{\includegraphics[width=1in,height=1.25in,clip,keepaspectratio]{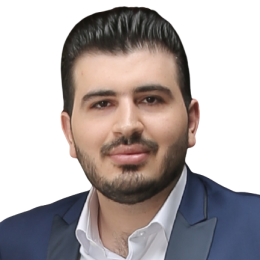}}]{Ali Beikmohammadi}
earned his B.Sc. in Electrical Engineering from Bu-Ali Sina University, Hamedan, Iran, in 2017, and subsequently completed his M.Sc. in Electrical Engineering at Amirkabir University of Technology, Tehran, Iran, in 2019. Currently, he is a Ph.D. candidate in Computer and Systems Sciences at Stockholm University, Sweden, focusing on research areas such as Reinforcement Learning, Deep Learning, and Federated Learning, both in theory and applications.
\end{IEEEbiography}
\begin{IEEEbiography}[{\includegraphics[width=1in,height=1.25in,clip,keepaspectratio]{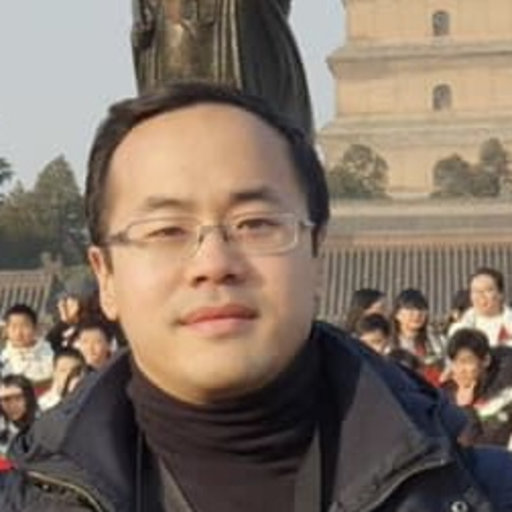}}]{Sarit Khirirat}
 received his B.Eng. in Electrical Engineering from Chulalongkorn University, Thailand, in 2013; the Master’s degree in Systems, Control and Robotics from KTH Royal Institute of Technology, Sweden, in 2017; and the PhD at the Division of Decision and Control Systems from the same institution in 2022, supported by the Wallenberg AI, Autonomous Systems and Software program, Sweden’s largest individual research funding program. He is currently a postdoctoral fellow  at King Abdullah University of Science and Technology (KAUST). His research interests include distributed optimization algorithms for federated learning applications. 
\end{IEEEbiography}
\begin{IEEEbiography}[{\includegraphics[width=1in,height=1.25in,clip,keepaspectratio]{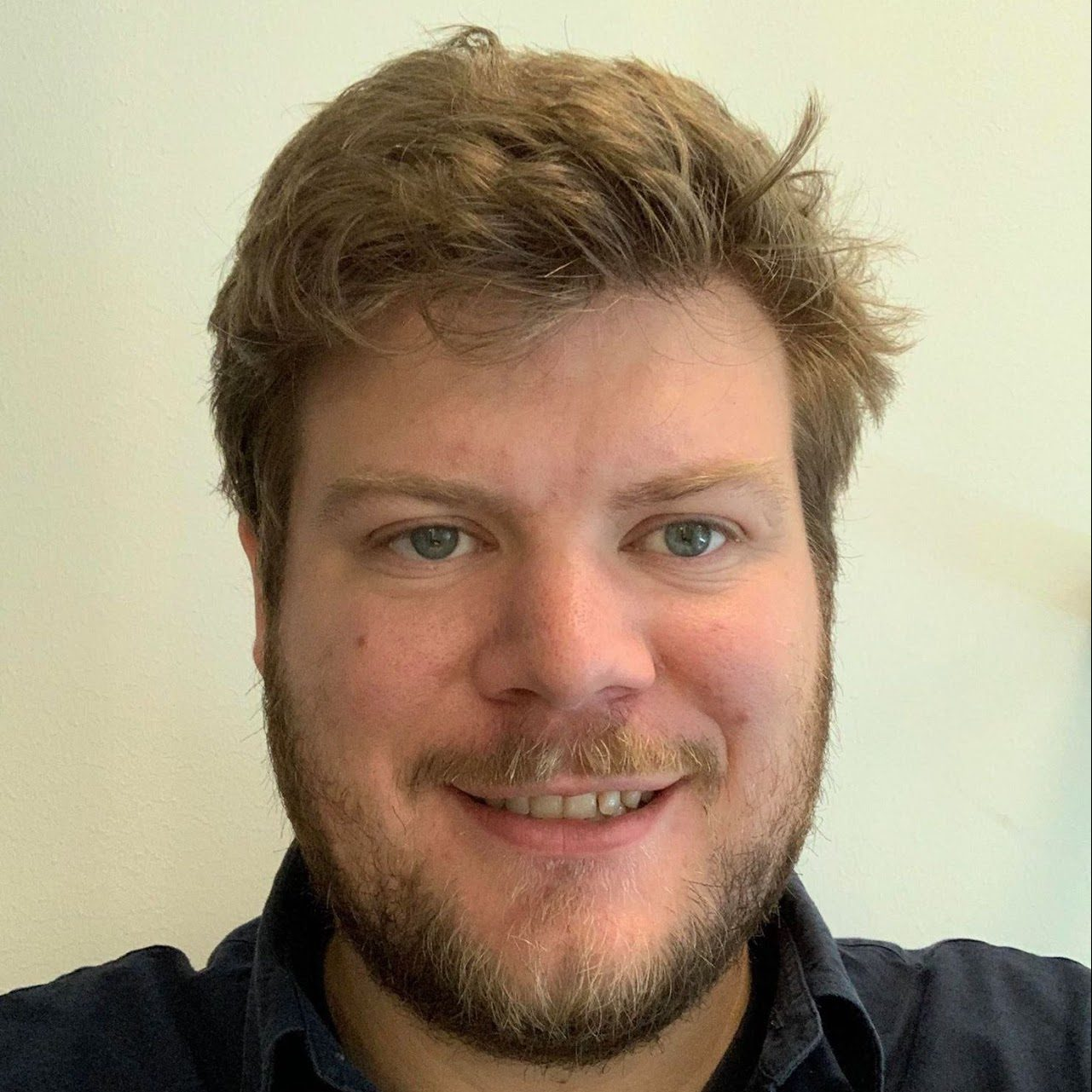}}]{Sindri Magn\'usson}
is an Associate Professor in the Department of Computer and Systems Science at Stockholm University, Sweden. He received a B.Sc. degree in Mathematics from the University of Iceland, Reykjav´ık Iceland, in 2011, a Masters degree in Applied Mathematics (Optimization and Systems Theory) from KTH Royal Institute of Technology, Stockholm, Sweden, in 2013, and the PhD in Electrical Engineering from the same institution, in 2017. He was a postdoctoral researcher 2018-2019 at Harvard University, Cambridge, MA, and a visiting PhD student at Harvard University for 9 months in 2015 and 2016. His research interests include large-scale distributed/parallel optimization, machine learning, and control.
\end{IEEEbiography}

\clearpage
\appendices
\section{Useful Inequalities}

\noindent We present the following inequalities from linear algebra. 
For $\theta>0$ and $x_1,\ldots,x_n,y\in\mathbb{R}^d$, 
\begin{eqnarray}
	\left\| \sum_{i=1}^n x_i \right\|^2 & \leq & n \sum_{i=1}^n \| x_i\|^2.  \label{eq:Triangle_sq} \\ 
	\| x+y \|^2 & \leq & (1+\theta)\| x  \|^2 + (1+1/\theta)\| y \|^2. \label{eq:Young} \\
	-2\langle x,y \rangle & = & - \| x\|^2 - \| y \|^2 + \| x-y \|^2. \label{eq:Exact_sq} \\
	2\langle x,y \rangle & \leq & \| x \|^2 + \| y \|^2.  \label{eq:InExact_sq}
\end{eqnarray}	
%
%The next lemmas allow us to derive the upper-bounds, which depend on specific step size choices. 

\begin{lemma}\label{lemma:trick_fixed}
$(1+a\gamma^2)^K \leq \exp(ac^2)$ if $\gamma = c/\sqrt{K}$ for $a,c>0$ and $K\in\mathbb{N}$.
\end{lemma}
\begin{proof}
By the fact that $x = \exp(\ln(x))$ and that $\ln(1+x)\leq x$ for $x\geq -1$, we have
%\begin{align*}
  $(1+a\gamma^2)^K 
   =  \exp(K \ln(1+a\gamma^2)) 
   \leq \exp(K a\gamma^2)$, 
%\end{align*}
for $a>0$ and $K\in\mathbb{N}$.
If $\gamma =c/\sqrt{K}$ for $c>0$, then  
%\begin{align*}
    $(1+a\gamma^2)^K \leq \exp(ac^2)$.
%\end{align*}
\end{proof}

\begin{lemma}\label{lemma:trick_decreasing}
Let $\gamma_k = c/(k+1)^{\nu}$ for $\nu \in (1/2,1)$. Then, $\sum_{k=0}^{K-1} \gamma_k^2 \leq {2\nu c^2}/(2\nu-1)$.
\end{lemma}
\begin{proof}
 By the fact that $\gamma_k = c/(k+1)^{\nu}$ for $\nu \in (1/2,1)$ decreases with respect to $k$, 
  \begin{eqnarray*}
\sum_{k=0}^{K-1} \gamma_k^2 
    %& = & c^2 \sum_{k=0}^{K-1} \frac{1}{(k+1)^{2\nu}} \\
     =  c^2 + c^2 \sum_{k=1}^{K-1} \frac{1}{(k+1)^{2\nu}} 
%     \leq  c^2 + c^2 \int_{k=0}^{K-1} \frac{dk}{(k+1)^{2\nu}} \\
     \leq  c^2 {+} c^2 \int_{k=0}^{\infty} \frac{dk}{(k+1)^{2\nu}}.
 \end{eqnarray*}
 Since 
\(   \int_{k=0}^{\infty} \frac{dk}{(k+1)^{2\nu}} 
    % & = & \frac{1-1/(K+1)^{2\nu-1}}{2\nu-1}  \\
      =  \frac{1}{2\nu-1},
\) we complete the proof. 
%  \begin{eqnarray*}
%\sum_{k=0}^{K-1} \gamma_k^2 
%    & \leq & c^2 + \frac{c^2}{2\nu-1} = \frac{2\nu c^2}{2\nu-1}.
% \end{eqnarray*}
\end{proof}

\section{Proof of Theorem~\ref{thm:fixed}}
\noindent Define $\alpha_{-1}=1$ and 
\begin{eqnarray}\label{eq:alpha_fixedstepsize}
    \alpha_k = \frac{\alpha_{k-1}}{1+b_1\gamma^2}, \quad \text{for} \quad k \geq 0.
\end{eqnarray}
By \eqref{eq:alpha_fixedstepsize}, the sequence $\{\alpha_k\}$ can be expressed equivalently as: 
\begin{eqnarray}\label{eq:alpha_fixedstepsize_equi}
    \alpha_k = \begin{cases} 1 & \text{for} \quad  k = -1 \\ \frac{1}{(1+b_1\gamma^2)^{k+1}} & \text{for} \quad  k \geq 0. \end{cases}
\end{eqnarray}
Therefore, $\alpha_k>0$ decreases with respect to $k$. Next, by setting $\gamma_k = \gamma$ into \eqref{eqn:mainIneq}, by re-arranging the terms,
\begin{eqnarray*}
    \alpha_k W_k 
    & \leq &  \frac{\alpha_k (1+b_1\gamma^2)V_k}{b_2\gamma} - \frac{\alpha_k V_{k+1}}{b_2\gamma} + \frac{b_3\gamma}{b_2}\alpha_k \\
    & \overset{\eqref{eq:alpha_fixedstepsize}}{=} & \frac{\alpha_{k-1} V_k}{b_2\gamma} - \frac{\alpha_k V_{k+1}}{b_2\gamma} + \frac{b_3\gamma}{b_2}\alpha_k.
\end{eqnarray*}
Next, denote $\tilde V_k = \alpha_{k-1} V_k > 0$ for $k\geq 0$. Then, 
\begin{align*}
    \alpha_k W_k
    & \leq \frac{\tilde V_k}{b_2\gamma} - \frac{\tilde V_{k+1}}{b_2\gamma} + \frac{b_3\gamma}{b_2}\alpha_k.
\end{align*}
Next, by re-arranging the terms, 
\begin{align*}
\mathop{\min}_{0\leq k \leq K-1} W_k
&\leq \frac{1}{\sum_{k=0}^{K-1} \alpha_k} \sum_{k=0}^{K-1} \alpha_k W_k \\ 
& = \frac{\tilde V_0 - \tilde V_K}{b_2\gamma \sum_{k=0}^{K-1} \alpha_k} + \frac{b_3\gamma}{b_2} \\
& \leq  \frac{\tilde V_0 }{b_2\gamma \sum_{k=0}^{K-1} \alpha_k} + \frac{b_3\gamma}{b_2}.
\end{align*}
Since $\tilde V_0=\alpha_{-1}V_0=V_0$ and also 
\begin{eqnarray*}
    \sum_{k=0}^{K-1} \alpha_k 
    \geq K \alpha_{K-1} 
     \overset{\eqref{eq:alpha_fixedstepsize_equi}}{=} \frac{K}{(1+b_1\gamma^2)^K},
\end{eqnarray*}
we have 
\begin{align*}
\mathop{\min}_{0\leq k \leq K-1} W_k
&  \leq \frac{(1+b_1\gamma^2)^K}{K}\frac{V_0}{b_2\gamma}  + \frac{b_3\gamma}{b_2}.
\end{align*}

Finally, if $\gamma=c/\sqrt{K}$, then from Lemma~\ref{lemma:trick_fixed} we complete the proof. 
%\begin{align*}
%\mathop{\min}_{0\leq k \leq K-1} W_k
%&  \leq \frac{1}{\sqrt{K}} \left( \frac{\exp(b_1 c^2) V_0}{b_2c}  + \frac{b_3c}{b_2} \right). 
%\end{align*}
%Hence, we complete the proof. 

\section{Proof of Theorem~\ref{thm:diminishing}}
\noindent Define $\alpha_{-1}=1$ and
\begin{eqnarray}\label{eq:alpha_decreasing_stepsize}
    \alpha_k = \alpha_{k-1} \frac{\gamma_k}{\gamma_{k-1}(1+b_1\gamma_k^2)} \quad \text{for} \quad k \geq 0.
\end{eqnarray}
By \eqref{eq:alpha_decreasing_stepsize}, the sequence $\{\alpha_k\}$ can be rewritten into: 
\begin{eqnarray}\label{eq:alpha_decreasing_stepsize_equi}
    \alpha_k = \begin{cases}
        1 & \text{for} \quad k=-1 \\ 
        \frac{\gamma_k}{\gamma_{-1} \prod_{l=0}^{k}(1+b_1\gamma_l^2)} & \text{for} \quad k \geq 0.
    \end{cases}
\end{eqnarray}
Notice that  $\alpha_k$  decreases with $k$ if $\gamma_k$ decreases with $k$.
Next, by \eqref{eqn:mainIneq}, by re-arranging the terms and by the fact that $\frac{\alpha_{k-1}}{\gamma_{k-1}} = \frac{\alpha_k(1+b_1\gamma_k^2)}{\gamma_k}$, 
\begin{eqnarray*}
    \alpha_k W_k 
    & \leq & \frac{\alpha_k(1+ b_1 \gamma_k^2 )}{b_2 \gamma_k}V_k - \frac{\alpha_k}{b_2\gamma_k}V_{k+1} + \frac{b_3}{b_2} \alpha_k \gamma_k \\
    & \overset{\eqref{eq:alpha_decreasing_stepsize}}{=} & \frac{\tilde V_k - \tilde V_{k+1}}{b_2}  + \frac{b_3}{b_2} \alpha_k \gamma_k,
\end{eqnarray*}
where $\tilde V_k = \frac{\alpha_{k-1}}{\gamma_{k-1}}V_{k}>0$. Therefore, 
\begin{align*}
\mathop{\min}_{0\leq k \leq K-1} W_k
&\leq \frac{1}{\sum_{k=0}^{K-1} \alpha_k} \sum_{k=0}^{K-1} \alpha_k W_k \\ 
& \leq  \frac{\tilde V_0 - \tilde V_{K}}{b_2\sum_{k=0}^{K-1} \alpha_k}  + \frac{b_3}{b_2} \frac{\sum_{k=0}^{K-1}\alpha_k \gamma_k} {\sum_{k=0}^{K-1} \alpha_k} \\
& \leq \frac{\tilde V_0 }{b_2\sum_{k=0}^{K-1} \alpha_k}  + \frac{b_3}{b_2} \frac{\sum_{k=0}^{K-1}\alpha_k \gamma_k} {\sum_{k=0}^{K-1} \alpha_k}.
\end{align*}
Since $\alpha_{-1}=1$ and 
\begin{align*}
    \sum_{k=0}^{K-1}\alpha_k \gamma_k & = \frac{\alpha_{-1}}{\gamma_{-1}}\sum_{k=0}^{K-1} \frac{\gamma_k^2}{\prod_{l=0}^{k}(1+b_1\gamma_l^2)}  \\
    %& \leq \frac{\alpha_{-1}}{\gamma_{-1}}\sum_{k=0}^{K-1} {\gamma_k^2} \\
    & \leq  \frac{1}{\gamma_{-1}}\sum_{k=0}^{K-1} {\gamma_k^2},
\end{align*}
where the last inequality comes from the fact that $1/\prod_{l=0}^{k}(1+b_1\gamma_l^2) \leq 1$ with $b_1>0$ and $\gamma_k>0$ for all $k\geq 0$, we have 
\begin{align*}
\mathop{\min}_{0\leq k \leq K-1} W_k
&\leq \left( \frac{\tilde V_0}{b_2}+ \frac{b_3}{b_2 \gamma_{-1}} \sum_{k=0}^{K-1} \gamma_k^2 \right) \frac{1}{\sum_{k=0}^{K-1} \alpha_k}.
\end{align*}
Next, by the fact that  
\begin{eqnarray*}
    \sum_{k=0}^{K-1} \alpha_k 
    \geq K \alpha_{K-1}  \overset{\eqref{eq:alpha_decreasing_stepsize_equi}}{=}  \frac{K \gamma_{K-1}}{\gamma_{-1} \prod_{k=0}^{K-1} (1+b_1 \gamma_k^2)}, 
\end{eqnarray*}
that $x = \exp(\ln(x))$ and that $\ln(1+x) \leq x$ for $x> -1$,
\begin{align*}
& \mathop{\min}_{0\leq k \leq K-1} W_k
\leq \left( \frac{\tilde V_0}{b_2}+ \frac{b_3}{b_2\gamma_{-1}} \sum_{k=0}^{K-1} \gamma_k^2 \right) \frac{\gamma_{-1}  \prod_{k=0}^{K-1} (1+b_1 \gamma_k^2) }{K \gamma_{K-1}} \\
& =\left( \frac{\tilde V_0}{b_2}+ \frac{b_3}{b_2\gamma_{-1}} \sum_{k=0}^{K-1} \gamma_k^2 \right) \frac{\gamma_{-1}  \exp\left( \sum_{k=0}^{K-1} \ln(1+b_1 \gamma_k^2) \right)}{K \gamma_{K-1}} \\
& \leq \left( \frac{\tilde V_0}{b_2}+ \frac{b_3}{b_2 \gamma_{-1}} \sum_{k=0}^{K-1} \gamma_k^2 \right) \frac{\gamma_{-1}  \exp\left( b_1 \sum_{k=0}^{K-1} \gamma_k^2 \right)}{K \gamma_{K-1}}.
\end{align*}

If $\gamma_k = c/(k+1)^{\nu}$ for $\nu \in (1/2,1)$ and $k\geq 0$, and $\gamma_{-1}=1$,
then 
\begin{align*}
\mathop{\min}_{0\leq k \leq K-1} W_k
&\leq \left( \frac{\tilde V_0}{b_2}+ \frac{b_3}{b_2 \gamma_{-1}} \frac{2\nu c^2}{2\nu-1} \right) \frac{\gamma_{-1}  \exp\left( b_1 \frac{2\nu c^2}{2\nu-1}  \right)}{K^{1-\nu} c}.
\end{align*}
Finally, plugging $\tilde V_0= \frac{\alpha_{-1}}{\gamma_{-1}}V_0 = \frac{1}{\gamma_{-1}}V_0$ and $\gamma_{-1}=1$ yields
\begin{align*}
\mathop{\min}_{0\leq k \leq K-1} W_k
&\leq \left( \frac{V_0}{b_2\gamma_{-1}}+ \frac{b_3}{b_2 \gamma_{-1}} \frac{2\nu c^2}{2\nu-1} \right) \frac{\gamma_{-1}  \exp\left( b_1 \frac{2\nu c^2}{2\nu-1}  \right)}{K^{1-\nu} c} \\
& = \left( \frac{V_0}{b_2}+ \frac{b_3}{b_2} \frac{2\nu c^2}{2\nu-1} \right) \frac{\exp\left( b_1 \frac{2\nu c^2}{2\nu-1}  \right)}{K^{1-\nu} c}.
\end{align*}

\section{Proof of Theorem~\ref{thm:stepdecay}}
\noindent Given a fixed value $T>0$ and $\alpha>1$, the step-decay step-size can be expressed equivalently as
\begin{eqnarray*}
	\gamma_k = \gamma_0 / \alpha^{\lfloor k/T \rfloor} = \gamma_0 /  \alpha^m := \gamma_m,
\end{eqnarray*}	
for $mT \leq k \leq (m+1)T - 1$ and $m=0,1,\ldots, M-1$. Therefore, 
\begin{eqnarray}\label{eq:inequality_step_decay}
     V_{k+1} \leq (1+ b_1 \gamma_m^2 )V_k - b_2 \gamma_m W_k + b_3 \gamma_m^2,
\end{eqnarray}
for $mT \leq k \leq (m+1)T - 1$ and $m=0,1,\ldots, M-1$. 
By summing \eqref{eq:inequality_step_decay} over $k=mT,mT+1,\ldots,(m+1)T-1$, and by the fact that $1+ b_1\gamma_k^2 \geq 1$, 
\begin{eqnarray*}
	V_{(m+1)T} 
 &\leq & (1+ b_1\gamma_m^2)^{T} V_{mT} - b_2\gamma_m \sum_{j=mT}^{(m+1)T-1} W_j  \\
 && + b_3 \gamma_m^2 \sum_{j=0}^{T-1} (1+b_1\gamma_m^2)^j, 
\end{eqnarray*}	
for $m=0,1,\ldots, M-1$. Next, since 
\begin{eqnarray*}
	\sum_{j=0}^{T-1} (1+b_1\gamma_m^2)^j  =  \frac{(1+b_1\gamma_m^2)^{T} - 1}{1+b_1\gamma_m^2 -1 } \leq   \frac{(1+b_1\gamma_m^2)^{T} }{b_1\gamma_m^2},
\end{eqnarray*}	
we have
 \begin{eqnarray*}
 	V_{(m+1)T} 
  &\leq&  (1+ b_1\gamma_m^2)^{T} V_{mT} - b_2\gamma_m \sum_{j=mT}^{(m+1)T-1} W_j  \\
  && + \frac{b_3}{b_1} (1+b_1\gamma_m^2)^{T},
 \end{eqnarray*}	
 for $m=0,1,\ldots, M-1$. 
Next, by re-arranging the terms, 
\begin{align*}
 \sum_{j=mT}^{(m+1)T-1} W_j  
% &\leq \frac{ (1+ b_1\gamma_m^2)^{T} }{b_2\gamma_m} V_{mT} - \frac{1}{b_2\gamma_m} V_{(m+1)T} + \frac{b_3}{b_2b_1} \frac{(1+ b_1\gamma_m^2)^{T}}{\gamma_m} \\
 & \leq \frac{V_{mT} - V_{(m+1)T}}{b_2\gamma_m} + \frac{(1+ b_1\gamma_m^2)^{T}-1}{b_2\gamma_m} V_{mT} \\
 &\hspace{0.5cm}+ \frac{b_3}{b_2b_1} \frac{(1+ b_1\gamma_m^2)^{T}}{\gamma_m} \\
 & \leq \frac{V_{mT} - V_{(m+1)T}}{b_2\gamma_m}  + \frac{(1+ b_1\gamma_m^2)^{T}}{b_2\gamma_m} V_{mT} \\
 & \hspace{0.5cm}+ \frac{b_3}{b_2b_1} \frac{(1+ b_1\gamma_m^2)^{T}}{\gamma_m},
\end{align*}	
for $m=0,1,\ldots, M-1$. 
Therefore, 
\begin{align*}
& \frac{1}{MT}\sum_{m=0}^{M-1} \sum_{j=mT}^{(m+1)T-1} W_j  \\
& \leq  \frac{V_{0} -  V_{MT}}{\tilde \Gamma MT} + \frac{A}{\tilde \Gamma } \frac{1}{MT} \sum_{m=0}^{M-1} V_{mT} + \frac{b_3 A}{b_1 \tilde \Gamma} \frac{1}{T}, 
\end{align*}	
where $\tilde \Gamma = b_2 \min_{m} \gamma_m$ and 
$A = \max_{m}\left(  (1+ b_1\gamma_m^2)^{T}  \right)$. 

If $0 \leq V_k \leq R$ for some positive constant $R$ and for all $k$, then 
\begin{align*}
 \frac{1}{MT}\sum_{m=0}^{M-1} \sum_{j=mT}^{(m+1)T-1} W_j  
 \leq&  \Gamma_1\frac{R}{MT} +  \frac{\Gamma_2}{T},
\end{align*}	
where $\Gamma_1 = \frac{1}{b_2 \min_{m} \gamma_m}$, $\Gamma_2 = C\frac{A}{ \min_{m} \gamma_m}$ and 
$C=(R+b_3/b_1)/b_2$. Since  
\begin{eqnarray*}
	(1+ b_1\gamma_m^2)^{T}
	& = & \exp(T \ln(1+b_1\gamma_m^2))  
	\leq  \exp(T b_1 \gamma_m^2) \\
	& = & \exp(b_1 \gamma_0^2 T / \alpha^{2m}), \quad \text{and} \\ 
        \min_{m} \gamma_m & = & \gamma_0 \min_m (1/\alpha^m)  
        \geq   \gamma_0/\alpha^M
\end{eqnarray*}	
we have 
\begin{align*}
 \frac{1}{MT}\sum_{m=0}^{M-1} \sum_{j=mT}^{(m+1)T-1} W_j  
 &\leq&  \frac{\alpha^M }{b_2 \gamma_0}\frac{R}{MT} + C\frac{ \alpha^M \bar A}{ \gamma_0} \frac{1}{T},
\end{align*}	
where $\bar A = \max_{m}\left( \exp(b_1 \gamma_0^2 T/  \alpha^{2m}) \right)$.

%\begin{eqnarray*}
%	\frac{1}{MT}\sum_{m=0}^{M-1} \sum_{j=mT}^{(m+1)T-1} W_j  \leq \max_{m}\left( \frac{ \exp(b_1 \gamma_0^2 T/  \alpha^{2m}) }{\gamma_m} \right) \frac{V_{0} -  V_{MT}}{b_2 MT} + \frac{b_3}{b_2b_1} \frac{1}{M}  \max_{m}\left( \frac{ \exp(b_1 \gamma_0^2 T / \alpha^{2m}) }{\gamma_m} \right).
%\end{eqnarray*}	

If $T = 2K/\log_{\alpha} K$ and $M=\log_{\alpha} K/2$, then  
\begin{eqnarray*}
	\bar A
	 \leq   \exp(b_1 \gamma_0^2 T / \alpha^{2M})  
	% & = & \exp \left(b_1 \gamma_0^2 \frac{2K}{\log_{\alpha} K} / \alpha^{\log_{\alpha} K} \right)\\
	 = \exp \left( 2b_1 \gamma_0^2 \frac{1}{\log_{\alpha} K} \right) \leq  B, 
\end{eqnarray*}	
where $B = \exp \left( 2b_1 \gamma_0^2 \frac{1}{\min(\log_{\alpha} 2, 1)} \right)$.
Hence, 
\begin{align*}
	 \frac{1}{MT}\sum_{m=0}^{M-1} \sum_{j=mT}^{(m+1)T-1} W_j &\leq  \frac{\alpha^M }{b_2 \gamma_0}\frac{R}{MT} + C\frac{ \alpha^M B}{ \gamma_0} \frac{1}{T}.
\end{align*}	
By the fact that  $\alpha^M  =  \alpha^{\log_{\alpha} K /2 } = \sqrt{K}$,
\begin{align*}
	 \frac{1}{MT}\sum_{m=0}^{M-1} \sum_{j=mT}^{(m+1)T-1} W_j &\leq  \frac{\sqrt{K} }{b_2 \gamma_0}\frac{R}{MT} + C\frac{ \sqrt{K} B}{ \gamma_0} \frac{1}{T}.
\end{align*}
Next, by the fact that $MT = K$ and $T=2K/\log_{\alpha} K$,
\begin{align*}
	 \frac{1}{MT}\sum_{m=0}^{M-1} \sum_{j=mT}^{(m+1)T-1} W_j &\leq  \frac{1}{b_2 \gamma_0}\frac{R}{\sqrt{K}} + C\frac{  B}{ \gamma_0} \frac{\log_{\alpha}(K)}{2\sqrt{K}}.
\end{align*}
Finally, since 
\begin{eqnarray*}
    \mathop{\min}_{0\leq k \leq K-1} W_k
	  \leq   \frac{1}{K}\sum_{k=0}^{K-1} W_k =  \frac{1}{MT}\sum_{m=0}^{M-1} \sum_{j=mT}^{(m+1)T-1} W_j,
\end{eqnarray*}
 we complete the proof.

\section{Other Applications for Convergence Theorems}
\noindent We can apply our convergence theorems to establish convergence results for stochastic optimization algorithms on non-convex problems by characterizing \eqref{eqn:mainIneq}. 
For instance, stochastic gradient descent according to (47) in \cite{khaled2022better} satisfies  \eqref{eqn:mainIneq} with $V_k= \mathbf{E}[f(x_k)-f^{\inf}]$, $W_k = \mathbf{E}\| \nabla f(x_k)\|^2$, $b_1=L A$, $b_2 = 1/2$, $b_3 = L C/2$, while byzantine stochastic gradient descent according to (3.31) in \cite{roberts2022simplified} satisfies \eqref{eqn:mainIneq} with $V_k= \mathbf{E}[f(x_k)-f^{\inf}]$, $W_k = \mathbf{E}\| \nabla f(x_k)\|^2$, $b_1=L A^\prime$, $b_2 = (1-\sin(\alpha))/2$, $b_3 = L C^\prime/2$.

\section{Proof of Proposition~\ref{prop:fedavg}}
\noindent	Algorithm~\ref{alg:fp_fl} with $\mathcal{T}_{\gamma F}(x)=x-\gamma \nabla F(x)$ is FedAvg, which can be described equivalently in Algorithm~\ref{alg:fedavg}. 
	The update for Algorithm~\ref{alg:fedavg} with $\gamma^k = \alpha^k/T$ can be written as: 
	\begin{align}\label{eq:Fedavg_equi}
		x^{k+1} = x^k - \frac{\alpha^k}{nT} \sum_{i=1}^n \sum_{t=0}^{T-1} \nabla F_i(x_i^{k,t};\xi_i^{k,t}).
	\end{align}	
	Also, from Algorithm~\ref{alg:fedavg}, we can show easily that 
	\begin{align}\label{Fedavg_local_and_global}
		x^k - x_i^{k,t} = \gamma^k \sum_{l=0}^{t-1} \nabla F_i(x_i^{k,l};\xi_i^{k,l}).
	\end{align}	
	\begin{algorithm}
		\caption{ {FedAvg} }\label{alg:fedavg}
		\begin{algorithmic}
			\State \textbf{Input:} The number of iterations $K,T$, the step-size $\gamma^k > 0$, and the initial point $x^0\in\mathbb{R}^d$.
			\For{$k=0,1,\ldots,K-1$}
			\State The server broadcasts $x^k$ to every worker node 
			\For{every worker $i=1,\ldots,n$}
			\State Set $x_i^{k,0}=x^k$
			\For{$t=0,1,\ldots,T-1$}
			\State Compute $ x_i^{k,t+1}= x_i^{k,t} - \gamma^k \nabla F_i(x_i^{k,t}; \xi_i^{k,t})$
			\EndFor 
			\State Send $x_i^{k,T}$ to the server 	
			\EndFor 
			\State The server updates $x^{k+1} = \frac{1}{n}\sum_{i=1}^n x_i^{k,t+1}$	
			\EndFor
		\end{algorithmic}
	\end{algorithm}
	Before deriving the result, we present one useful lemma: 
	\begin{lemma}\label{lemma:Fedavg_trick}
		Consider Problem~\eqref{eq:Problem} where Assumptions~\ref{assum:L_Lipschitz} and~\ref{assum:variance_stoc} hold.
		Then, the iterates $\{x^k\}$ generated by Algorithm~\ref{alg:fedavg}	with $\gamma^k \leq 1/(\sqrt{6} T L)$ satisfy
		\begin{align}
			\sum_{l=0}^{T-1}	\mathbf{E}\| x^k - x_i^{k,l} \|^2 
			 \leq  \Gamma^k T^3 \mathbf{E} \| \nabla f_i(x^k) \|^2   + \Gamma^k T^3 \sigma^2,
   \label{eq:Fedavg_trick}
		\end{align}	
  where $\Gamma^k = 6(\gamma^k)^2$
	\end{lemma}
	\begin{proof}
		From the definition of the Euclidean norm, 
		\begin{eqnarray*}
			\| x^k - x_i^{k,t} \|^2 
			& \overset{\eqref{Fedavg_local_and_global}}{=}& (\gamma^k)^2 \left\|  \sum_{l=0}^{t-1} \nabla F_i(x_i^{k,l};\xi_i^{k,l}) \right\|^2 \\ 
			& \overset{\eqref{eq:Triangle_sq}}{\leq}& (\gamma^k)^2 t \sum_{l=0}^{t-1}  \| \nabla F_i(x_i^{k,l};\xi_i^{k,l})\|^2.
		\end{eqnarray*}		
		Since $t\leq T$, we have 
		\begin{eqnarray*}
			\| x^k - x_i^{k,t} \|^2 
			& \leq & (\gamma^k)^2 T \sum_{l=0}^{T-1}  \| \nabla F_i(x_i^{k,l};\xi_i^{k,l})\|^2. %\\ 
			%& \overset{\eqref{eq:Triangle_sq}}{\leq}& 3  (\gamma^k)^2 T^2 \| \nabla f_i(x^k) \|^2 + 3  (\gamma^k)^2 T \sum_{l=0}^{T-1}  \|  \nabla f_i(x^k) -  \nabla f_i(x_i^{k,l})\|^2 \\
			%&& + 3  (\gamma^k)^2 T \sum_{l=0}^{T-1}  \|  \nabla F_i(x_i^{k,l};\xi_i^{k,l}) -  \nabla f_i(x_i^{k,l})\|^2 \\ 
			%	& \overset{\eqref{eq:L_Lipschitz}}{\leq}& 3  (\gamma^k)^2 T^2 \| \nabla f_i(x^k) \|^2 + 3  (\gamma^k)^2 L^2 T \sum_{l=0}^{T-1}  \| x^k -  x_i^{k,l}\|^2 \\
			%	&& + 3  (\gamma^k)^2 T \sum_{l=0}^{T-1}  \|  \nabla F_i(x_i^{k,l};\xi_i^{k,l}) -  \nabla f_i(x_i^{k,l})\|^2. 
		\end{eqnarray*}		
		
		Next, by \eqref{eq:Triangle_sq} with $n=3$, $x_1 = \nabla f_i(x^k)$, $x_2 = \nabla f_i(x^k)-\nabla f_i(x_i^{k,l})$, and $x_3 = \nabla F_i(x_i^{k,l};\xi_i^{k,l}) -  \nabla f_i(x_i^{k,l})$ and then by \eqref{eq:L_Lipschitz}, we get  
		\begin{align*}
			& \| x^k - x_i^{k,t} \|^2 
			%& \leq & (\gamma^k)^2 T \sum_{l=0}^{T-1}  \| \nabla F_i(x_i^{k,l};\xi_i^{k,l})\|^2 \\ 
	%		& \leq & 3  (\gamma^k)^2 T^2 \| \nabla f_i(x^k) \|^2 \\
	%		&& + 3  (\gamma^k)^2 T \sum_{l=0}^{T-1}  \|  \nabla f_i(x^k) -  \nabla f_i(x_i^{k,l})\|^2 \\
	%		&& + 3  (\gamma^k)^2 T \sum_{l=0}^{T-1}  B_i^{k,l}  \\ 
			 \leq  3  (\gamma^k)^2 T^2 \| \nabla f_i(x^k) \|^2 \\
			& \hspace{0.5cm}+ 3  (\gamma^k)^2 L^2 T \sum_{l=0}^{T-1}  \| x^k -  x_i^{k,l}\|^2  + 3  (\gamma^k)^2 T \sum_{l=0}^{T-1}  B_i^{k,l},
		\end{align*}		
		where $B_i^{k,l}  = \|  \nabla F_i(x_i^{k,l};\xi_i^{k,l}) -  \nabla f_i(x_i^{k,l})\|^2$.
		Therefore, 
		\begin{align*}
			& \sum_{l=0}^{T-1}	\| x^k - x_i^{k,l} \|^2 
			 \leq  3  (\gamma^k)^2 T^3 \| \nabla f_i(x^k) \|^2 \\
			& \hspace{0.5cm}+ 3  (\gamma^k)^2 L^2 T^2 \sum_{l=0}^{T-1}  \| x^k -  x_i^{k,l}\|^2  + 3  (\gamma^k)^2 T^2 \sum_{l=0}^{T-1} B_i^{k,l}.
		\end{align*}		
		
		Finally, if $\gamma^k \leq 1/(\sqrt{6} TL)$, then by taking the expectation and by \eqref{eq:variance_stoc}, we complete the proof.
%		\begin{eqnarray*}
%			\sum_{l=0}^{T-1}	\| x^k - x_i^{k,l} \|^2 
%			& \leq & 6 (\gamma^k)^2 T^3 \| \nabla f_i(x^k) \|^2  \\
%			&& + 6 (\gamma^k)^2 T^2 \sum_{l=0}^{T-1}  B_i^{k,l}.
%		\end{eqnarray*}		
%
%		\begin{eqnarray*}
%			\sum_{l=0}^{T-1}	\mathbf{E}\| x^k - x_i^{k,l} \|^2 
%			& \leq & 6 (\gamma^k)^2 T^3 \mathbf{E} \| \nabla f_i(x^k) \|^2  \\
%			&& + 6 (\gamma^k)^2 T^3 \sigma^2. 
%		\end{eqnarray*}		
	\end{proof}

	Now, we prove the main result. 	
	From Assumption~\ref{assum:L_Lipschitz}, we can prove that $f(x)$ has also $L$-Lipschitz continuous gradient. Let $f^{\inf}$ is the lower bound for $f(x)$. 
	If $\gamma^k = \alpha^k/T$, then by \eqref{eq:L_Lipschitz_ineq} and \eqref{eq:Fedavg_equi},
	\begin{eqnarray*}
		r^{k+1}
		\leq r^k  - \alpha^k \left\langle \nabla f(x^k),   v^k \right\rangle + \frac{L (\alpha^k)^2}{2} \left\|  v^k  \right\|^2,
	\end{eqnarray*}	
	where $r^k = f(x^k) - f^{\inf}$ and also 
 $v^k  = \frac{1}{nT} \sum_{i=1}^n \sum_{t=0}^{T-1} \nabla F_i(x_i^{k,t};\xi_i^{k,t})$.
	By taking the expectation, 
	\begin{eqnarray*}
		V^{k+1}
		\leq   V^k - \alpha^k \mathbf{E}\left\langle \nabla f(x^k),    \bar v^k\right\rangle + \frac{L (\alpha^k)^2}{2} \mathbf{E}\left\|  v^k \right\|^2,
	\end{eqnarray*}	
	where $V^k =\mathbf{E}[r^k]$ and $\bar v^k = \frac{1}{nT} \sum_{i=1}^n \sum_{t=0}^{T-1} \nabla f_i(x_i^{k,t})$.
	Next, since 
	\begin{eqnarray*}
		- \alpha^k \mathbf{E}\left\langle \nabla f(x^k),   \bar v^k \right\rangle 
		& \overset{\eqref{eq:Exact_sq}}{=} & - \frac{\alpha^k}{2}\mathbf{E}\|  \nabla f(x^k) \|^2 - \frac{\alpha^k}{2}\mathbf{E}\left\| \bar v^k \right\|^2\\
		&& + \frac{\alpha^k}{2} \mathbf{E}\left\|  \nabla f(x^k) - \bar v^k \right\|^2, \quad \text{and} \\
		\mathbf{E}\left\|  v^k  \right\|^2
		& \overset{\eqref{eq:Triangle_sq} }{\leq} & 2 	\mathbf{E}\left\|  \frac{1}{nT} \sum_{i=1}^n \sum_{t=0}^{T-1} B_i^{k,t}\right\|^2 \\
		&& + 2 \mathbf{E}\left\| \bar v^k \right\|^2,
	\end{eqnarray*}	
	where $B_i^{k,t} = \nabla F_i(x_i^{k,t};\xi_i^{k,t}) - \nabla f_i(x_i^{k,t})$,
	we get  
	\begin{eqnarray*}
		V^{k+1}
		%&\leq&  \mathbf{E}[f(x^k) - f^{\inf}] - \alpha^k \mathbf{E}\left\langle \nabla f(x^k),    \frac{1}{nT} \sum_{i=1}^n \sum_{t=0}^{T-1} \nabla f_i(x_i^{k,t})\right\rangle \\
		%&& + \frac{L (\alpha^k)^2}{2} \mathbf{E}\left\|  \frac{1}{nT} \sum_{i=1}^n \sum_{t=0}^{T-1} \nabla F_i(x_i^{k,t};\xi_i^{k,t}) \right\|^2 \\
		& \leq &  V^k - \frac{\alpha^k}{2} \mathbf{E}\| \nabla f(x^k) \|^2 - \beta^k \mathbf{E}\left\|  \bar v^k  \right\|^2 \\
		&& + \frac{\alpha^k}{2} A_1^k  + L (\alpha^k)^2 A_2^k,
	\end{eqnarray*}	
	where $\beta^k = {\alpha^k}/{2} - L (\alpha^k)^2$, $A_1^k = \mathbf{E}\left\| \nabla f(x^k) - \frac{1}{nT} \sum_{i=1}^n \sum_{t=0}^{T-1} \nabla f_i(x_i^{k,t}) \right\|^2$, and also $A_2^k =\mathbf{E}\left\|  \frac{1}{nT} \sum_{i=1}^n \sum_{t=0}^{T-1} B_i^{k,t} \right\|^2$.
	
	If $\alpha^k \leq 1/(\sqrt{6}L)$, then $\alpha^k \leq 1/(2L)$. Hence, $\beta^k \geq 0$ and 
	\begin{eqnarray*}
		V^{k+1}
		& \leq & V^k - \frac{\alpha^k}{2} \mathbf{E}\| \nabla f(x^k) \|^2  
		%&& - \left( \frac{\alpha^k}{2} - L (\alpha^k)^2 \right)\mathbf{E}\left\|  \frac{1}{nT} \sum_{i=1}^n \sum_{t=0}^{T-1} \nabla f_i(x_i^{k,t}) \right\|^2 \\
		+ \frac{\alpha^k}{2}A^k_1 + L (\alpha^k)^2A_2^k.
	\end{eqnarray*}	
	%where $A_1^k = \mathbf{E}\left\| \nabla f(x^k) - \frac{1}{nT} \sum_{i=1}^n \sum_{t=0}^{T-1} \nabla f_i(x_i^{k,t}) \right\|^2$ and $A_2^k =\mathbf{E}\left\|  \frac{1}{nT} \sum_{i=1}^n \sum_{t=0}^{T-1} B_i^{k,t} \right\|^2$.
	Next, since $\nabla f(x) = \frac{1}{nT}\sum_{i=1}^n\sum_{t=0}^{T-1} \nabla f_i(x)$ and since 
	\begin{eqnarray*}
		A_1^k
		%& \overset{\eqref{eq:Triangle_sq}}{\leq}& \frac{1}{nT}\sum_{i=1}^n \sum_{t=0}^{T-1} \mathbf{E}\| \nabla f_i(x^k) -  \nabla f_i(x_i^{k,t})  \|^2 \\ 
		&\overset{\eqref{eq:Triangle_sq}+ \eqref{eq:L_Lipschitz}}{\leq}& \frac{L^2}{nT}\sum_{i=1}^n \sum_{t=0}^{T-1} \mathbf{E}\| x^k -  x_i^{k,t} \|^2, \quad \text{and} \\ 
		A_2^k  & \overset{\eqref{eq:Triangle_sq}}{\leq}&   \frac{1}{nT} \sum_{i=1}^n \sum_{t=0}^{T-1} \mathbf{E} \| B_i^{k,t} \|^2 \overset{\eqref{eq:variance_stoc}}{\leq}  \sigma^2,
	\end{eqnarray*}	 
	we have 
	%
	%By \eqref{eq:Triangle_sq} and \eqref{eq:variance_stoc},
	\begin{eqnarray*}
		V^{k+1}
		%  & \leq  & \mathbf{E}[f(x^k) - f^{\inf}] - \frac{\alpha^k}{2} \mathbf{E}\| \nabla f(x^k) \|^2 \\ 
		%	&& + \frac{\alpha^k}{2} \frac{1}{nT}  \sum_{i=1}^n \sum_{t=0}^{T-1}  \mathbf{E}\left\| \nabla f_i(x^k) - \nabla f_i(x_i^{k,t}) \right\|^2  + L (\alpha^k)^2 \sigma^2 \\ 
		& \leq  & V^k - \frac{\alpha^k}{2} \mathbf{E}\| \nabla f(x^k) \|^2 + L (\alpha^k)^2 \sigma^2 \\ 
		&& + \frac{\alpha^k L^2 }{2} \frac{1}{nT}  \sum_{i=1}^n \sum_{t=0}^{T-1}  \mathbf{E}\left\| x^k - x_i^{k,t} \right\|^2.
	\end{eqnarray*}	
	Next, from Lemma~\ref{lemma:Fedavg_trick} with $\gamma^k=\alpha^k/T$ and $\alpha^k \leq 1/(\sqrt{6}L)$
	\begin{eqnarray*}
		V^{k+1}
		& \leq & V^k - \frac{\alpha^k}{2} \mathbf{E}\| \nabla f(x^k) \|^2 + L (\alpha^k)^2 \sigma^2  \\ 
		&&   +  \frac{3(\alpha^k)^3 L^2 T}{n} \sum_{i=1}^n  \mathbf{E} \| \nabla f_i(x^k) \|^2 + 3(\alpha^k)^3 L^2 T \sigma^2 .
		%	& \overset{\eqref{eq:L_Lipschitz_trick}}{\leq} & (1+6(\alpha^k)^3 L^3 T) \mathbf{E}[f(x^k) - f^{\inf}] - \frac{\alpha^k}{2} \mathbf{E}\| \nabla f(x^k) \|^2 \\ 
		%	&& + 6(\alpha^k)^3 L^3 T \Delta^{\inf}+ 3(\alpha^k)^3 L^2 T \sigma^2 + L (\alpha^k)^2 \sigma^2.
	\end{eqnarray*}	
	Next, denote $f_i^{\inf}$  as the lower-bound of each component function $f_i(x)$. Since
	\begin{eqnarray*}
		\frac{1}{n}\sum_{i=1}^n  \mathbf{E} \| \nabla f_i(x^k) \|^2
		&  \overset{\eqref{eq:L_Lipschitz_trick}}{\leq} & 	\frac{2L}{n}\sum_{i=1}^n  \mathbf{E} [f_i(x^k) - f_i^{\inf}] \\
		& = & 2L \mathbf{E}[f(x^k) - f^{\inf}] + 2L \Delta^{\inf},
	\end{eqnarray*}	
	where $\Delta^{\inf} = f^{\inf} - \frac{1}{n}\sum_{i=1}^n f_i^{\inf}$, we get
	\begin{eqnarray*}
		V^{k+1}
		%	& \leq & \mathbf{E}[f(x^k) - f^{\inf}] +  \frac{3(\alpha^k)^3 L^2 T}{n} \sum_{i=1}^n  \mathbf{E} \| \nabla f_i(x^k) \|^2 - \frac{\alpha^k}{2} \mathbf{E}\| \nabla f(x^k) \|^2 \\ 
		%	&& + 3(\alpha^k)^3 L^2 T \sigma^2 + L (\alpha^k)^2 \sigma^2 \\ 
		& \leq & (1+6(\alpha^k)^3 L^3 T) V^k - \frac{\alpha^k}{2} \mathbf{E}\| \nabla f(x^k) \|^2 \\ 
		&& + 6(\alpha^k)^3 L^3 T \Delta^{\inf}+ 3(\alpha^k)^3 L^2 T \sigma^2 + L (\alpha^k)^2 \sigma^2.
	\end{eqnarray*}	
	Finally, by the fact that $\alpha^k \leq 1/(\sqrt{6}L)$, 
	\begin{eqnarray*}
		V^{k+1}
		\leq  (1+c_1(\alpha^k)^2 ) V^k - \frac{\alpha^k}{2} \mathbf{E}\| \nabla f(x^k) \|^2 + (\alpha^k)^2 e, 
	\end{eqnarray*}	
	where $c_1 = \sqrt{6}L^2 T$ and $e=\sqrt{6} L^2 T \Delta^{\inf}+ (3/\sqrt{6}) L T \sigma^2 + L  \sigma^2.$

\section{Proof of Proposition~\ref{prop:FedProx}}
\noindent Recall from the definition and first-optimality condition of the proximal operator that
	\begin{align}\label{eq:prox_trick}
		p^k_i = x^k - \gamma^k \nabla F_i(p^k_i;\xi_i^k),
	\end{align}	
where $p^k_i:=\prox(x^k)$. Therefore, FedProx, Algorithm~\ref{alg:fp_fl} with $\mathcal{T}_{\gamma F}(x)=\textbf{prox}_{\gamma F}(x)$ and $T=1$,  can be expressed equivalently as: 
	\begin{align}\label{eq:FedProx_equi}
		x^{k+1} = x^k - \frac{\gamma^k}{n}\sum_{i=1}^n \nabla F_i(p^k_i;\xi_i^k).
	\end{align}	
	
	We begin by stating one useful lemma. 
	\begin{lemma}
		Consider Problem~\eqref{eq:Problem} where Assumptions~\ref{assum:L_Lipschitz} and~\ref{assum:variance_stoc} hold. Let $\gamma^k \leq 1/(\sqrt{6} L)$. Then, 
		%Then, the iterates $\{x^k\}$ generated by 	Algorithm~\ref{alg:fp_fl} with $\mathcal{T}_{\gamma F}(x)=\textrm{\bf prox}_{\gamma F}(x)$, with $T=1$ and  with $\gamma^k \leq 1/(\sqrt{6} L)$ satisfy
		\begin{align}
			\mathbf{E}\| x^k -  \prox(x^k) \|^2 \leq 6(\gamma^k)^2 \mathbf{E}\| \nabla f_i(x^k) \|^2  + 6 (\gamma^k)^2 \sigma^2.\label{eq:FedProx_trick}
		\end{align}	
	\end{lemma}
	\begin{proof}
		Define  $p^k_i:=\prox(x^k)$. 
		From the definition of the Euclidean norm, 
		\begin{eqnarray*}
			\| x^k -  p^k_i \|^2
	%		& \overset{\eqref{eq:prox_trick}}{=} & (\gamma^k)^2 \| \nabla F_i(p_i^k;\xi_i^k) \|^2 \\ 
			& \overset{\eqref{eq:prox_trick}+\eqref{eq:Triangle_sq}}{\leq}& 3 (\gamma^k)^2 \left( \| \nabla f_i(x^k) \|^2 + e_1^k + e_2^k\right).
		\end{eqnarray*}		
  where $e_1^k = \| \nabla f_i(x^k) - \nabla f_i(p_i^k) \|^2$ and $e_2^k = \|  \nabla F_i(p_i^k;\xi_i^k)  - \nabla f_i(p_i^k) \|^2$. 	Next, by taking the expectation, by \eqref{eq:variance_stoc} and by \eqref{eq:L_Lipschitz}, 
		\begin{eqnarray*}
			\mathbf{E}\| x^k -  p^k_i \|^2
			%	& \overset{\eqref{eq:variance_stoc}}{\leq}& 3 (\gamma^k)^2 \mathbf{E}\| \nabla f_i(x^k) \|^2 +  3 (\gamma^k)^2 \mathbf{E}\| \nabla f_i(x^k) - \nabla f_i(p_i^k) \|^2 +  3 (\gamma^k)^2 \sigma^2 \\
			& \leq & 3 (\gamma^k)^2 \mathbf{E}\| \nabla f_i(x^k) \|^2 \\
			&& +  3 (\gamma^k)^2 L^2 \mathbf{E}\| x^k - p_i^k \|^2 +  3 (\gamma^k)^2 \sigma^2.
		\end{eqnarray*}		
		
		Finally, if $\gamma \leq 1/(\sqrt{6} L)$, then by re-arranging the terms, we complete the proof. 
	\end{proof}

	Now, we prove the main result. From Assumption~\ref{assum:L_Lipschitz},  $f(x)$ has also $L$-Lipschitz continuous gradient. Let $f^{\inf}$ is the lower bound for $f(x)$. From \eqref{eq:L_Lipschitz_ineq} and \eqref{eq:FedProx_equi}, 
	\begin{eqnarray*}
		r^{k+1}
		&\leq&  r^k  - \gamma^k \left\langle \nabla f(x^k),  v^k\right\rangle  + \frac{L (\gamma^k)^2}{2} \left\|  v^k \right\|^2,
	\end{eqnarray*}
 where $r^k =  f(x^k) - f^{\inf}$ and  $v^k =  \frac{1}{n}\sum_{i=1}^n \nabla F_i(p^k_i;\xi_i^k)$. By taking the expectation,
	\begin{eqnarray*}
		V^{k+1} 
		&\leq&  V^k  - \gamma^k T_1^k  + \frac{L (\gamma^k)^2}{2} T_2^k,
	\end{eqnarray*}	
	where $V^k = \mathbf{E}[r^k]$, $T_1^k=\mathbf{E} \left\langle \nabla f(x^k),   \frac{1}{n}\sum_{i=1}^n \nabla f_i(p^k_i)\right\rangle$ and $T_2^k= \mathbf{E}\left\|  \frac{1}{n}\sum_{i=1}^n \nabla F_i(p^k_i;\xi_i^k) \right\|^2$.
	Since 
	\begin{eqnarray*}
		- \gamma^k T_1^k 
		& \overset{\eqref{eq:Exact_sq}}{=} & - \frac{\gamma^k}{2}\mathbf{E} \| \nabla f(x^k) \|^2 - \frac{\gamma^k}{2} \bar T_2^k\\
		&& +  \mathbf{E} \left\| \nabla f(x^k) -  \frac{1}{n}\sum_{i=1}^n \nabla f_i(p_i^k) \right\|^2 \\ 
%		&  \overset{\eqref{eq:Triangle_sq}}{\leq} &-  \frac{\gamma^k}{2}\mathbf{E} \| \nabla f(x^k) \|^2 - \frac{\gamma^k}{2} \bar T_2^k  \\
%		&& + \frac{1}{n}\sum_{i=1}^n  \mathbf{E} \left\| \nabla f_i(x^k) -   \nabla f_i(p_i^k) \right\|^2 \\ 
		& \overset{\eqref{eq:Triangle_sq}+\eqref{eq:L_Lipschitz}}{\leq} &- \frac{\gamma^k}{2}\mathbf{E} \| \nabla f(x^k) \|^2 - \frac{\gamma^k}{2} \bar T_2^k  \\
		&& + \frac{L^2}{n}\sum_{i=1}^n  \mathbf{E} \left\| x^k -   p_i^k \right\|^2,
	\end{eqnarray*}	
	and since 
	\begin{eqnarray*}
		T_2^k 
	%	& \overset{\eqref{eq:Triangle_sq}}{\leq}& 2 \bar T_2^k + 2 \mathbf{E}\left\| \frac{1}{n}\sum_{i=1}^n[\nabla F_i(p^k_i;\xi_i^k) - \nabla f_i(p_i^k)]  \right\|^2 \\ 
		& \overset{\eqref{eq:Triangle_sq}}{\leq}&  2 \bar T_2^k + \frac{2}{n} \sum_{i=1}^n \mathbf{E}\left\| \nabla F_i(p^k_i;\xi_i^k) - \nabla f_i(p_i^k) \right\|^2 \\
		& \overset{ \eqref{eq:variance_stoc} }{\leq} & 2 \bar T_2^k + 2\sigma^2,
	\end{eqnarray*}	
	where $\bar T_2^k = \mathbf{E}\left\|  \frac{1}{n}\sum_{i=1}^n \nabla f_i(p_i^k) \right\|^2$, we have 
	\begin{eqnarray*}
		V^{k+1}
		& \leq & V^k- \frac{\gamma^k}{2} \mathbf{E}\| \nabla f(x^k) \|^2 - \left( \frac{\gamma^k}{2} - L (\gamma^k)^2 \right)\bar T_2^k \\
		&& + L (\gamma^k)^2 \sigma^2  + \frac{L^2\gamma^k}{2n}\sum_{i=1}^n \mathbf{E}\left\| x^k - p_i^k \right\|^2.
	\end{eqnarray*}

	If $\gamma^k \leq {1}/{(\sqrt{6}L)}$, then $\gamma^k \leq {1}/{(2L)}$. Hence, $\frac{\gamma^k}{2} - L (\gamma^k)^2 \geq 0$ and 
	\begin{eqnarray*}
		V^{k+1}
	%	& \leq & V^k - \frac{\gamma^k}{2} \mathbf{E}\| \nabla f(x^k) \|^2+ L (\gamma^k)^2 \sigma^2  \\
	%	&&  + \frac{L^2\gamma^k}{2n}\sum_{i=1}^n \mathbf{E}\left\| x^k - p_i^k \right\|^2 \\ 
		& \overset{\eqref{eq:FedProx_trick}}{\leq} & V^k + 3 L^2 (\gamma^k)^3 \frac{1}{n}\sum_{i=1}^n \mathbf{E}\| \nabla f_i(x^k) \|^2    \\
		&&  - \frac{\gamma^k}{2} \mathbf{E}\| \nabla f(x^k) \|^2 + [L (\gamma^k)^2 + 3L^2 (\gamma^k)^3] \sigma^2. 
		%	& \overset{\eqref{eq:L_Lipschitz_trick}}{\leq} & (1+6L^3(\gamma^k)^3)\mathbf{E}[f(x^{k}) - f^{\inf}]  - \frac{\gamma^k}{2} \mathbf{E}\| \nabla f(x^k) \|^2 \\
		%	&& + 6L^3(\gamma^k)^3 \Delta^{\inf}  + [L (\gamma^k)^2 + 3L^2 (\gamma^k)^3] \sigma^2.
	\end{eqnarray*}	
	Next, denote $f_i^{\inf}$  as the lower-bound of each component function $f_i(x)$. Since
	\begin{eqnarray*}
		\frac{1}{n}\sum_{i=1}^n \mathbf{E}\| \nabla f_i(x^k) \|^2 
		& \overset{\eqref{eq:L_Lipschitz_trick} }{\leq} &  \frac{2L}{n}\sum_{i=1}^n \mathbf{E}[f_i(x_k)-f_i^{\inf}] \\
		& = &2L \mathbf{E}[f(x_k)-f^{\inf}] + 2L \Delta^{\inf},
	\end{eqnarray*}	
	where $ \Delta^{\inf} = f^{\inf} - \frac{1}{n} \sum_{i=1}^n f_i^{\inf}$, we get 
	\begin{eqnarray*}
		V^{k+1}
		& \leq & (1+6L^3(\gamma^k)^3)V^k - \frac{\gamma^k}{2} \mathbf{E}\| \nabla f(x^k) \|^2 \\
		&& + 6L^3(\gamma^k)^3 \Delta^{\inf}  + [L (\gamma^k)^2 + 3L^2 (\gamma^k)^3] \sigma^2.
	\end{eqnarray*}

	Finally, By the fact that $\gamma^k \leq {1}/{(\sqrt{6}L)}$, 
	\begin{eqnarray*}
		V^{k+1}
		& \leq & (1+\sqrt{6}L^2(\gamma^k)^2)V^k - \frac{\gamma^k}{2} \mathbf{E}\| \nabla f(x^k) \|^2 \\
		&& + \sqrt{6}L^2(\gamma^k)^2 \Delta^{\inf}  + L(\gamma^k)^2 [1 + 3/\sqrt{6}] \sigma^2.
	\end{eqnarray*}	
	We complete the proof.

\section{Proof of Proposition~\ref{prop:EF_fedavg}}
\noindent By setting $\mathcal{T}_{\gamma F}(x)=x-\gamma \nabla F(x)$ Algorithm~\ref{alg:EC_FL}  with $\mathcal{T}_{\gamma F}(x)=x-\gamma \nabla F(x)$  is error-feedback FedAvg, see Algorithm~\ref{alg:EC_fedavg} below.  
The update for Algorithm~\ref{alg:EC_fedavg} with $\gamma^k{=}\alpha^k/T$ can be expressed as:
\begin{eqnarray}
		z^{k+1}
%  & = & z^k + \frac{1}{n}\sum_{i=1}^n (x_i^{k,T} - x^k) \notag \\
  & \overset{\eqref{Fedavg_local_and_global}}{=} & z^k - \frac{\alpha^k}{nT} \sum_{i=1}^n \sum_{t=0}^{T-1} \nabla F_i(x_i^{k,t};\xi_i^{k,t}),\label{eq:EC_FedAvg_equi}
\end{eqnarray}	
where $z^k = x^k + \sum_{i=1}^n e_i^k/n$. Also, from Algorithm~\ref{alg:EC_fedavg}, we can prove  \eqref{Fedavg_local_and_global}.

\begin{algorithm}
	\caption{Error-feedback FedAvg }\label{alg:EC_fedavg}
	\begin{algorithmic}
		\State \textbf{Input:} The number of iterations $K,T$, the step-size $\gamma^k > 0$, the initial point $x^0\in\mathbb{R}^d$, and $e_i^0=0$ for all $i$.
		\For{$k=0,1,\ldots,K-1$}
		\State The server broadcasts $x^k$ to every worker node 
		\For{every worker $i=1,\ldots,n$}
		\State Set $x_i^{k,0}=x^k$
		\For{$t=0,1,\ldots,T-1$}
		\State Compute $ x_i^{k,t+1}= x_i^{k,t} - \gamma^k \nabla F_i(x_i^{k,t}; \xi_i^{k,t})$
		\EndFor 
		\State Send $Q(x_i^{k,T} - x^k + e_i^k)$ to the server
		\State Update $e_i^{k+1} =x_i^{k,T} - x^k + e_i^k - Q(x_i^{k,T} - x^k + e_i^k)$ 	
		\EndFor 
		\State The server updates $x^{k+1} = x^k + \frac{1}{n}\sum_{i=1}^n Q(x_i^{k,T} - x^k + e_i^k)$	
		\EndFor
	\end{algorithmic}
\end{algorithm}

From Assumption~\ref{assum:L_Lipschitz},  $f(x)$ has also $L$-Lipschitz continuous gradient and let $f^{\inf}$ is the lower bound for $f(x)$. From \eqref{eq:L_Lipschitz_ineq} and \eqref{eq:EC_FedAvg_equi}, 
	\begin{eqnarray*}
	r^{k+1} \leq  r^k   - \alpha^k \left\langle \nabla f(z^k), g^k \right\rangle  + \frac{L (\alpha^k)^2}{2} \left\| g^k \right\|^2,
\end{eqnarray*}	
where $r^k=f(z^k) - f^{\inf}$ and $g^k = \frac{1}{nT}\sum_{i=1}^n \sum_{t=0}^{T-1} \nabla F_i(x_i^{k,t};\xi_i^{k,t})$.
Next, by taking the expectation, and by using the unbiased property of the stochastic gradient and the fact that $\nabla f(x) = (1/n)\sum_{i=1}^n \nabla f_i(x)$,
\begin{eqnarray*}
	V^{k+1} 
	%&\leq& V^k - \alpha^k \mathbf{E} \left\langle \nabla f(z^k) , \bar g^k  \right\rangle  +   \frac{L (\alpha^k)^2}{2} \mathbf{E} \left\| g^k \right\|^2 \\
	& \leq & V^k - \alpha^k T_1 + \alpha^k T_2 +  \frac{L (\alpha^k)^2}{2} \mathbf{E} \left\|  g^k \right\|^2, 
\end{eqnarray*}	
where $T_1 = \mathbf{E} \left\langle \nabla f(z^k) , \nabla f(x^k)  \right\rangle$, $T_2= \mathbf{E} \left\langle \nabla f(z^k) , \nabla f(x^k) - \bar g^k \right\rangle$, and also  $\bar g^k = \frac{1}{nT}\sum_{i=1}^n \sum_{t=0}^{T-1} \nabla f_i(x_i^{k,t})$. 
Since 
\begin{eqnarray*}
	 - \alpha^k T_1  & \overset{ \eqref{eq:Exact_sq}}{=} & - \frac{\alpha^k}{2} \mathbf{E} \|  \nabla f(z^k)\|^2 -   \frac{\alpha^k}{2} \mathbf{E} \|  \nabla f(x^k)\|^2  \\
	 && + \frac{\alpha^k}{2} \mathbf{E} \|  \nabla f(z^k) - \nabla f(x^k)\|^2, \quad \text{and} \\ 
	 \alpha^k T_2 & \overset{\eqref{eq:InExact_sq} + \eqref{eq:Triangle_sq}}{\leq} &  \frac{\alpha^k}{2} \frac{1}{nT} \sum_{i=1}^n \sum_{t=0}^{T-1} \mathbf{E} \| \nabla f_i(x^k) - \nabla f_i(x_i^{k,t})\|^2  \\
	 &&  + \frac{\alpha^k}{2} \mathbf{E}\| \nabla f(z^k) \|^2,
\end{eqnarray*}	
we have 
\begin{eqnarray*}
	V^{k+1} 
	&\leq& V^k - \frac{\alpha^k}{2} \mathbf{E} \| \nabla f(x^k)\|^2 +   \frac{L (\alpha^k)^2}{2} \mathbf{E} \left\| g^k \right\|^2  \\
	&& + \frac{\alpha^k}{2} \mathbf{E} \|  \nabla f(z^k) - \nabla f(x^k)\|^2 \\
	&& + \frac{\alpha^k}{2} \frac{1}{nT} \sum_{i=1}^n \sum_{t=0}^{T-1} \mathbf{E} \| \nabla f_i(x^k) - \nabla f_i(x_i^{k,t})\|^2. 
\end{eqnarray*}	
Next, since 
\begin{align*}
	  & \frac{L (\alpha^k)^2}{2} \mathbf{E} \left\| g^k \right\|^2 \\
	  & \overset{ \eqref{eq:Triangle_sq}}{\leq}  \frac{3L (\alpha^k)^2}{2} \mathbf{E} \left\| g^k - \bar g^k  \right\|^2 +  \frac{3L (\alpha^k)^2}{2} \mathbf{E} \left\| \bar g^k - \nabla f(x^k)  \right\|^2  \\
	  & \hspace{0.5cm}+  \frac{3L (\alpha^k)^2}{2} \mathbf{E} \left\| \nabla f(x^k)  \right\|^2 \\
	 % & \overset{ \eqref{eq:Triangle_sq}}{\leq }\frac{3L (\alpha^k)^2}{2}  \frac{1}{nT} \sum_{i=1}^n \sum_{t=0}^{T-1}\mathbf{E} \|  \nabla F_i(x_i^{k,t};\xi_i^{k,t}) -  \nabla f_i(x_i^{k,t})\|^2 \\
	  %&  \hspace{0.5cm}+ \frac{3L (\alpha^k)^2}{2}  \frac{1}{nT} \sum_{i=1}^n \sum_{t=0}^{T-1}\mathbf{E} \|    \nabla f_i(x_i^{k,t}) - \nabla f_i(x^k)\|^2 \\
	  %&  \hspace{0.5cm}+  \frac{3L (\alpha^k)^2}{2} \mathbf{E} \left\| \nabla f(x^k)  \right\|^2 \\ 
	  & \overset{\eqref{eq:Triangle_sq} + \eqref{eq:variance_stoc}}{\leq}\frac{3L (\alpha^k)^2}{2} \sigma^2 +  \frac{3L (\alpha^k)^2}{2} \mathbf{E} \left\| \nabla f(x^k)  \right\|^2  \\
	  &  \hspace{0.5cm}+  \frac{3L (\alpha^k)^2}{2}  \frac{1}{nT} \sum_{i=1}^n \sum_{t=0}^{T-1}\mathbf{E} \|    \nabla f_i(x_i^{k,t}) - \nabla f_i(x^k)\|^2,
\end{align*}	
we have 
\begin{align*}
	V^{k+1} 
	\leq& V^k - \frac{\alpha^k (1-3L\alpha^k)}{2} \mathbf{E} \| \nabla f(x^k)\|^2   \\
	& + \frac{\alpha^k}{2} \mathbf{E} \|  \nabla f(z^k) - \nabla f(x^k)\|^2 + \frac{3L(\alpha^k)^2}{2}\sigma^2 \\
	& + \frac{\alpha^k\beta^k}{2} \frac{1}{nT} \sum_{i=1}^n\sum_{t=0}^{T-1} \mathbf{E} \| \nabla f_i(x^k) - \nabla f_i(x_i^{k,t})\|^2,
\end{align*}	
where $\beta^k = 1+3L\alpha^k$.

If $\alpha^k \leq 1/(6L)$, then 
\begin{align*}
	V^{k+1} 
	\leq& V^k - \frac{\alpha^k}{4} \mathbf{E} \| \nabla f(x^k)\|^2  \\
	& + \frac{\alpha^k}{2} \mathbf{E} \|  \nabla f(z^k) - \nabla f(x^k)\|^2 + \frac{3L(\alpha^k)^2}{2}\sigma^2 \\
	& + \frac{3\alpha^k}{4} \frac{1}{nT} \sum_{i=1}^n\sum_{t=0}^{T-1} \mathbf{E} \| \nabla f_i(x^k) - \nabla f_i(x_i^{k,t})\|^2.
\end{align*}	
By Assumption~\ref{assum:L_Lipschitz},
\begin{align*}
	V^{k+1} 
	\leq& V^k - \frac{\alpha^k}{4} \mathbf{E} \| \nabla f(x^k)\|^2   + \frac{L^2 \alpha^k}{2} \mathbf{E} \|  z^k - x^k \|^2 \\
	& + \frac{3 L^2 \alpha^k}{4} \frac{1}{nT} \sum_{i=1}^n\sum_{t=0}^{T-1} \mathbf{E} \| x^k - x_i^{k,t}\|^2  + \frac{3L(\alpha^k)^2}{2}\sigma^2.
\end{align*}	
By the fact that  $z^k = x^k + \frac{1}{n}\sum_{i=1}^n e_i^k$ and by \eqref{eq:Triangle_sq},
\begin{align*}
	V^{k+1} 
	\leq& V^k - \frac{\alpha^k}{4} \mathbf{E} \| \nabla f(x^k)\|^2   + \frac{L^2 \alpha^k}{2} \frac{1}{n} \sum_{i=1}^n \mathbf{E} \|  e_i^k \|^2 \\
	& + \frac{3 L^2 \alpha^k}{4} \frac{1}{nT} \sum_{i=1}^n\sum_{t=0}^{T-1} \mathbf{E} \| x^k - x_i^{k,t}\|^2  + \frac{3L(\alpha^k)^2}{2}\sigma^2.
\end{align*}	
Next, we bound $\sum_{t=0}^{T-1}	\mathbf{E}\| x^k - x_i^{k,t} \|^2$ to complete the convergence bound. By the fact that  $\gamma^k=\alpha^k/T$, $\alpha^k \leq 1/(6L) \leq 1/(\sqrt{6}L)$, and by  \eqref{eq:Fedavg_trick}, \eqref{eq:Triangle_sq} and \eqref{eq:L_Lipschitz}
\begin{eqnarray}
		&& \sum_{t=0}^{T-1}	\mathbf{E}\| x^k - x_i^{k,t} \|^2 \notag \\
	%	& \overset{ \eqref{eq:Fedavg_trick} }{\leq} &	6 (\alpha^k)^2 T \mathbf{E} \| \nabla f_i(x^k) \|^2  \notag  \\
%		&&  + 6 (\alpha^k)^2 T \sigma^2 \notag \\
		& \leq &  12 (\alpha^k)^2 T \mathbf{E} \| \nabla f_i(z^k) \|^2 \notag   + 12 L^2 (\alpha^k)^2 T \mathbf{E} \| z^k - x^k \|^2 \notag \\
		&& + 6 (\alpha^k)^2 T \sigma^2 \notag  \\
		& \overset{\eqref{eq:Triangle_sq} }{\leq } &  12 (\alpha^k)^2 T \mathbf{E} \| \nabla f_i(z^k) \|^2 + 12 L^2 (\alpha^k)^2 T  \frac{1}{n}\sum_{i=1}^n \mathbf{E} \| e_i^k  \|^2 \notag\\
		&& + 6 (\alpha^k)^2 T \sigma^2.  \label{eq:EC_fedavg_localiter_globaliter}
\end{eqnarray}	
Plugging this bound into the main inequality yields
\begin{align*}
	V^{k+1} 
	\leq& V^k - \frac{\alpha^k}{4} \mathbf{E} \| \nabla f(x^k)\|^2    + \frac{L^2 \alpha^k \beta_1^k}{2} \frac{1}{n} \sum_{i=1}^n \mathbf{E} \|  e_i^k \|^2 \\
	& +  9 L^2 (\alpha^k)^3 \frac{1}{n}\sum_{i=1}^n \mathbf{E} \| \nabla f_i(z^k)\|^2  + \frac{3L(\alpha^k)^2}{2}\beta_2^k\sigma^2, 
\end{align*}	
where $\beta_1^k=1+9L^2\alpha^k$ and $\beta_2^k =  1+ \frac{L\alpha^k}{2}$. 
By the fact that $\alpha^k \leq 1/(6L)$,
\begin{eqnarray*}
	V^{k+1} 
	&\leq& V^k - \frac{\alpha^k}{4} \mathbf{E} \| \nabla f(x^k)\|^2  + \frac{ c_1\alpha^k }{2} \frac{1}{n} \sum_{i=1}^n \mathbf{E} \|  e_i^k \|^2 \\
	&& +  \frac{3 L (\alpha^k)^2}{2} \frac{1}{n}\sum_{i=1}^n \mathbf{E} \| \nabla f_i(z^k)\|^2  + \frac{13L(\alpha^k)^2}{8} \sigma^2,
\end{eqnarray*}	
where $c_1 = (1+1.5L)L^2$.
Next, define $\bar V^k = V^k + A^k \frac{1}{n} \sum_{i=1}^n \mathbf{E}\| e_i^k  \|^2$ with $A^k>0$. Then, 
\begin{align*}
	& \bar V^{k+1} 
	\leq V^k - \frac{\alpha^k}{4} \mathbf{E} \| \nabla f(x^k)\|^2 + A^{k+1} \frac{1}{n} \sum_{i=1}^n \mathbf{E}\| e_i^{k+1}  \|^2  \notag   \\
	&  + \frac{c_1\alpha^k }{2} \frac{1}{n} \sum_{i=1}^n \mathbf{E} \|  e_i^k \|^2  +  \frac{3 L (\alpha^k)^2}{2} \frac{1}{n}\sum_{i=1}^n \mathbf{E} \| \nabla f_i(z^k)\|^2 \notag \\
 & + \frac{13L(\alpha^k)^2}{8} \sigma^2. 
\end{align*} \label{eq:EC_fedavg_final_inequality}	
To complete the proof, we must bound $\frac{1}{n} \sum_{i=1}^n \mathbf{E}\| e_i^{k+1}  \|^2$.  
By the definition of $e_i^{k+1}$, 
\begin{align*}
& \frac{1}{n} \sum_{i=1}^n \mathbf{E}\| e_i^{k+1}  \|^2
 \overset{\eqref{eq:contractive_comp}}{\leq} \frac{1-\alpha}{n} \sum_{i=1}^n \mathbf{E}\| x_i^{k,T} - x^k + e_i^{k}  \|^2 \\
& \overset{ \eqref{eq:Young}}{\leq}  \frac{A_1(\alpha)}{n}\sum_{i=1}^n \mathbf{E}\| e_i^{k}  \|^2 + \frac{A_2(\alpha)}{n}\sum_{i=1}^n \mathbf{E}\| x_i^{k,T} - x^k \|^2 \\
& \leq \frac{(1-\alpha/2)}{n}\sum_{i=1}^n \mathbf{E}\| e_i^{k}  \|^2  +  A_2(\alpha)(\alpha^k)^2 B, 
\end{align*}	
where $A_1(\alpha)=(1-\alpha)(1+\alpha/2)$,
$A_2(\alpha) = (1-\alpha)(1+2/\alpha)$ and
$B = \frac{1}{nT}\sum_{i=1}^n \sum_{t=0}^{T-1} \mathbf{E}\|  \nabla F_i(x_i^{k,t};\xi_i^{k,t}) \|^2$. We now bound $B$  to bound $\frac{1}{n} \sum_{i=1}^n \mathbf{E}\| e_i^{k+1}  \|^2$: By \eqref{eq:Triangle_sq}, \eqref{eq:L_Lipschitz} and \eqref{eq:variance_stoc}, 
\begin{eqnarray*}
	B 
%	& \overset{\eqref{eq:Triangle_sq}}{\leq} &  \frac{4}{nT}\sum_{i=1}^n \sum_{t=0}^{T-1} \mathbf{E}\|  \nabla f_i(z^k) \|^2 \\ 
%	&& +  \frac{4}{nT}\sum_{i=1}^n \sum_{t=0}^{T-1} \mathbf{E}\|  \nabla f_i(z^k) - \nabla f_i(x^k) \|^2 \\
%	&& +   \frac{4}{nT}\sum_{i=1}^n \sum_{t=0}^{T-1} \mathbf{E}\|   \nabla f_i(x^k) - \nabla f_i(x_i^{k,t}) \|^2 \\
%	&& +  \frac{4}{nT}\sum_{i=1}^n \sum_{t=0}^{T-1} \mathbf{E}\|   \nabla F_i(x_i^{k,t}; \xi_i^{k,t}) - \nabla f_i(x_i^{k,t}) \|^2 \\
	& \leq & \frac{4}{n}\sum_{i=1}^n  \mathbf{E}\|  \nabla f_i(z^k) \|^2 +  4L^2 \mathbf{E}\|  z^k - x^k \|^2  + 4\sigma^2 \\
	&& + \frac{4L^2}{nT}\sum_{i=1}^n \sum_{t=0}^{T-1} \mathbf{E}\|   x_i^{k,t} - x^k \|^2 \\
	& \overset{\eqref{eq:Triangle_sq}}{\leq} &   \frac{4}{n}\sum_{i=1}^n  \mathbf{E}\|  \nabla f_i(z^k) \|^2 +  \frac{4L^2}{n} \sum_{i=1}^n \mathbf{E}\|  e_i^k \|^2  + 4\sigma^2 \\
	&& + \frac{4L^2}{nT}\sum_{i=1}^n \sum_{t=0}^{T-1} \mathbf{E}\|   x_i^{k,t} - x^k \|^2. 
\end{eqnarray*}	
Since $\gamma^k=\alpha^k/T$, $\alpha^k \leq 1/(6L) \leq 1/(\sqrt{6}L)$, and 
\begin{eqnarray*}
	&&  \frac{4L^2}{nT}\sum_{i=1}^n \sum_{t=0}^{T-1} \mathbf{E}\|   x_i^{k,t} - x^k \|^2 \\ & \overset{\eqref{eq:EC_fedavg_localiter_globaliter}}{\leq} & \frac{4}{3} \frac{1}{n}\sum_{i=1}^n \mathbf{E} \|  \nabla f_i(z^k) \|^2  + \frac{4L^2}{3} \frac{1}{n}\sum_{i=1}^n \mathbf{E} \| e_i^k \|^2  + \frac{2}{3}\sigma^2,
\end{eqnarray*}	
we get 
\begin{eqnarray*}
	B & \leq &   \frac{16}{3n}\sum_{i=1}^n  \mathbf{E}\|  \nabla f_i(z^k) \|^2 +  \frac{16L^2}{3n} \sum_{i=1}^n \mathbf{E}\|  e_i^k \|^2  + \frac{14}{3}\sigma^2.
\end{eqnarray*}	
Therefore, 
\begin{align*}
& \frac{1}{n} \sum_{i=1}^n \mathbf{E}\| e_i^{k+1}  \|^2 \leq \frac{[1-\alpha/2 +  Q^k] }{n}\sum_{i=1}^n \mathbf{E}\| e_i^{k}  \|^2  \\
& \hspace{0.5cm} + \frac{D_1^k}{n} \sum_{i=1}^n  \mathbf{E}\|  \nabla f_i(z^k) \|^2  +  D_2^k \sigma^2, 
\end{align*}	
where $Q^k = 16L^2(1-\alpha)(1+2/\alpha) (\alpha^k)^2/3$, $D_1^k = \frac{16(1-\alpha)(1+2/\alpha) (\alpha^k)^2}{3}$ and $D_2^k =  \frac{14(1-\alpha)(1+2/\alpha) (\alpha^k)^2}{3}$.

If $\alpha^k \leq \frac{1}{L} \sqrt{\frac{3\alpha}{64(1-\alpha)(1+2/\alpha)}}$, then 
\begin{align*}
	& \frac{1}{n} \sum_{i=1}^n \mathbf{E}\| e_i^{k+1}  \|^2 \leq \frac{(1-\alpha/4)}{n}\sum_{i=1}^n \mathbf{E}\| e_i^{k}  \|^2  \\
	& \hspace{0.5cm}+   \frac{D_1^k}{n}\sum_{i=1}^n  \mathbf{E}\|  \nabla f_i(z^k) \|^2   +  D_2^k\sigma^2.
\end{align*}	
By plugging the bound for $\frac{1}{n} \sum_{i=1}^n \mathbf{E}\| e_i^{k+1}  \|^2$ into the main inequality, 
\begin{eqnarray*}
		\bar V^{k+1} 
	&\leq& V^k - \frac{\alpha^k}{4} \mathbf{E} \| \nabla f(x^k)\|^2  + C_1^k \frac{1}{n} \sum_{i=1}^n \mathbf{E} \|  e_i^k \|^2  \\
	&& +  C_2^k \frac{(\alpha^k)^2}{n}\sum_{i=1}^n \mathbf{E} \| \nabla f_i(z^k)\|^2  + C_3^k(\alpha^k)^2 \sigma^2,
\end{eqnarray*}	
where $C_1^k = A^{k+1} \left(1-\frac{\alpha}{4} \right)+ \frac{(1+1.5L)L^2 \alpha^k}{2}$, $C_2^k =  \frac{16(1-\alpha)(1+2/\alpha)A^{k+1}}{3} + \frac{3 L }{2}$ and $C_3^k =  \frac{14(1-\alpha)(1+2/\alpha) A^{k+1}}{3}  +  \frac{13L}{8}$. 

If $A^k = \frac{4 (1+1.5L) L^2}{\alpha} \alpha^k$ and $0<\alpha^{k+1} \leq \alpha^k$ for all $k\in\mathbb{N}$, then $A^{k+1} \leq A^k$ and 
 \begin{eqnarray*}
 	C_1^k & = & \frac{4 (1+1.5L) L^2}{\alpha} \alpha^{k+1} (1-\alpha/4) + \frac{(1+1.5L)L^2 \alpha^k}{2} \\
 	& \leq & \frac{4 (1+1.5L) L^2}{\alpha} \alpha^{k} (1-\alpha/4) + \frac{(1+1.5L)L^2 \alpha^k}{2} \\
        & \leq & A^k - \frac{(1+1.5L)L^2 \alpha^k}{2} \leq A^k. 
 \end{eqnarray*}	
 Therefore, 
 \begin{eqnarray*}
 	\bar V^{k+1} 
 	&\leq& \bar V^k - \frac{\alpha^k}{4} \mathbf{E} \| \nabla f(x^k)\|^2  + \tilde C_3^k(\alpha^k)^2 \sigma^2 \\
 	&& +  \tilde C_2^k \frac{(\alpha^k)^2}{n}\sum_{i=1}^n \mathbf{E} \| \nabla f_i(z^k)\|^2,
 \end{eqnarray*}	
where  $\tilde C_2^k =  \frac{16(1-\alpha)(1+2/\alpha)A^{k}}{3} + \frac{3 L }{2}$ and $\tilde C_3^k =  \frac{14(1-\alpha)(1+2/\alpha) A^{k}}{3}  +  \frac{13L}{8}$. 
 By the fact that $\alpha^k \leq \hat\alpha := \frac{1}{L}\min\left( \frac{1}{6} , \sqrt{\frac{3\alpha}{64(1-\alpha)(1+2/\alpha)}}  \right)$, 
  \begin{eqnarray*}
 	\bar V^{k+1} 
 	&\leq& \bar V^k - \frac{\alpha^k}{4} \mathbf{E} \| \nabla f(x^k)\|^2  + \tilde C_3 (\alpha^k)^2 \sigma^2 \\
 	&& +  \tilde C_2 \frac{(\alpha^k)^2}{n}\sum_{i=1}^n \mathbf{E} \| \nabla f_i(z^k)\|^2,
 \end{eqnarray*}	
 where $\tilde C_2 =  \frac{16(1-\alpha)(1+2/\alpha)A}{3} + \frac{3 L }{2}$, $\tilde C_3 =  \frac{14(1-\alpha)(1+2/\alpha) A}{3}  +  \frac{13L}{8}$, and $A = \frac{4 (1+1.5L) L^2}{\alpha} \hat \alpha$.
Next, denote $f_i^{\inf}$ and $f^{\inf}$ as the lower-bound of each component function $f_i(x)$ and of the whole objective function $f(x)=(1/n)\sum_{i=1}^n f_i(x)$, respectively. Since
\begin{eqnarray*}
	\frac{1}{n}\sum_{i=1}^n \mathbf{E}\| \nabla f_i(z^k) \|^2 
	& \overset{\eqref{eq:L_Lipschitz_trick} }{\leq} &  \frac{2L}{n}\sum_{i=1}^n \mathbf{E}[f_i(z^k)-f_i^{\inf}] \\
	& = &2L \mathbf{E}[f(z^k)-f^{\inf}] + 2L \Delta^{\inf} \\
	& \leq & 2L \bar V^k + 2L \Delta^{\inf},
\end{eqnarray*}	
where $ \Delta^{\inf} = f^{\inf} - \frac{1}{n} \sum_{i=1}^n f_i^{\inf}$, we obtain the final convergence bound.   
%  \begin{eqnarray*}
% 	\bar V^{k+1} 
% 	&\leq& (1 + 2L\tilde C_2 (\alpha^k)^2 )\bar V^k - \frac{\alpha^k}{4} \mathbf{E} \| \nabla f(x^k)\|^2  \\
% 	&& + \tilde C_3 (\alpha^k)^2 \sigma^2 + 2L \tilde C_2 (\alpha^k)^2 \Delta^{\inf}.
% \end{eqnarray*}	
% We thus complete the proof. 

\section{Proof of Proposition~\ref{prop:EF_fedprox}}
\noindent Error-feedback FedProx, Algorithm~\ref{alg:EC_FL} with $\mathcal{T}_{\gamma F}(x)=\textbf{prox}_{\gamma F}(x)$ and $T=1$,  can be expressed equivalently as: 
\begin{eqnarray*}\label{eq:EF_FedProx_equi}
	z^{k+1} 
 & = & z^k + \frac{1}{n}\sum_{i=1}^n [\mathcal{T}_{\gamma^k F_i}(x^k)-x^k]  \notag \\
 & \overset{\eqref{eq:prox_trick}}{=} & z^k - \frac{\gamma^k}{n}\sum_{i=1}^n \nabla F_i(p^k_i;\xi_i^k),
\end{eqnarray*}	
where $p^k_i=\prox(x^k)$ and $z^k = x^k + \frac{1}{n}\sum_{i=1}^n e_i^k$.

From Assumption~\ref{assum:L_Lipschitz},  $f(x)$ has also $L$-Lipschitz continuous gradient. Let $f^{\inf}$ is the lower bound for $f(x)$. From \eqref{eq:L_Lipschitz_ineq} and \eqref{eq:EF_FedProx_equi}, we get
	\begin{eqnarray*}
	r^{k+1} 
	\leq  r^k  - \gamma^k \left\langle \nabla f(z^k),  g^k \right\rangle  + \frac{L (\gamma^k)^2}{2} \left\|  g^k \right\|^2.
\end{eqnarray*}	
where $r^k = f(z^k) - f^{\inf}$ and $g^k =  \frac{1}{n}\sum_{i=1}^n \nabla F_i(p^k_i;\xi_i^k)$. 
By taking the expectation and by the fact that $\nabla f(x) = (1/n)\sum_{i=1}^n \nabla f_i(x)$,
	\begin{eqnarray*}
	V^{k+1}
	%&\leq&  V^k   - \gamma^k \mathbf{E}\left\langle \nabla f(z^k),   \frac{1}{n}\sum_{i=1}^n \nabla f_i(p^k_i) \right\rangle \\
	%&& + \frac{L (\gamma^k)^2}{2} \mathbf{E}\left\|  \frac{1}{n}\sum_{i=1}^n \nabla F_i(p^k_i;\xi_i^k) \right\|^2 \\
	& \leq & V^k   - \gamma^k T_1  +  \gamma^k T_2  + \frac{L (\gamma^k)^2}{2} T_3, \\
\end{eqnarray*}	
where $V^k = \mathbf{E}[r^k]$,  $T_1 = \mathbf{E}\left\langle \nabla f(z^k),  \nabla f(x^k) \right\rangle$, $T_2 =  \mathbf{E}\left\langle \nabla f(z^k),   \frac{1}{n}\sum_{i=1}^n \nabla f_i(x^k)- \nabla f_i(p^k_i) \right\rangle$ and  $T_3 = \mathbf{E}\left\|  \frac{1}{n}\sum_{i=1}^n \nabla F_i(p^k_i;\xi_i^k) \right\|^2$.
Since
\begin{eqnarray*}
	 - \gamma^k T_1 
	 & \overset{\eqref{eq:Exact_sq}}{=} & - \frac{\gamma^k}{2} \mathbf{E} \| \nabla f(z^k) \|^2 -  \frac{\gamma^k}{2} \mathbf{E} \| \nabla f(x^k) \|^2 \\
	 && + \frac{\gamma^k}{2}\mathbf{E}\|\nabla f(z^k)  - \nabla f(x^k) \|^2, \quad \text{and} \\
	 \gamma^k T_2 & \overset{\eqref{eq:InExact_sq} + \eqref{eq:Triangle_sq}}{\leq} & \frac{\gamma^k}{2}\mathbf{E}\| \nabla f(z^k)\|^2  \\
	 && + \frac{\gamma^k}{2}\frac{1}{n}\sum_{i=1}^n\mathbf{E} \left\|   \nabla f_i(x^k)- \nabla f_i(p^k_i)   \right\|^2,
\end{eqnarray*}	
we have 
	\begin{eqnarray*}
	V^{k+1}
	&\leq&  V^k   - \frac{\gamma^k}{2} \mathbf{E}\| \nabla f(x^k ) \|^2 +  \frac{L (\gamma^k)^2}{2} T_3 \\
	&& +  \frac{\gamma^k}{2}\mathbf{E}\|\nabla f(z^k)  - \nabla f(x^k) \|^2 \\
	 && + \frac{\gamma^k}{2}  \frac{1}{n}\sum_{i=1}^n  \mathbf{E} \left\| \nabla f_i(x^k)- \nabla f_i(p^k_i)   \right\|^2. 
\end{eqnarray*}	
Next, since 
\begin{align*}
 \frac{L (\gamma^k)^2}{2} T_3  
%	& \overset{\eqref{eq:Triangle_sq}}{\leq}   \frac{3L (\gamma^k)^2}{2}  \mathbf{E}\left\|  \frac{1}{n}\sum_{i=1}^n \nabla F_i(p^k_i;\xi_i^k) - \nabla f_i(p^k_i) \right\|^2 \\
%	&  \hspace{0.5cm}+ \frac{3L (\gamma^k)^2}{2}  \mathbf{E}\left\|  \frac{1}{n}\sum_{i=1}^n  \nabla f_i(p^k_i) - \nabla f_i(x^k) \right\|^2 \\
%	& \hspace{0.5cm}+ \frac{3L (\gamma^k)^2}{2}  \mathbf{E}\left\|  \frac{1}{n}\sum_{i=1}^n  \nabla f_i(x^k) \right\|^2 \\ 
	& \overset{\eqref{eq:Triangle_sq} + \eqref{eq:variance_stoc}}{\leq}   \frac{3L (\gamma^k)^2}{2}  \frac{1}{n}\sum_{i=1}^n \mathbf{E}\left\|    \nabla f_i(p^k_i) - \nabla f_i(x^k) \right\|^2 \\
	& \hspace{0.7cm} +  \frac{3L (\gamma^k)^2}{2} \sigma^2 + \frac{3L (\gamma^k)^2}{2}  \mathbf{E}\left\|   \nabla f(x^k) \right\|^2,
\end{align*}	
we get 
	\begin{eqnarray*}
	V^{k+1}
	&\leq&  V^k   - \frac{\gamma^k (1-3L \gamma^k)}{2} \mathbf{E}\| \nabla f(x^k ) \|^2  \\
	&&  +  \frac{\gamma^k}{2}\mathbf{E}\|\nabla f(z^k)  - \nabla f(x^k) \|^2 + \frac{3L (\gamma^k)^2}{2} \sigma^2\\
	&& + \frac{\gamma^k \beta^k}{2}   \frac{1}{n}\sum_{i=1}^n  \mathbf{E} \left\| \nabla f_i(x^k)- \nabla f_i(p^k_i)   \right\|^2,
\end{eqnarray*}	
where $\beta^k = 1+ 3L \gamma^k$. 
By Assumption~\ref{assum:L_Lipschitz},
	\begin{align*}
	V^{k+1}
	\leq&  V^k   - \frac{\gamma^k (1-3L\gamma^k)}{2} \mathbf{E}\| \nabla f(x^k ) \|^2 \\
	& +  \frac{L^2\gamma^k}{2}\mathbf{E}\| z^k  - x^k \|^2 + \frac{3L (\gamma^k)^2}{2} \sigma^2 \\
	& + \frac{ L^2 \gamma^k \beta^k}{2}   \frac{1}{n}\sum_{i=1}^n  \mathbf{E} \left\| x^k- p^k_i  \right\|^2.
\end{align*}	
By the fact that  $z^k = x^k + \frac{1}{n}\sum_{i=1}^n e_i^k$ and by \eqref{eq:Triangle_sq},
	\begin{align*}
	V^{k+1}
	\leq&  V^k   - \frac{\gamma^k (1-3L\gamma^k)}{2} \mathbf{E}\| \nabla f(x^k ) \|^2 \\
	& +  \frac{L^2\gamma^k}{2}  \frac{1}{n}\sum_{i=1}^n \mathbf{E}\| e_i^k  \|^2 + \frac{3L (\gamma^k)^2}{2} \sigma^2 \\
	& + \frac{ L^2 \gamma^k \beta^k}{2}   \frac{1}{n}\sum_{i=1}^n  \mathbf{E} \left\| x^k- p^k_i  \right\|^2.
\end{align*}	

If $\gamma^k \leq 1/(6L)$, then $\beta^k \leq 3/2$ and
	\begin{eqnarray*}
	V^{k+1}
	&\leq&  V^k   - \frac{\gamma^k}{4} \mathbf{E}\| \nabla f(x^k ) \|^2  +  \frac{L^2\gamma^k}{2}  \frac{1}{n}\sum_{i=1}^n \mathbf{E}\| e_i^k  \|^2 \\
	&& + \frac{ 3L^2 \gamma^k}{4}   \frac{1}{n}\sum_{i=1}^n  \mathbf{E} \left\| x^k- p^k_i  \right\|^2  + \frac{3L (\gamma^k)^2}{2} \sigma^2.
\end{eqnarray*}	
Next, we bound $ \mathbf{E} \left\| x^k- p^k_i  \right\|^2$ to complete the convergence bound. 
By the fact that \eqref{eq:FedProx_trick} and \eqref{eq:Triangle_sq}

\begin{eqnarray*}
 \mathbf{E} \left\| x^k- p^k_i  \right\|^2
 %& \overset{	\eqref{eq:FedProx_trick} }{\leq} & 6(\gamma^k)^2 \mathbf{E}\| \nabla f_i(x^k) \|^2 + 6 (\gamma^k)^2 \sigma^2  \notag \\ 
 & \leq & 12 (\gamma^k)^2 \mathbf{E}\| \nabla f_i(z^k) \|^2  + 6 (\gamma^k)^2 \sigma^2  \notag  \\
 && + 12 (\gamma^k)^2  \mathbf{E}\| \nabla f_i(x^k) - \nabla f_i(z^k) \|^2 \notag \\
 & \overset{\eqref{eq:L_Lipschitz}}{\leq} & 12 (\gamma^k)^2 \mathbf{E}\| \nabla f_i(z^k) \|^2  + 6 (\gamma^k)^2 \sigma^2  \notag \\
 && + 12 L^2 (\gamma^k)^2 \frac{1}{n} \sum_{i=1}^n  \mathbf{E}\| e_i^k \|^2. \notag 
\end{eqnarray*}	
By the fact that $ \gamma^k \leq \frac{1}{6L}$, 
\begin{align}
    \mathbf{E} \left\| x^k- p^k_i  \right\|^2
 %& \overset{	\eqref{eq:FedProx_trick} }{\leq} & 6(\gamma^k)^2 \mathbf{E}\| \nabla f_i(x^k) \|^2 + 6 (\gamma^k)^2 \sigma^2  \notag \\ 
   \leq & \frac{2\gamma^k}{L}\mathbf{E}\| \nabla f_i(z^k) \|^2  + \frac{\gamma^k \sigma^2}{L}   + \frac{1}{3n} \sum_{i=1}^n  \mathbf{E}\| e_i^k \|^2. \label{eq:bound_x_p_EFFEDPROX}
\end{align}
Plugging \eqref{eq:bound_x_p_EFFEDPROX} into the main inequality yields 
	\begin{eqnarray*}
	V^{k+1}
	&\leq&  V^k   - \frac{\gamma^k}{4} \mathbf{E}\| \nabla f(x^k ) \|^2 +  \frac{3L^2\gamma^k}{4}  \frac{1}{n}\sum_{i=1}^n \mathbf{E}\| e_i^k  \|^2 \\
	&& + \frac{ 3L (\gamma^k)^2}{2}   \frac{1}{n}\sum_{i=1}^n  \mathbf{E} \left\|  \nabla f_i (z^k)\right\|^2  +  \frac{9L}{4}(\gamma^k)^2\sigma^2.
\end{eqnarray*}	
Next, define $\bar V^k = V^k + A^k \frac{1}{n} \sum_{i=1}^n \mathbf{E}\| e_i^k  \|^2$ with $A^k>0$. Then, 
	\begin{eqnarray*}
	\bar V^{k+1}
	&\leq&  V^k   - \frac{\gamma^k}{4} \mathbf{E}\| \nabla f(x^k ) \|^2 \notag \\
	&&  +  A^{k+1}  \frac{1}{n}\sum_{i=1}^n \mathbf{E}\| e_i^{k+1}  \|^2  +  \frac{3L^2\gamma^k}{4}  \frac{1}{n}\sum_{i=1}^n \mathbf{E}\| e_i^k  \|^2 \notag \\
	&& + \frac{ 3L (\gamma^k)^2}{2}   \frac{1}{n}\sum_{i=1}^n  \mathbf{E} \left\|  \nabla f_i (z^k)\right\|^2  +  \frac{9L}{4} (\gamma^k)^2\sigma^2. \label{eq:final_ineq_EF_fedprox}
\end{eqnarray*}
To complete the proof, we must bound $\frac{1}{n}\sum_{i=1}^n \mathbf{E}\| e_i^{k+1}  \|^2$. 
By the fact that error-feedback FedProx is Algorithm~\ref{alg:EC_FL} with $\mathcal{T}_{\gamma F}(x)=\textbf{prox}_{\gamma F}(x)$ and $T=1$, 
%\begin{eqnarray*}
	$\| x_i^{k,T} - x^k \|^2 
	 =  (\gamma^k)^2 \| \nabla F_i(p_i^k;\xi^k_i) \|^2$.
%\end{eqnarray*}	
Therefore, by \eqref{eq:contractive_comp} and  \eqref{eq:Young}, and by the fact that $(1-\alpha)(1+\alpha/2) \leq 1-\alpha/2$, 
\begin{align*}
	\| e_i^{k+1} \|^2
%	& \overset{\eqref{eq:contractive_comp}}{\leq} & (1-\alpha) \| x_i^{k,T} - x^k + e_i^k   \|^2 \\
%	&  \overset{\eqref{eq:contractive_comp} + \eqref{eq:Young} }{\leq} & (1-\alpha)(1+\alpha/2) \| e_i^k \|^2 \\
%	&&  + (1-\alpha)(1+2/\alpha) \| x_i^{k,T} - x^k \|^2  \\
	  {\leq}& (1{-}\alpha/2) \| e_i^k \|^2 
	  + (1-\alpha)(1+2/\alpha) \| x_i^{k,T} - x^k \|^2 \\
	  {\leq}&  (1{-}\frac{\alpha}{2}) \| e_i^k \|^2 
	  {+} (1{-}\alpha)(1{+}\frac{2}{\alpha}) (\gamma^k)^2 \| \nabla F_i(p_i^k;\xi^k_i) \|^2. 
\end{align*}	
We hence get 
\begin{align*}
& \frac{1}{n}\sum_{i=1}^n \mathbf{E}\| e_i^{k+1}  \|^2  \leq  (1-\alpha/2) \frac{1}{n}\sum_{i=1}^n \mathbf{E} \| e_i^k \|^2 \\
	&  + (1-\alpha)(1+2/\alpha) (\gamma^k)^2 \frac{1}{n}\sum_{i=1}^n \mathbf{E}\| \nabla F_i(p_i^k;\xi^k_i) \|^2. 
\end{align*}	
We bound $\frac{1}{n}\sum_{i=1}^n \mathbf{E}\| e_i^{k+1}  \|^2$ by bounding  $\mathbf{E}\| \nabla F_i(p_i^k;\xi^k_i) \|^2$: 
By \eqref{eq:Triangle_sq}, \eqref{eq:L_Lipschitz} and \eqref{eq:variance_stoc}
\begin{align*}
	 \mathbf{E}\| \nabla F_i(p_i^k;\xi^k_i) \|^2
%	 & \overset{\eqref{eq:Triangle_sq}}{\leq} & 4  \mathbf{E}\| \nabla F_i(p_i^k;\xi^k_i) - \nabla f_i(p_i^k) \|^2 \\
%	 &&  + 4  \mathbf{E}\| \nabla f_i(p_i^k) - \nabla f_i(x^k ) \|^2  \\
%	 && + 4  \mathbf{E}\| \nabla f_i(x^k) - \nabla f_i(z^k) \|^2 \\
%	 &&  + 4  \mathbf{E}\| \nabla f_i(z^k) \|^2 \\
	   \leq&   4 \sigma^2  + 4  \mathbf{E}\| \nabla f_i(z^k) \|^2  
	   {+} 4 L^2  \mathbf{E}\| p_i^k {-} x^k \|^2  \\
	  & + 4 L^2 \mathbf{E}\| x^k - z^k \|^2 \\
	   \overset{\eqref{eq:Triangle_sq}}{\leq} &  4 \sigma^2  + 4  \mathbf{E}\| \nabla f_i(z^k) \|^2  
	+ 4 L^2  \mathbf{E}\| p_i^k {-} x^k \|^2  \\
	& + 4 L^2 \frac{1}{n}\sum_{i=1}^n \mathbf{E}\|  e_k^i \|^2.
%	& \overset{\eqref{eq:bound_x_p_EFFEDPROX}}{\leq} & (4+ 4L\gamma^k) \sigma^2 \\
%	&& + (4 + 8L\gamma^k) \mathbf{E} \| \nabla f_i(z^k) \|^2 \\
%	&& + (4+4L^2/3) \frac{1}{n}\sum_{i=1}^n \mathbf{E}\| e_k^i\|^2 \\
%	& \overset{\eqref{eq:bound_x_p_EFFEDPROX}  + \ \gamma^k \leq \frac{1}{6L}}{\leq} & (4+ 2/3) \sigma^2 \\
%	&& + (4 + 4/3) \mathbf{E} \| \nabla f_i(z^k) \|^2 \\
%	&& + (4+4L^2/3) \frac{1}{n}\sum_{i=1}^n \mathbf{E}\| e_k^i\|^2, 
\end{align*}	
By \eqref{eq:bound_x_p_EFFEDPROX}  and by the fact that $ \gamma^k \leq \frac{1}{6L}$, 
\begin{align*}
	 \mathbf{E}\| \nabla F_i(p_i^k;\xi^k_i) \|^2
 \leq & (4+ 2/3) \sigma^2 
	+ (4 + 4/3) \mathbf{E} \| \nabla f_i(z^k) \|^2 \\
	& + (4+4L^2/3) \frac{1}{n}\sum_{i=1}^n \mathbf{E}\| e_k^i\|^2.
\end{align*}	
Plugging the bound for $ \mathbf{E}\| \nabla F_i(p_i^k;\xi^k_i) \|^2$, we have
\begin{align*}
	& \frac{1}{n}\sum_{i=1}^n \mathbf{E}\| e_i^{k+1}  \|^2  \leq  (1-\alpha/2 + C_1 (\gamma^k)^2) \frac{1}{n}\sum_{i=1}^n \mathbf{E} \| e_i^k \|^2 \\
	& +  C_2 (\gamma^k)^2 \frac{1}{n}\sum_{i=1}^n \mathbf{E}\| \nabla f_i(z^k) \|^2  + C_3 (\gamma^k)^2 \sigma^2,
\end{align*}
where $C_1 = (1-\alpha)(1+2/\alpha)(4+4L^2/3)$, $C_2 =(1-\alpha)(1+2/\alpha)(4+4/3)$ and  $C_3 =(1-\alpha)(1+2/\alpha)(4+2/3)$.

If $\gamma^k \leq \frac{1}{2}\sqrt{\frac{\alpha}{C_1}}$, then 
\begin{align*}
	& \frac{1}{n}\sum_{i=1}^n \mathbf{E}\| e_i^{k+1}  \|^2  \leq  (1-\alpha/4) \frac{1}{n}\sum_{i=1}^n \mathbf{E} \| e_i^k \|^2 \\
	& +  C_2 (\gamma^k)^2 \frac{1}{n}\sum_{i=1}^n \mathbf{E}\| \nabla f_i(z^k) \|^2  + C_3 (\gamma^k)^2 \sigma^2.
\end{align*}
Plugging this result into \eqref{eq:EC_FedAvg_equi} yields 
	\begin{align*}
	\bar V^{k+1}
	\leq&  V^k   - \frac{\gamma^k}{4} \mathbf{E}\| \nabla f(x^k ) \|^2   +  \tilde B_1^k  \frac{1}{n}\sum_{i=1}^n \mathbf{E}\| e_i^k  \|^2  \\
	& +  \tilde B_2^k \frac{(\gamma^k)^2}{n}\sum_{i=1}^n  \mathbf{E} \left\|  \nabla f_i (z^k)\right\|^2 + \tilde B_3^k (\gamma^k)^2\sigma^2,
\end{align*}
where $\tilde B_1^k = A^{k+1}(1-\alpha/4) +  \frac{3L^2\gamma^k}{4}$, $\tilde B_2^k =  \frac{ 3L}{2} + A^{k+1} C_2 $ and $\tilde B_3^k =  \frac{9L}{4} + A^{k+1} C_3$. 

If $A^k = \frac{3L^2 \gamma^k}{\alpha}$ and $\gamma^{k+1} \leq \gamma^k$ for all $k\in\mathbb{N}$, then $A^{k+1} \leq A^k$ and 
\begin{align*}
	%A^{k+1} & \leq &  A^k, \quad \text{and} \\
	A^{k+1}(1-\frac{\alpha}{4}) +   \frac{3L^2\gamma^k}{4}   
	 = & \frac{3L^2 \gamma^{k+1}}{\alpha}(1-\frac{\alpha}{4}) +   \frac{3L^2\gamma^k}{4}   \\
	 \leq & \frac{3L^2 \gamma^{k}}{\alpha}(1-\frac{\alpha}{4}) +   \frac{3L^2\gamma^k}{4} %\\
	 %= & \frac{3L^2 \gamma^{k}}{\alpha} 
  = A^k.
\end{align*}	
Therefore, 
	\begin{align*}
	\bar V^{k+1}
	\leq&  \bar V^k   - \frac{\gamma^k}{4} \mathbf{E}\| \nabla f(x^k ) \|^2   \\
 & +  B_1^k \frac{(\gamma^k)^2}{n}\sum_{i=1}^n  \mathbf{E} \left\|  \nabla f_i (z^k)\right\|^2 
	+ B_2^k(\gamma^k)^2\sigma^2,
\end{align*}
where $B_1^k=\frac{ 3L}{2} + A^{k} C_2$ and $B_2^k =  \frac{9L}{4} + A^{k} C_3$.
By the fact that $\gamma^k \leq \gamma$ where $\gamma = \min\left( \frac{1}{6L}, \frac{1}{2}\sqrt{\frac{\alpha}{C_1}} \right)$,  we have $A^k \leq A := 3L^2 \gamma/\alpha$ and 
	\begin{align*}
	\bar V^{k+1}
	\leq&   \bar V^k   - \frac{\gamma^k}{4} \mathbf{E}\| \nabla f(x^k ) \|^2   \\
 & +  B_1 \frac{(\gamma^k)^2}{n}\sum_{i=1}^n  \mathbf{E} \left\|  \nabla f_i (z^k)\right\|^2 
 + B_2(\gamma^k)^2\sigma^2,
\end{align*}
where $B_1=\frac{ 3L}{2} + A C_2$ and $B_2=\frac{9L}{4} + A C_3$. 
Next, denote $f_i^{\inf}$ as the lower-bound of each component function $f_i(x)$. Since
\begin{eqnarray*}
	\frac{1}{n}\sum_{i=1}^n \mathbf{E}\| \nabla f_i(z^k) \|^2 
	& \overset{\eqref{eq:L_Lipschitz_trick} }{\leq} &  \frac{2L}{n}\sum_{i=1}^n \mathbf{E}[f_i(z^k)-f_i^{\inf}] \\
	& = &2L \mathbf{E}[f(z^k)-f^{\inf}] + 2L \Delta^{\inf} \\
	& \leq & 2L \bar V^k + 2L \Delta^{\inf},
\end{eqnarray*}	
where $ \Delta^{\inf} = f^{\inf} - \frac{1}{n} \sum_{i=1}^n f_i^{\inf}$, we complete the proof.

\end{document}